\documentclass{article}

    \PassOptionsToPackage{numbers, sort, compress}{natbib}

 \usepackage[main, final]{neurips_2026}

\usepackage[utf8]{inputenc} %
\usepackage[T1]{fontenc}    %
\usepackage{hyperref}       %
\usepackage{url}            %
\usepackage{booktabs}       %
\usepackage{amsfonts}       %
\usepackage{nicefrac}       %
\usepackage{microtype}      %
\usepackage{xcolor}         %
\usepackage{fontawesome5}   %

\usepackage{amsmath}
\usepackage{amssymb}
\usepackage{mathtools}
\usepackage{amsthm}
\usepackage{bbm}
\usepackage{subcaption}
\usepackage{float}

\usepackage{algorithm,algorithmic}

\usepackage[capitalize,noabbrev]{cleveref}
\usepackage{enumitem}
\usepackage{mathrsfs}
\usepackage[accept]{editing}
\usepackage{xcolor}
\usepackage{wrapfig}
\usepackage{multirow}
\usepackage{booktabs} %
\usepackage{graphicx} %
\usepackage{adjustbox} %
\usepackage{caption}   %
\usepackage{makecell} %

\usepackage{aliascnt}

\theoremstyle{plain}
\newtheorem{theorem}{Theorem}[section]

\newaliascnt{proposition}{theorem}

\aliascntresetthe{proposition}

\newaliascnt{lemma}{theorem}

\aliascntresetthe{lemma}

\newaliascnt{corollary}{theorem}

\aliascntresetthe{corollary}

\theoremstyle{definition}

\newaliascnt{definition}{theorem}
\newtheorem{definition}[definition]{Definition}
\aliascntresetthe{definition}

\newaliascnt{assumption}{theorem}
\newtheorem{assumption}[assumption]{Assumption}
\aliascntresetthe{assumption}

\newtheorem{example}{Example}

\theoremstyle{remark}

\newaliascnt{remark}{theorem}
\newtheorem{remark}[remark]{Remark}
\aliascntresetthe{remark}

\Crefname{equation}{Eq.}{Eqs.}
\Crefname{align}{Eq.}{Eqs.}
\Crefname{figure}{Fig.}{Figs.}
\Crefname{tabular}{Tab.}{Tabs.}
\Crefname{theorem}{Thm.}{Thms.}
\Crefname{lemma}{Lem.}{Lems.}
\Crefname{proposition}{Prop.}{Props.}
\Crefname{definition}{Def.}{Defs.}
\Crefname{algorithm}{Algo.}{Algs.}
\Crefname{corollary}{Corol.}{Corol.}
\Crefname{section}{Sec.}{Sec.}
\Crefname{appendix}{App.}{Apps.}
\Crefname{prop}{Prop.}{Props.}
\Crefname{task}{Task.}{Tasks.}
\Crefname{setting}{Setting}{Settings}
\Crefname{example}{Example}{Expamples}
\Crefname{assumption}{Assumption}{Assumptions}

\def\*#1{\boldsymbol{#1}}
\def\1#1{\mathcal{#1}}
\def\2#1{\mathscr{#1}}
\def\3#1{\mathbb{#1}}

\makeatletter
\renewcommand*{\@fnsymbol}[1]{\ifcase#1\or\else\fi}
\makeatother

\usepackage{pgfplots}
\DeclareUnicodeCharacter{2212}{−}
\usepgfplotslibrary{groupplots,dateplot}
\usetikzlibrary{patterns,shapes.arrows}
\pgfplotsset{compat=newest}

\usetikzlibrary{positioning,arrows,shapes,shapes.arrows,arrows.meta,trees,shapes.misc,shapes.geometric,decorations.pathreplacing,automata}
\tikzset{%
  >={Latex[width=1.5mm,length=2mm]},
  vertex/.style={draw,circle,inner sep=0mm,semithick,minimum width=4mm},
  point/.style = {circle, draw, inner sep=0.04cm,fill,node contents={}},
  uvertex/.style={outer sep=0},
  bidir/.style={<->,dashed, line width=0.25mm},
  dir/.style={->, line width=0.25mm},
  regime/.style={shape=rectangle,fill=black,inner sep=0pt,minimum size=3pt,draw},
  action/.style={shape=circle,fill=black,inner sep=0pt,minimum size=8pt,draw},
  node distance=1cm,
  font=\scriptsize\sffamily
}

\definecolor{betterred}{RGB}{228,26,28}
\definecolor{betterblue}{RGB}{55,126,184}
\definecolor{bettergreen}{RGB}{77,175,74}
\definecolor{betterpurple}{RGB}{152,78,163}
\definecolor{LightCyan}{rgb}{0.88,1,1}

\pgfdeclarelayer{back}
\pgfsetlayers{back,main}

\makeatletter
\pgfkeys{%
  /tikz/on layer/.code={
    \def\tikz@path@do@at@end{\endpgfonlayer\endgroup\tikz@path@do@at@end}%
    \pgfonlayer{#1}\begingroup%
  }%
}
\makeatother

\makeatletter
\newcommand{\xdashleftrightarrow}[2][]{\ext@arrow 3359\leftrightarrowfill@@{#1}{#2}}
\def\rightarrowfill@@{\arrowfill@@\relax\relbar\rightarrow}
\def\leftarrowfill@@{\arrowfill@@\leftarrow\relbar\relax}
\def\leftrightarrowfill@@{\arrowfill@@\leftarrow\relbar\rightarrow}
\def\arrowfill@@#1#2#3#4{%
  $\m@th\thickmuskip0mu\medmuskip\thickmuskip\thinmuskip\thickmuskip
   \relax#4#1
   \xleaders\hbox{$#4#2$}\hfill
   #3$%
}
\makeatother

\usepackage{titletoc}
\newcommand\DoToC{%
  \setcounter{tocdepth}{4} %
  \startcontents
  \printcontents{}{1}{\textbf{Contents}\vskip3pt\hrule\vskip5pt}
  \vskip3pt\hrule\vskip5pt
}

\title{Counterfactual Debugging \\the World Model Transfer Gap}

\author{
  Mingxuan Li$^{1*}$ \quad Kai-Zhan Lee$^{1*}$ \thanks{*Equal contribution.} \quad Michael Dennis$^{2}$ \quad Elias Bareinboim$^1$ \\
  $^1$ Columbia University, $^2$ Google DeepMind\\
  $^1$\texttt{\{ml,kz,eb\}@cs.columbia.edu}, $^2$\texttt{dennismi@google.com}
}

\begin{document}

\maketitle

\begin{abstract}
  Policies achieving strong performance in simulators or learned world models can fail when deployed into the real environment. While there must be environmental differences that account for these failures, not all differences are equally responsible. A benign error of irrelevant visual details may co-exist with a failure to predict a single critical transition causing an inevitable catastrophic failure. In this paper, we introduce \emph{counterfactual debugging} as an approach for \emph{world model transfer gap attribution} to identify the time steps whose transition or reward errors are \emph{causally responsible} for a performance degradation exhibited in a real trajectory, rather than only visually or statistically different. 
  We develop a scalable algorithm for computing these attributions that exploits the sparsity of causal errors by recursively divide-and-conquer, significantly reducing computational costs, achieving an exponential speedup \xadd{while maintaining a high attribution quality}.
  Experiments across distinct environments \xadd{with various common} world model failures (observation corruption, reward misprediction, and physics violations) demonstrate that counterfactual debugging correctly identifies the errors that cause the transfer gap, \xadd{providing principled, actionable insights for continual learning in world models}. 

  \vspace{0.5ex}
  {\centering\hypersetup{hidelinks}\href{https://mingxuan.me/debugwm.html}{\textcolor{betterblue}{\faGlobe}\enspace Project page:\enspace\textcolor{betterblue}{\texttt{mingxuan.me/debugwm.html}}}\par}
\end{abstract}

\section{Introduction}

A world model can be thought of as a simulator of the domain in which the agent aims to operate.
Learning such a predictive model of the environment's transition dynamics and reward function, can unlock the ability to simulate experience for planning or policy optimization \citep{suttonDynaIntegratedArchitecture1991, rlbook}. In practice, recent world models have achieved remarkable success in complex domains, from mastering Atari games with discrete latent representations \citep{dreamer, schrittwieserMasteringAtariGo2020} to planning with learned dynamics in continuous control \citep{jannerWhenTrustYour2019a, haRecurrentWorldModels2018a}, to generating fully imagined worlds\citep{bruce2024genie}. Yet, a persistent challenge remains: {when the simulator or learned world model is wrong, policies trained on it can fail catastrophically once deployed in the real environment, despite success in simulation} \citep{jannerWhenTrustYour2019a}.

The difficulty is not merely that world models make errors, every learned model does, but that only some of these errors matter. A world model may hallucinate visual details that are irrelevant to the task, or subtly mispredict a single transition that cascades into a qualitatively different outcome. Such a single misprediction could lead to autonomous driving car crash or robots hurting its operator but it may not be distinguishable visually nor from the average prediction accuracy. 
\begin{figure}[t]
    \centering
    \includegraphics[width=.82\linewidth]{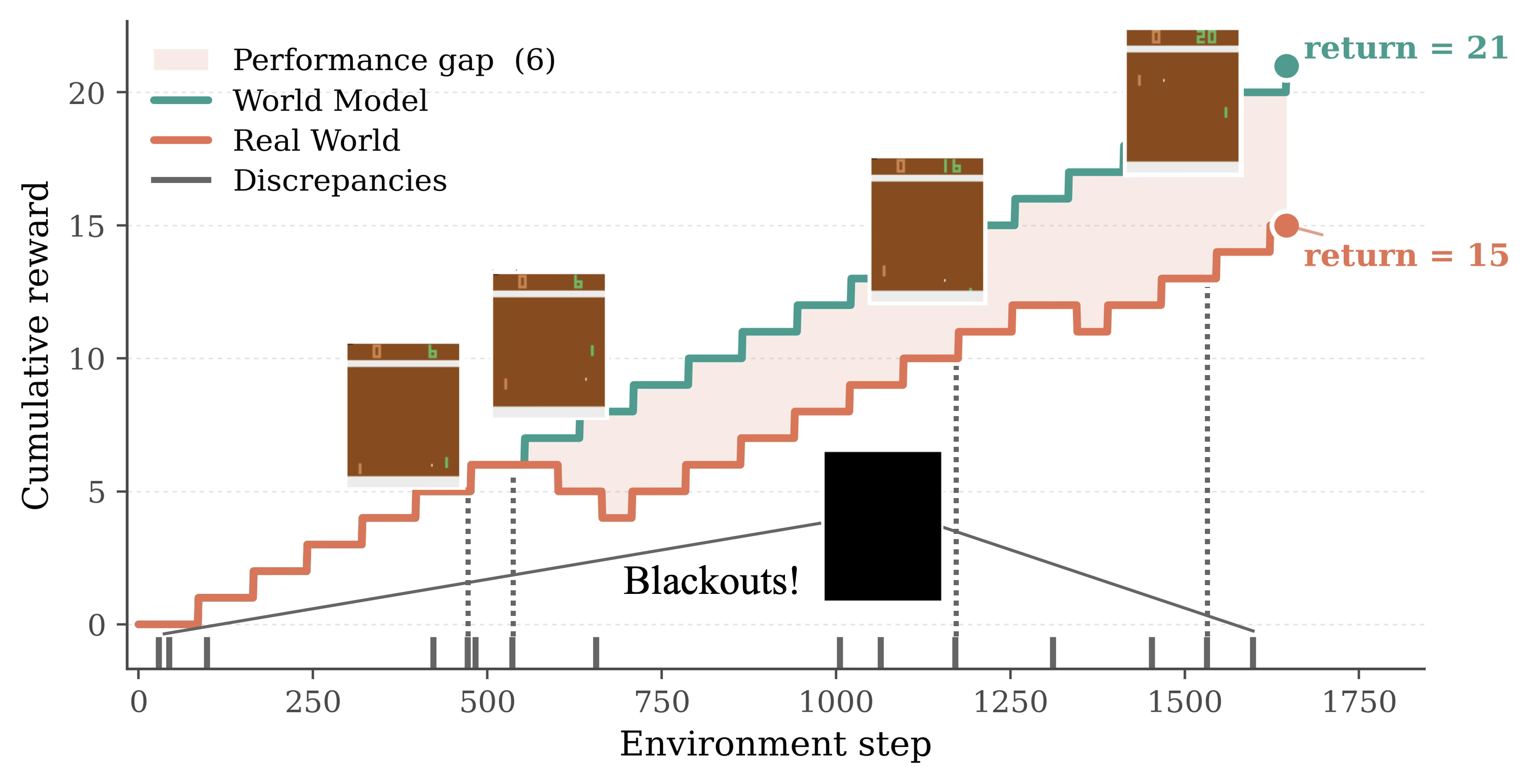}
    \caption{\textbf{Which transition discrepancy should be blamed for the performance degradation?} A PPO policy on Atari Pong running under a world model (green) scores full mark, $21$, while the same policy running against a turbulent ``real world", where random blackouts occlude the screen (red) only scores $15$. Crucially, most of the occluded frames have little bearing on the final outcome, only a small, hard-to-identify subset is \emph{causally responsible} for the collapse. \textbf{In this example, somewhat surprisingly, only two of the blackout frames are causally responsible for the performance degradation that happens more than 50 steps later}. See more details in \cref{sec:results-obsocclusion}.}
    \label{fig:teaser-cumreward}
    \vspace{-18pt}
\end{figure}
\cref{fig:teaser-cumreward} makes this concrete: a few occluded frames in a simple Pong game rollout suffice to knock the agent's return from $21$ down to $15$, even though visually most of the two trajectories look nearly identical. A raw visual comparison flags every perturbed frame equally, but only a small subset of those frames is the actual cause of the transfer gap in the real world; pinpointing that causal subset, rather than the full set of visual or statistical anomalies, is the diagnostic we should aim for. 

Existing tools fall short of finding such causal subsets: aggregated prediction-error metrics \citep{schrittwieserMasteringAtariGo2020,dreamer} and model-uncertainty heuristics \citep{jannerWhenTrustYour2019a} flag many or all bad transitions but cannot single out which ones drive the transfer gap, while saliency based methods \citep{DBLP:conf/icml/GreydanusKDF18visatari} and policy explanation over state features methods \citep{DBLP:conf/aaai/Madumal0SV20,explaining_rl_with_shapley_values} surface correlations in visual features or actions rather than causal attributions on the world model's dynamics. The most closely related solution is the line of feature attribution methods (e.g., SHAP\citep{lundberg2017unified} and LIME\citep{ribeiro2016should}). However, they condition on static inputs and cannot model the sequential, causal structure of a multi-step trajectory, where an error at step $t$ propagates forward through subsequent transitions.
Therefore, none of these answers the targeted question: at which time steps do the world model's prediction errors cause the policy to underperform in the real world?
\footnote{See \Cref{sec:extended-related-work} for an extended discussion.}

In this paper, we develop a method that answers the above question directly by allocating counterfactual Shapley credits to world model's predicted dynamics at each time step (rather than actions~\citep{li2026credit}). We treat each time step's model-predicted transition as a player in a game and measure its causal contribution to the transfer gap. We do so by constructing \emph{counterfactual rollouts}: at each time step, we can choose whether to use the world model's prediction or the sampled real world's transition, then measure how the trajectory unfolds, and potentially changing the rewards. By systematically intervening at different subsets of time steps and measuring the resulting return differences, we obtain per-time-step counterfactual Shapley values ($\phi$-values) that quantify how much each time step's dynamics causally contributes to the transfer gap.
This formulation yields actionable diagnostics. 
The world model's prediction deviating from the real world at those most blamed steps is the root cause of the transfer gap. This also directly informs immediate remediation: steps with high attribution are where the model needs improvements, such as more training data, architectural capacity, or calibration. 
Specifically, our contributions are:
\begin{itemize}[label=\textbullet, leftmargin=20pt, topsep=0pt, parsep=0pt, itemsep=1pt]
    \item We develop an \emph{explainable} and \emph{scalable} counterfactual debugging algorithm for attributing the world model transfer gap to individual time steps with a coarse-to-fine recursive scheme that exploits causal sparsity to remain tractable even for trajectories up to \textbf{1 million} steps.
    \item We develop the theoretical foundation for \emph{faithful} world-model transfer gap attribution, establishing correctness conditions under which the newly proposed counterfactual debugging method pinpoints all causally responsible steps and allocates credits proportionally to what could have been achieved if the real world dynamics at those steps had been the same as the world model's.
    \item We validate the method on Atari, ProcGen, MiniGrid \xadd{and OGBench environments with injected and also natural world model failures including} observation corruption, physics violations, and reward misprediction, showing that it correctly localizes a variety of errors and outperforms prior methods significantly in both attribution quality and sample efficiency.
\end{itemize}

\section{Background}

In this section, we introduce the theoretical framework for explaining world model behaviors from a causal perspective. We use uppercase letters to denote random variables ($X$), lowercase for their realizations ($x$) and bold letters for sets ($\mathbf{V}$) throughout.

\paragraph{Structural Causal Models and Markov Decision Processes.} We follow the semantics of Structural Causal Models (SCMs) to ground the discussion. SCMs provide the mathematical foundation for counterfactual reasoning needed to isolate causal contributions \citep{pearl2009causality,caibook2026elias}.
An SCM $\mathcal{M}$ is a tuple $\langle \mathbf{V}, \mathbf{U}, \mathcal{F}, P(\mathbf{U}) \rangle$ where $\mathbf{V}$ represents the endogenous variables, $\mathbf{U}$ represents the exogenous variables, $\mathcal{F} = \{f_i\}$ are causal mechanisms $V_i := f_i(\mathbf{Pa}_i, \mathbf{U}_i)$, $\mathbf{Pa}_i \subseteq \mathbf{V}$ are the endogenous parents of $V_i$, and $P(\mathbf{U})$ represents the exogenous distribution.
The counterfactual variable $V_{\mathbf{x}}(\mathbf{u})$ is the value of $V$ when $\mathbf{X}$ is set to $\mathbf{x}$ for $\mathbf{U} = \mathbf{u}$.

In the standard reinforcement learning literature~\citep{rlbook}, the environment is usually modeled as a Markov Decision Process (MDP), which is a tuple $\mathcal{M} = \langle \mathcal{S}, \mathcal{A}, \mathcal{P}, \mathcal{R}, \gamma \rangle$, where $\mathcal{S}$ is the state space, $\mathcal{A}$ is the action space, $\mathcal{P}$ is the transition probability, $\mathcal{R}$ is the reward function, and $\gamma$ is the discount factor. To examine MDP-based world models from a causal perspective, we cast MDPs as SCMs. \footnote{For more details, see Sec. 8 in \citet{caibook2026elias}.}

\begin{definition}[MDP-SCM]
\label{def:mdp-scm}
An MDP-SCM is an SCM with endogenous variables $\mathbf{V} = \{S_t, X_t, Y_t\}_{t=1}^T \cup \{Y\}$ (states, actions, per-step rewards, and outcome $Y = f_Y(Y_{1:T})$), mutually independent exogenous variables $\mathbf{U} = \{U_{S_t}, U_{X_t}, U_{Y_t}\}_{t=1}^T$, and structural equations defining system dynamics:
$
    S_1 = f_S(U_{S_1}), S_{t+1} = f_S(S_t, X_t, U_{S_{t+1}}),
    X_t = \pi(S_t, U_{X_t}),\ Y_t = r(S_t, X_t, U_{Y_t}) 
$.
\end{definition} 
We use $P(S)$ to denote the initial state distribution induced by $U_{S_1}$ and the discounted total reward as the outcome $Y = \sum_t \gamma^{t-1} Y_t$. Value functions $V(s)$, Q-values $Q(s,x)$, and advantages $A(s,x)$ under MDP-SCM follow the standard RL definitions~\cite{rlbook}.

\paragraph{Counterfactual Shapley Values.} Shapley values~\cite{shapley1953} distribute credit among players in cooperative games. 
Given players $\mathbf{X}$ and value function $f: 2^\mathbf{X} \to \mathbb{R}$, a \emph{coalition} $\mathbf{Z} \subseteq \mathbf{X}$ is a subset of players, and $f(\mathbf{Z})$ measures their joint contribution.
The Shapley value $\phi_t$ quantifies player $X_t$'s fair share of the total value $f(\mathbf{X}) - f(\emptyset)$ with desirable properties~\citep{shapley1953}, including \emph{efficiency}: $\sum_t \phi_t = f(\mathbf{X}) - f(\emptyset)$. A commonly used representation for efficient Shapley value computation is the kernel formulation.
Let $\mathbf{z} \in \{0,1\}^T$ denote the coalition mask ($z_t = 1$ iff $X_t \in \mathbf{Z}$), the Shapley value $\phi_t$ is written as:
\begin{equation}
    \phi_t = \sum_{\mathbf{z} \in \{0,1\}^T} \kappa_t(\mathbf{z}) \cdot f(\mathbf{z}),
    \label{eqn:kernel}
\end{equation}
where $|\mathbf{z}| = \sum_t z_t$ and 
$\kappa_t(\mathbf{z}) = \frac{1}{T\binom{T-1}{|\mathbf{z}|-1}}$ if $ z_t = 1$ while 
$\kappa_t(\mathbf{z}) = -\frac{1}{T\binom{T-1}{|\mathbf{z}|}}$ if $ z_t = 0$.

For credit assignment, \citet{li2026credit} proposes to treat actions $\mathbf{X} = \{X_1, \ldots, X_T\}$ as players, and given trajectory $\tau = (s_{1:T}, x_{1:T}, y_{1:T})$ (the \emph{evidence} for counterfactual reasoning), one can calculate Counterfactual Shapley values ($\phi$-values) to quantify how much $X_t$ contributed to outcome $Y$. They use counterfactual natural total effect (NTE, \cite{lee2025}) to measure a set of actions' causal contributions to the outcome, which is the expected difference between the observed outcome and a counterfactual outcome where a subset of actions are replaced.
And it's defined as,
\begin{align}
    \label{eq:ctf nte}
    f(\mathbf{Z}) &= \mathrm{NTE}(\mathbf{Z}, Y | \tau) 
    = \mathbb{E}_{\substack{\mathbf{u'} \sim P(\mathbf{U}),  \mathbf{u} \sim P(\mathbf{U} \mid \tau)}}\bigl[Y(\mathbf{u}) - Y_{\mathbf{Z}(\mathbf{u}')}(\mathbf{u})\bigr],
\end{align}
where $Y(\mathbf{u})$ is the observed outcome and $Y_{\mathbf{Z}(\mathbf{u}')}(\mathbf{u})$ is the counterfactual with actions within the coalition $\mathbf{Z}$ replaced by values sampled from a baseline policy $\pi_{\mathrm{base}}$ under $\mathbf{u}'$, while exogenous $\mathbf{u}$ for other variables is fixed.
In our world model attribution setting, time steps can be thought of as the players and the total credit is the return gap of deploying the policy in the world model and the real world which is defined formally in the next section (\Cref{def:transfer-gap}).

\section{Counterfactual Attribution for World Model Transfer Gap}
\label{sec:debugwm}
A visual discrepancy between the predicted and real world transitions does not always cause performance degradation, as illustrated in \cref{fig:teaser-cumreward}. In this section, we define such performance degradation as the world model transfer gap formally and introduce our causal attribution method for explaining it. Proof,  definitions and assumptions details are provided in \Cref{sec:theory-details}. 

\subsection{The World Model Transfer Gap from A Causal Lens}
\label{sec:explain-performance-degradation}
We call the performance difference resulting from deploying a policy trained on the world model in the real world environment the "world model transfer gap" which we define formally as follows. 

\begin{definition}[World Model Transfer Gap]
\label{def:transfer-gap}
Let MDP-SCM $\mathcal{M}_{\mathrm{wm}}$ and $\mathcal{M}_{\mathrm{env}}$ denote the world-model and real world whose transitions and rewards may differ, $f_S^{\mathrm{wm}} \neq f_S^{\mathrm{env}}$ and $r^{\mathrm{wm}} \neq r^{\mathrm{env}}$, while other components are the same. Let $\pi^*_{\mathrm{wm}}$ be an optimal policy under $\mathcal{M}_{\mathrm{wm}}$ and trajectory $\tau \sim \mathcal{M}^{\pi^*_{\mathrm{wm}}}_{\mathrm{env}}$ be generated under the real world $\mathcal{M}_{\mathrm{env}}$ following the same policy. The \emph{world model transfer gap} is,
\begin{align}
    \Delta(\mathcal{M}_{\mathrm{env}}, \mathcal{M}_{\mathrm{wm}}) \;:=\; Y^{\tau}_{\mathcal{M}_{\mathrm{env}}} - \mathbb{E}_{\mathcal{M}_{\mathrm{wm}}}^{\pi^*_{\mathrm{wm}}}[Y],
\end{align}
\end{definition}
We are interested in identifying the time steps at which the transitions or rewards are causally responsible for the transfer gap between its observed outcome in the real world, $Y^{\tau}_{\mathcal{M}_{\mathrm{env}}}$, and the expected outcome under the world model, $\mathbb{E}_{\mathcal{M}_{\mathrm{wm}}}^{\pi^*_{\mathrm{wm}}}[Y]$. In the environment underlying \cref{fig:teaser-cumreward}, the reward function $r$ is the same but transition function $f_S$ in the real world occasionally occludes the screen, causing the agent to miss the ball and lose points. The blackouts occur in 15 steps but only two out of the fifteen steps, $536$ and $1311$, are causally responsible for the gap (\Cref{fig:atari-phi-trace}).
Formally, we cast this as an attribution problem for the world model transfer gap.

\begin{definition}[Attribution for World Model Transfer Gap, Informal]
\label{def:causal world-model attribution}
A world-model transfer gap attribution function $\phi$ maps a trajectory, a sequence of dynamics selectors, a world model, and a policy to a real vector denoting each step's attribution to the transfer gap,
\begin{align}
    \phi: \tau \times \{{\mathrm{wm}}, {\mathrm{env}}\}^T \times \mathcal{M}_{\mathrm{wm}} \times \Pi \times [T] \mapsto \mathbb{R}^T
\end{align}
where $\{{\mathrm{wm}}, {\mathrm{env}}\}^T$ is the sequence of dynamics selectors determining whether at time step t, $\{f_S, r\}$ is from $\mathcal{M}_{\mathrm{wm}}$ or $\mathcal{M}_{\mathrm{env}}$.
The attribution function $\phi$ is causal if it satisfies the following properties,
\begin{enumerate}[label={$\mathbf{D}_\arabic*$}, leftmargin=20pt, topsep=0pt, parsep=0pt, itemsep=1pt]
    \item \textbf{Causal Admissibility}: Steps that do not cause the performance gap have zero credit;
    \item \textbf{Causal Power}: Steps that cause the performance gap are assigned non-zero credits;
    \item \textbf{Causal Normality}: Steps that are more likely to affect the outcome have more credits;
    \item \textbf{Causal Effect Scaling}: Steps affecting the outcome more are assigned more credits.
\end{enumerate}
\end{definition}

\xadd{The selector sequence indexes the hybrid configurations between the fixed real trajectory and world-model rollouts, so the mismatch between $\mathcal{M}_{\mathrm{wm}}$ and $\mathcal{M}_{\mathrm{env}}$ enters only through which steps run which mechanism.}
A formal version of the causal attribution desiderata is presented in~\Cref{sec:phi-values-causal}.
These desiderata encode natural requirements for a causal attribution of the world model transfer gap. $\mathbf{D_1,D_2}$ together guarantee that the attribution generates both sufficient and necessary allocations: steps receive credit if and only if they cause the transfer gap. $\mathbf{D_3,D_4}$ require that the scale of the allocated credits reflects each step's relative contribution: using the dynamics from the world model instead of the real world at a time step that could mitigate the transfer gap more receives proportionally more credit. 

We adapt the Counterfactual Shapley value framework to solve this. The key idea is to treat each step's dynamics selector as a player in a cooperative game and measure its contributions to the gap. Specifically, given a trajectory $\tau$ sampled from the real world under policy $\pi^*_{\mathrm{wm}}$ (denoted as $\pi$ for simplicity), a subset of steps' causal contribution is measured by NTE,
\begin{align}
    \label{eq:ctf nte wm}
    f(\mathbf{Z}) &= \mathrm{NTE}(\mathbf{Z}, Y | \tau) 
    = \mathbb{E}_{\substack{\mathbf{u'} \sim P(\mathbf{U}),  \mathbf{u} \sim P(\mathbf{U} \mid \tau)}}\bigl[Y^{\pi}_{\mathcal{M}_{\mathrm{env}}}(\mathbf{u}) - Y^{\pi}_{\mathbf{Z}(\mathcal{M}_{\mathrm{wm}}, \mathbf{u}')}(\mathbf{u})\bigr],
\end{align}
where $Y^{\pi}_{\mathcal{M}_{\mathrm{env}}}(\mathbf{u})$ is the outcome under the real world $\mathcal{M}_{\mathrm{env}}$ following policy $\pi$ and $Y^{\pi}_{\mathbf{Z}(\mathcal{M}_{\mathrm{wm}}, \mathbf{u}')}(\mathbf{u})$ is the counterfactual outcome with transitions/reward functions replaced by the world model $\mathcal{M}_{\mathrm{wm}}$ at coalition steps $t, z_t = 1, z_t \in \mathbf{Z}$ with exogenous variables $\mathbf{u'}$ \xadd{while the remaining steps keep the transitions observed in $\tau$ for as long as the counterfactual rollout stays on $\tau$; once it leaves $\tau$, the later steps are generated by $\mathcal{M}_{\mathrm{wm}}$ as well (\cref{eqn:sigma-z}).}
This measures the difference between the return of a policy rolled out under the real world and that of when a subset of steps' dynamics are replaced by the world model. Unlike statistical or visual prediction error, this captures the \emph{downstream causal effect} of each step's dynamics on the transfer gap: what return could have been achieved if the real world dynamics had been the same as the world model's.

Inserting \cref{eq:ctf nte wm} into the counterfactual Shapley values ($\phi$-values, \citep{li2026credit}), we obtain a causal attribution method for explaining the world-model transfer gap.

\begin{theorem}
\label{thm:phi is causal}
    $\phi$-values satisfy the world model transfer gap causal attribution criteria.
\end{theorem}

\xadd{In $\phi$-values, the coalitions interpolate between the two sides of \Cref{def:transfer-gap}: 
the empty coalition reproduces the observed real rollout and the full coalition the world-model rollout, 
so by Shapley efficiency, each $\phi_t$ represents a share of the transfer gap on the task 
at hand.} 

\xadd{A related concept that may cause confusion is eligibility traces~\citep{sutton1988temporaldiff,schulmanHighDimensionalContinuousControl2016a}. 
It instead scores a step by $\sum_{t' \geq t} (\gamma\lambda)^{t'-t}\delta_{t'}$, a fixed recency 
kernel over the TD errors $\delta_{t'}$ of the factual trajectory in a single environment. 
This is not a causal attribution of the transfer gap as it only reflects value learning error 
of the learning algorithm.
In \cref{fig:atari-blackout}, the blackout at $t = 536$ causes the score loss at $t = 600$: 
a trace peaks near $t = 600$ and weighs the responsible step by $(\gamma\lambda)^{64}$, 
set by its distance from the loss rather than by its contribution to it. 
Traces thus mark steps temporally associated with the gap, whereas $\phi_t$ asks 
how the return would change had step $t$ run the mechanism of $\mathcal{M}_{\mathrm{wm}}$ 
instead of $\mathcal{M}_{\mathrm{env}}$.}

To compute $\phi$-values, the NTE need to be evaluated under a given coalition $\mathbf{z}$.
The NTE defined in~\cref{eq:ctf nte wm} averages over both posterior and prior exogenous variables which can be estimated by the \emph{conditional} causal contribution given trajectory $\tau$:
$
    f(\mathbf{z}) = Y^\tau_{\mathcal{M}_{\mathrm{env}}} - Y_{\sigma_{\mathbf{z}}},
$
where $Y^\tau_{\mathcal{M}_{\mathrm{env}}}$ is the observed return and $Y_{\sigma_{\mathbf{z}}}$ is the counterfactual return under intervention $\sigma_{\mathbf{z}}$
which specifies step dynamics (action, reward, next state) under coalition $\mathbf{z} \in \{0,1\}^T$:
\begin{align}
    \sigma_{\mathbf{z}}(t) = \begin{cases}
        x^{\mathbf{z}}_t = x_t, r^{\mathbf{z}}_t = r_t, s^{\mathbf{z}}_{t+1} = s_{t+1} & z_t = 0,\ s^{\mathbf{z}}_t = s_t \\
        x^{\mathbf{z}}_t \sim \pi, r^{\mathbf{z}}_t \sim r, s^{\mathbf{z}}_{t+1} \sim f_S, \text{ s.t. } r, f_S \in \mathcal{M}_{\mathrm{wm}} & \text{otherwise}
    \end{cases}
    \label{eqn:sigma-z}
\end{align}
where $s^{\mathbf{z}}_t, r^{\mathbf{z}}_t, s^{\mathbf{z}}_{t+1}$ are the counterfactual state, reward, and next state at time $t$. $s_t, x_t, r_t, s_{t+1}$ are the observed state, action, reward, and next state at time $t$ from trajectory $\tau$.
\Cref{alg:ctf-sim} implements \Cref{eqn:sigma-z}.

While theoretically appealing, $\phi$-values are very challenging to compute in general due to the combinatorial number of coalitions for exact evaluation and sampling variance for approximation.
In the next section, we develop an efficient estimation procedure exploiting the sparsity of causal steps, making $\phi$-values tractable even for extremely long trajectories up to \textbf{1 million} steps. \footnote{Even SOTA Genie model only runs at 24 FPS, retaining consistency for a few minutes, roughly $14$K steps~\citep{bruce2024genie}.} 

\subsection{Coarse-to-Fine Counterfactual Debugging}
\label{sec:computing}

Na\"ive $\phi$-value estimation treats each of the $T$ time steps as a separate player, requiring $O(2^T)$ coalition evaluations. Even with coalition sampling, accurate estimation is still computationally expensive~\citep{lee2025,li2026credit}. When causal responsibility is sparse: only a few time steps actually drive the transfer gap, most of this computation is wasted on steps of the trajectory that contribute negligibly.

We exploit this sparsity with a \emph{coarse-to-fine} recursive scheme (\Cref{alg:recursive-shapley-main}, \textsc{Recursive}-$\phi$). The trajectory is initially partitioned into $B$ contiguous blocks, each treated as a single composite player (line 3). $\phi$-values are estimated over this reduced $B$-player game by \Cref{eqn:kernel} and \Cref{eq:ctf nte wm} (line 5). Blocks whose per-step attribution $|\Phi_j|/|\mathcal{B}_j|$ falls below a pruning threshold $\epsilon$ are declared inactive and excluded from further computation (line 4, 6-7). The remaining active blocks are subdivided by a factor of $B$ and the process repeats, progressively refining resolution only where the block-level attribution is non-negligible.

\begin{example}[\textsc{Recursive}-$\phi$ in Atari Pong.]
    In \cref{fig:teaser-cumreward} example, the buggy Pong trajectory (rollout) spans $T \approx 1860$ environment steps with two causal blackouts (the occlusions that account for the $21\!\to\!15$ return drop) alongside several non-causal blackouts. With a split factor $B = 2$, round $0$ partitions the trajectory into two halves and estimates each block's attribution; any block containing only non-causal steps yields negligible attribution and is pruned, because intervening on a non-causal blackout leaves the two causal blackouts in place and the return unchanged. Each subsequent round halves whatever blocks still carry signal, recursively pruning sub-blocks of non-causal steps at every resolution. The algorithm isolates the two causal steps after $\lceil \log_2 1860 \rceil = 11$ rounds, spending a total budget that scales with the number of causal steps $K = 2$ rather than the length $T$.
\end{example}

\paragraph{Design choices.}
We make the following design choices for \textsc{Recursive}-$\phi$.

\textit{Block granularity and split factor.}
The split factor $B$ controls the resolution doubling rate. With $B = 2$ (binary splitting), the algorithm runs for at most $\lceil\log_2 T\rceil$ rounds. Each round re-estimates $\phi$-values only over the active blocks, pruning inactive blocks, so the effective number of players $n$ is bounded by the split size $B$ times the number of active blocks remaining. When $n$ is small enough that $2^n \leq M$, the algorithm switches from sampling to exact enumeration under counterfactual simulation rollout budget $M$, eliminating sampling variance entirely for this round.

\textit{Pruning threshold.}
The threshold $\epsilon$ governs the sparsity--accuracy trade-off. A block is pruned when its per-step attribution magnitude is below $\epsilon$, which sets the attribution of all its constituent time steps to zero. In the sparse-causality regime, where only $K \ll T$ time steps have non-negligible causal effect, aggressive pruning reduces the effective problem size from $T$ to $O(K)$ players within a few rounds, yielding a total cost of $O(2^K \log_B T)$ counterfactual simulations rather than $O(2^T)$.

\textit{Coalition sampling within each round.}
At each resolution level, if the total number of possible coalitions is less than the coalition budget $M$, we enumerate all coalitions to calculate $\phi$-values exactly. Otherwise, we use the optimal proposal distribution $Q^*_n$~\citep{li2026credit} to sample coalitions over the $n$ active blocks. Either way, each coalition $\mathbf{z} \in \{0,1\}^n$ is expanded to a full $T$-length mask by broadcasting: all time steps within an active block receive the same intervention indicator. 

\begin{algorithm}[t]
    \caption{\textsc{Recursive}-$\phi$}
    \label{alg:recursive-shapley-main}
    \begin{algorithmic}[1]
    \REQUIRE Trajectory $\tau$ (length $T$), policies $\pi$, world model $\mathcal{M}_{\mathrm{wm}}$, budget $M$, split $B$, threshold $\epsilon$
    \ENSURE Per-step attributions $\hat{\phi}_{1:T}$
    \STATE Initialize $\hat{\phi}_{1:T} \gets \mathbf{0}$, all steps active
    \FOR{$r = 0, 1, \ldots, \lceil\log_B T\rceil - 1$}
        \STATE Partition $\{1,\ldots,T\}$ into $\min(B^{r+1}, T)$ contiguous blocks
        \STATE Identify active blocks $\mathcal{A}$ (those containing at least one active step)
        \STATE Estimate block-level $\phi$-values $\{\Phi_j\}_{j \in \mathcal{A}}$ by \Cref{eqn:kernel} and \Cref{eqn:sigma-z}
        \STATE Distribute: $\hat{\phi}_t \gets \Phi_j / |\mathcal{B}_j|$ for $t \in \mathcal{B}_j$, $j \in \mathcal{A}$
        \STATE Prune: deactivate steps in block $j$ if $|\Phi_j|/|\mathcal{B}_j| < \epsilon$
    \ENDFOR
    \RETURN $\hat{\phi}_{1:T}$
    \end{algorithmic}
\end{algorithm}

\paragraph{Correctness of Coarse-to-Fine Estimation.}
The coarse-to-fine scheme is sound whenever the causal steps are sparse and do not cancel out other's effect on the transfer gap when grouped into a block. Concretely, we define a sparse transfer gap attribution problem where at most $K$ steps carry a meaningful signal of magnitude $\geq \delta$, and the rest are exactly zero (\Cref{def:sparse-attribution}). Two conditions then suffice to show the correctness of the algorithm. (i) \emph{Bounded interaction} (\Cref{assum:bounded-interaction-effect}): the block-level $\phi$-values $\Phi_j$ deviates from the sum of its per-step values by at most $\eta$. This holds with $\eta\!\ll\!\delta$ in our setting, because non-causal steps contribute zero to the transfer gap in every coalition, effectively killing all interaction terms that involve them~\citep{shapleyinteraction2023}. (ii) \emph{Sign consistency} (\Cref{assum:sign-consistency}): causal steps share a sign, which is natural because systematic world-model errors compound along a rollout rather than cancel~\citep{jannerWhenTrustYour2019a}. Under these conditions, the algorithm provably finds what it should:

\begin{theorem}[Correctness of \Cref{alg:recursive-shapley-main}, informal]
\label{thm:recursive-correctness-informal}
Under the conditions above, with a small enough threshold $\epsilon$, with high probability, \Cref{alg:recursive-shapley-main} (a) loses no causal step, (b) pinpoints all $K$ of them in $\log_B T$ rounds, and (c) recovers their attributions up to bounded errors while zeroing out the rest.
\end{theorem}

In words: as long as the causally responsible dynamics errors are few, sizeable, and compound with time, the recursive scheme zooms in on them in logarithmically many rounds without losing any along the way, and shrinks to exact enumeration once the surviving sub-game is small. 

\section{Experiments}
\label{sec:experiments}

We evaluate \textsc{Recursive}-$\phi$ on Atari~\citep{bellemareArcadeLearningEnvironment2013}, ProcGen~\citep{cobbeLeveragingProceduralGeneration2020}, MiniGrid~\citep{gym_minigrid} \xadd{and OGBench~\citep{park2025ogbench} spanning three common sources of a transfer gap: observation corruption, reward misspecification, and dynamics shift. The experiments answer} three questions:
\textbf{(Q1)}~can \textsc{Recursive}-$\phi$ localize the time steps which are causally responsible for the transfer gap;
\textbf{(Q2)}~does \textsc{Recursive}-$\phi$ explain more transfer gap than other baselines;
\textbf{(Q3)}~does \textsc{Recursive}-$\phi$ scale to long trajectories with limited number of simulations (rollouts) when baselines can not? All results are averaged over $10$ sampling seeds with \xadd{the observed trajectory's seed} held fixed for controlled world-model failures.

\begin{figure}[ht]
    \centering
    \includegraphics[width=.85\linewidth]{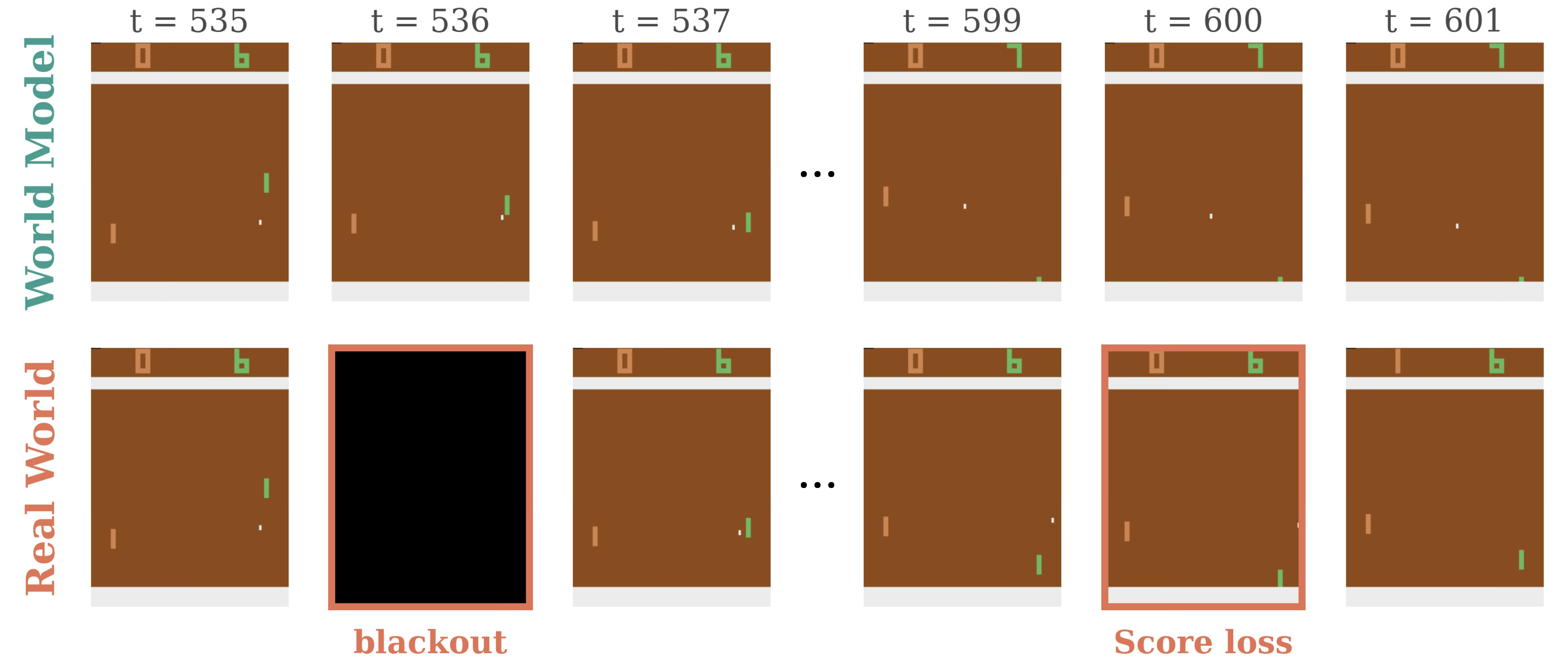}
    \caption{\textbf{Observation blackout on Pong.} Top row: the world model $\mathcal{M}_{\mathrm{wm}}$ observations. Bottom row: real world $\mathcal{M}_{\mathrm{env}}$ has blackouts at $t = 536$ but score loss does not surface until $t = 600$.}
    \label{fig:atari-blackout}
    \vspace{-8pt}
\end{figure}

\begin{figure}[ht]
    \centering
    \includegraphics[width=\linewidth]{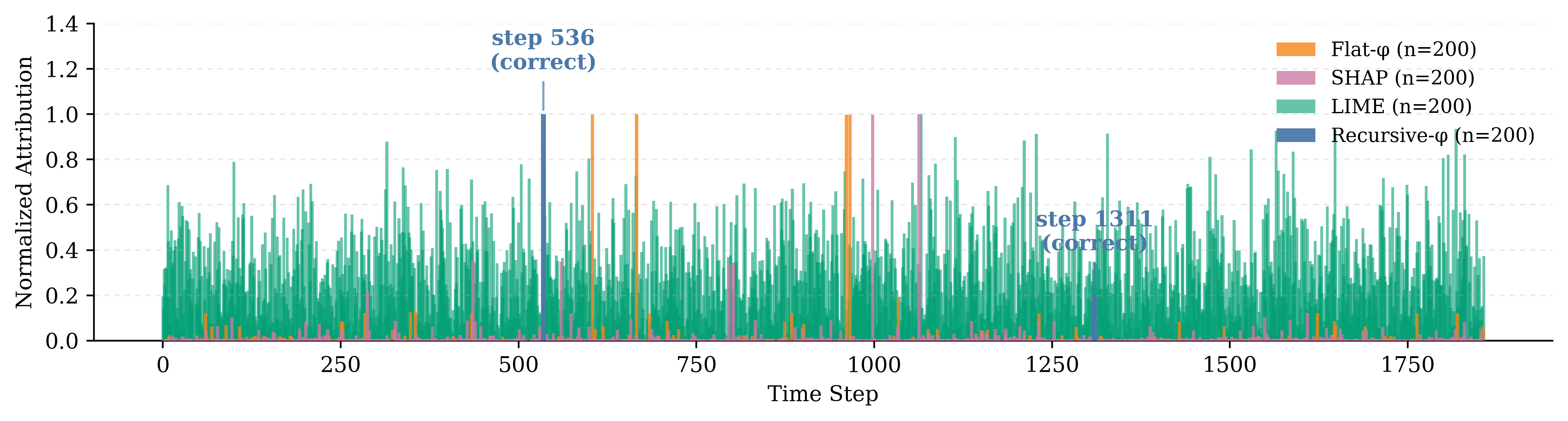}
    \caption{\textbf{Localization of causally important steps.} Only \textsc{Recursive}-$\phi$ (blue) correctly captures the two causally important steps $536$ and $1311$.}
    \label{fig:atari-phi-trace}
    \vspace{-12pt}
\end{figure}

\begin{table}[ht]
    \centering
    \small
    \caption{\textbf{Attribution quality and sample efficiency on Atari Pong.} Results are aggregated over $10$ seeds; ``-'' marks cells where not enough seeds succeed within the rollout budget (5k) to derive meaningful statistics.}
    \label{tab:atari-results}
    \vspace{2pt}
    \begin{tabular}{lccccc}
        \toprule
        Trajectory length $L$ & $100$ & $300$ & $500$ & $1000$ & $3000$ \\
        \midrule
        \multicolumn{6}{l}{\emph{transfer gap explained (\%, $\uparrow$)}} \\
        \quad \textsc{Recursive}-$\phi$ (ours) & $\mathbf{100 \pm 0}$ & $\mathbf{100 \pm 0}$ & $\mathbf{100 \pm 0}$ & $\mathbf{100 \pm 0}$ & $\mathbf{100 \pm 0}$ \\
        \quad \textsc{Flat}-$\phi$             & $\mathbf{100 \pm 0}$  & $\mathbf{100 \pm 0}$  & $10 \pm 32$ & $0 \pm 0$ & $0 \pm 0$ \\
        \quad SHAP                             & \xadd{$20 \pm 42$}  & \xadd{$0 \pm 0$}  & $0 \pm 0$   & $0 \pm 0$   & $0 \pm 0$      \\
        \quad LIME                             & \xadd{$10 \pm 32$}    & \xadd{$\mathbf{100 \pm 0}$}    & \xadd{$90 \pm 32$}   & \xadd{$70 \pm 48$} & $0 \pm 0$       \\
        \midrule
        \multicolumn{6}{l}{\emph{Rollouts required for successful attribution ($\downarrow$)}} \\
        \quad \textsc{Recursive}-$\phi$ (ours) & $\mathbf{28 \pm 0}$ & $\mathbf{34 \pm 0}$ & $\mathbf{132 \pm 0}$ & $\mathbf{148 \pm 0}$ & $\mathbf{156 \pm 0}$ \\
        \quad \textsc{Flat}-$\phi$             & $750 \pm 264$  & $4200 \pm 1687$ & - & - & - \\
        \quad SHAP                             & \xadd{-} & - & - & -            & -          \\
        \quad LIME                       & - & \xadd{$120 \pm 42$} & \xadd{$1089 \pm 1600$} & \xadd{$1214 \pm 1792$} & -    \\
        \bottomrule
    \end{tabular}
    \vspace{-2pt}
\end{table}

We compare the proposed counterfactual debugging method, \textsc{Recursive}-$\phi$ against a causal baseline \textsc{Flat}-$\phi$ (\Cref{alg:flat-shapley}) which is a non-recursive na\"ive implementation of the $\phi$-value and two non-causal baselines SHAP~\citep{lundberg2017unified} and LIME~\citep{ribeiro2016should}. Details and more results are provided in \cref{sec:exp design details,sec:extended-results}.
\paragraph{Sim-to-sim protocol.}
For the ease of testing controlled errors, other than the learned world models (one diffusion based~\citep{Alonso_2024}, one LeWM~\citep{maes2026leworldmodel}) tested in \cref{sec:results-learned-wm}, 
we mainly simulate real world vs. world model discrepencies by modifying the simulator: an unchanged base simulator serves as the world model $\mathcal{M}_{\mathrm{wm}}$, and a perturbed simulator $\mathcal{M}_{\mathrm{env}}$ stands for the real world with unexpected events (perturbations). A pretrained policy $\pi$ from $\mathcal{M}_{\mathrm{wm}}$ rolls out in $\mathcal{M}_{\mathrm{env}}$. 
Specfically, we evaluate each method by if removing the top $K$ (known by design) detected steps can close the transfer gap. 
\xadd{$K$ is used only for this scoring and is never an input to \textsc{Recursive}-$\phi$. The algorithm itself needs one observed real trajectory, the policy $\pi$, and a world model that can be rolled out from the states of that trajectory: its counterfactual rollouts query $\mathcal{M}_{\mathrm{wm}}$ alone and never reset or query $\mathcal{M}_{\mathrm{env}}$.}

\subsection{Experiment 1: Observation Blackouts on Atari Pong}
\label{sec:results-obsocclusion}

We test our method on Atari Pong with a pretrained PPO agent~\citep{schulmanProximalPolicyOptimization2017}.
At randomly selected time steps the observation emitted by $\mathcal{M}_{\mathrm{env}}$ is zeroed out, simulating a real world that experiences unexpected sensor disconnections. \cref{fig:atari-blackout} illustrates the error and its delayed effect: the causally relevant blackout happens at $t = 536$ yet the observable score loss does not surface until $t = 600$. 

\paragraph{Results.}
\emph{(i) Localization (Q1).} \cref{fig:atari-phi-trace} shows the per-step $|\phi|$-value trace on the full $\sim\!1860$-step Pong trajectory at \xadd{$L = 3000$}: \textsc{Recursive}-$\phi$ concentrates all mass on the two causal blackouts at steps $536$ and $1311$ with every other step at zero. Other baselines' attribution spikes scatter across the horizon\xadd{, and none of them recovers both} causally important steps. \emph{(ii) Attribution quality (Q2).} The top half of \cref{tab:atari-results} reports the fraction of the transfer gap recovered by removing each method's top-$K$ blackouts: \textsc{Recursive}-$\phi$ explains $100\%$ at every length, while \textsc{Flat}-$\phi$ matches at $L \leq 300$ then collapses (10\% at $L = 500$, $0\%$ beyond) once the $5$k-rollout sweep can no longer separate the two causal sites from sampling noise. \xadd{SHAP recovers only on $2/10$ seeds at $L = 100$. LIME favors the blackouts that cost the most points averaged over the reset states of its factual rollouts. That is the causal one at $L = 300$, but at $L = 100$ two others cost more and at $L \in \{500, 1000\}$ two nearly tie with it, so LIME recovers there only when sampling noise ranks it first at some budget (\cref{sec:exp design details}).} \emph{(iii) Sample efficiency (Q3).} The bottom half of \cref{tab:atari-results} gives the smallest budget at which the gap is fully explained: \textsc{Recursive}-$\phi$'s cost grows mildly with $L$ and has \emph{no} spread across $10$ seeds (pruning collapses the inactive horizon and the surviving sub-game switches to exact enumeration), whereas \textsc{Flat}-$\phi$ already needs $>\!25\times$ more at $L \in \{100, 300\}$ \xadd{and times out from $L = 500$, SHAP times out at every $L$, and LIME needs $4$--$8\times$ more where it succeeds}. The gap widens monotonically with $L$.

\subsection{Experiment 2: Mechanistic Shifts on ProcGen CoinRun}
\label{sec:results-mechanism}

\begin{figure}[t]
    \centering
    \begin{subfigure}[b]{.48\linewidth}
        \centering
        \includegraphics[width=\linewidth]{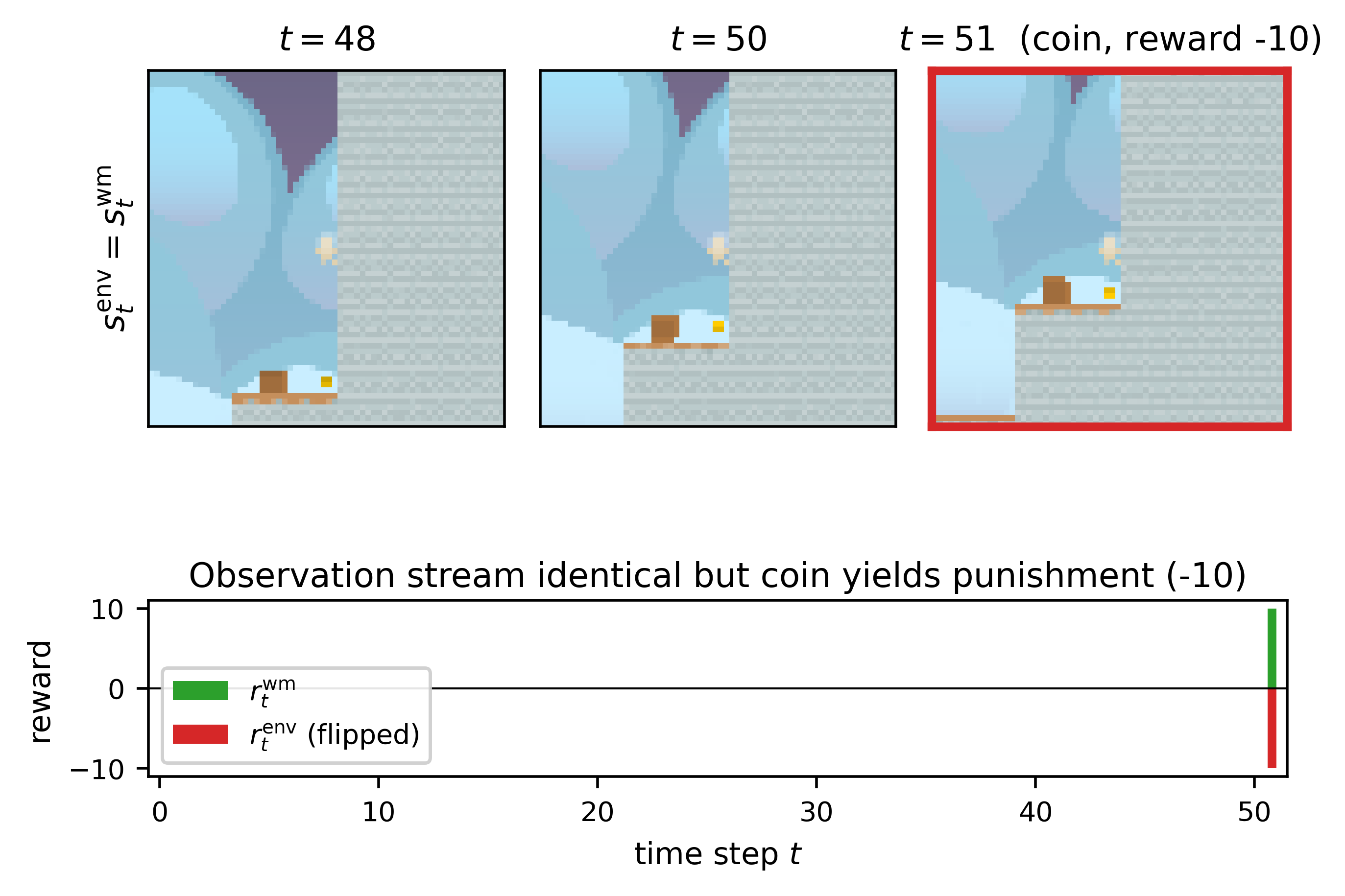}
        \caption{Variant A (poisonous coin).}
        \label{fig:coinrun-variant-a}
    \end{subfigure}
    \hfill
    \begin{subfigure}[b]{.48\linewidth}
        \centering
        \includegraphics[width=\linewidth]{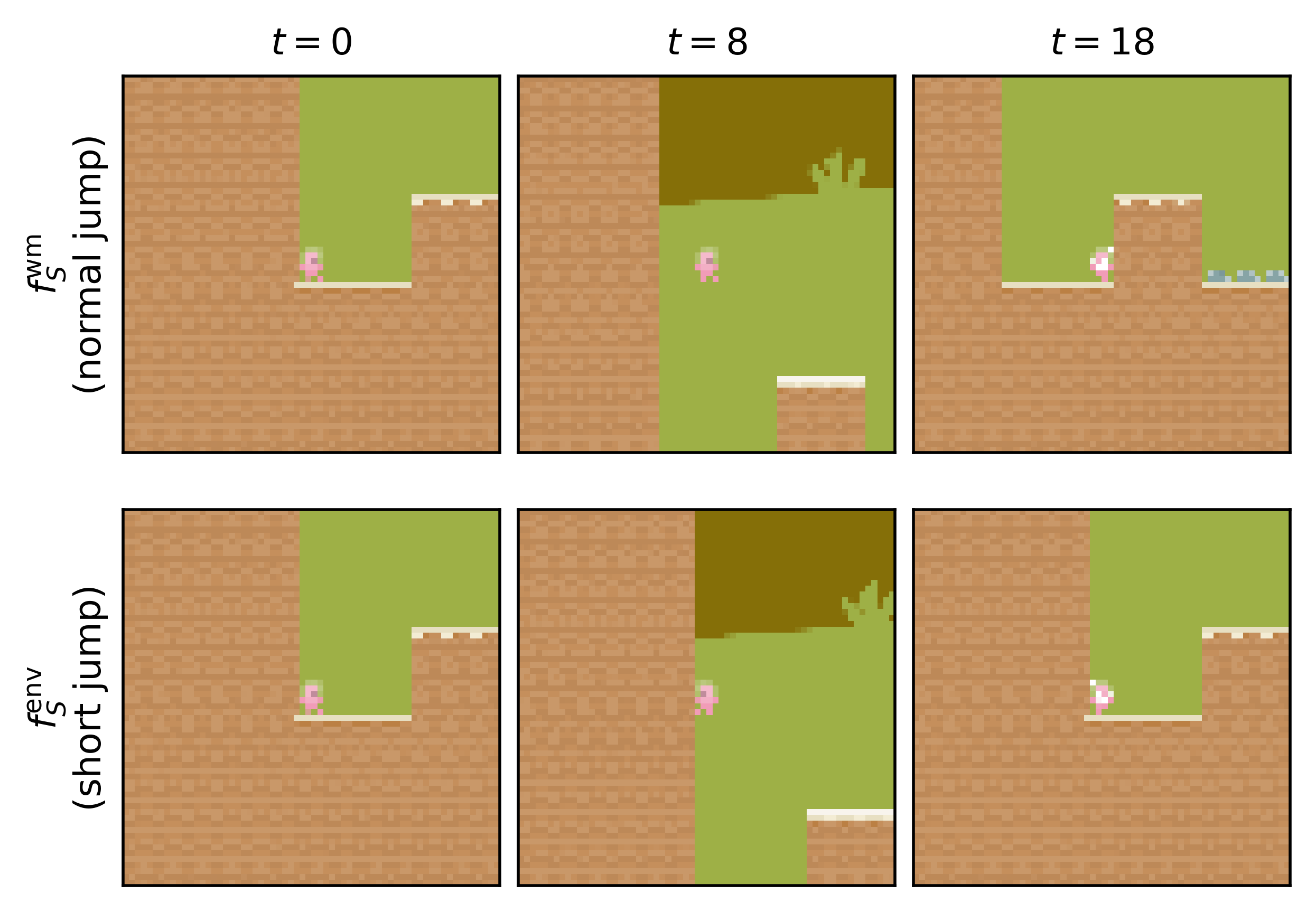}
        \caption{Variant B (windy day).}
        \label{fig:coinrun-variant-b}
    \end{subfigure}
    \caption{\textbf{CoinRun variants.} Both are visually plausible yet incur a large transfer gap.}
    \label{fig:coinrun-variants}
    \vspace{-15pt}
\end{figure}

We move from observational to mechanistic shifts on ProcGen CoinRun with a trained PPO policy under two variants with unexpected reward and transition dynamics shifts, respectively. 
\emph{Variant A (poisonous coin):} every time a coin is collected, the wrapper draws $\mathrm{Bernoulli}(p_A)$ and, on success, flips the reward sign ($-10$). The observation stream stays pixel-identical.
\emph{Variant B (windy day):} at every \texttt{UP}-containing action (jump) the wrapper draws $\mathrm{Bernoulli}(p_B)$ and, on success, applies a horizontal wind that pushes the agent to the left, so the jump falls short. \cref{fig:coinrun-variants} illustrates both.

\begin{table}[t]
    \centering
    \small
    \caption{\textbf{Attribution quality and sample efficiency on CoinRun variants.} Results are aggregated over $10$ seeds; ``-'' marks cells where not enough seeds succeed within the rollout budget (5k) to derive meaningful statistics.}
    \label{tab:coinrun-results}
    \vspace{2pt}
    \begin{tabular}{lcccc}
        \toprule
        Trajectory length $L$ & $100$ & $300$ & $500$ & $1000$ \\
        \midrule
        \multicolumn{5}{l}{\emph{transfer gap explained (\%, $\uparrow$)}} \\
        \quad Variant A: \textsc{Recursive}-$\phi$ (\textbf{ours}) & $\mathbf{100 \pm 0}$ & $\mathbf{100 \pm 0}$ & $\mathbf{100 \pm 0}$ & $\mathbf{100 \pm 0}$ \\
        \quad Variant A: \textsc{Flat}-$\phi$             & $\mathbf{100 \pm 0}$         & $\mathbf{100 \pm 0}$         & $0 \pm 0$           & $0 \pm 0$           \\
        \quad Variant A: SHAP                             & $\mathbf{100 \pm 0}$         & $\mathbf{100 \pm 0}$         & $0 \pm 0$           & $0 \pm 0$           \\
        \quad Variant A: LIME                             & $10 \pm 9$                   & $0 \pm 0$                    & $0 \pm 0$           & $0 \pm 0$           \\
        \quad Variant B: \textsc{Recursive}-$\phi$ (\textbf{ours}) & $\mathbf{100 \pm 0}$ & $\mathbf{100 \pm 0}$ & $\mathbf{100 \pm 0}$ & $\mathbf{100 \pm 0}$ \\
        \quad Variant B: \textsc{Flat}-$\phi$             & $\mathbf{100 \pm 0}$         & $90 \pm 9$                   & $40 \pm 15$         & $20 \pm 13$         \\
        \quad Variant B: SHAP                             & $\mathbf{100 \pm 0}$         & $50 \pm 16$                  & $20 \pm 13$         & $20 \pm 13$         \\
        \quad Variant B: LIME                             & $30 \pm 14$                  & $10 \pm 9$                   & $30 \pm 14$         & $0 \pm 0$           \\
        \midrule
        \multicolumn{5}{l}{\emph{Rollouts required for successful attribution ($\downarrow$)}} \\
        \quad Variant A: \textsc{Recursive}-$\phi$ (\textbf{ours}) & $\mathbf{28 \pm 0}$  & $\mathbf{34 \pm 0}$   & $\mathbf{132 \pm 0}$ & $\mathbf{136 \pm 0}$ \\
        \quad Variant A: \textsc{Flat}-$\phi$             & $620 \pm 114$        & $3800 \pm 611$        & -      & -      \\
        \quad Variant A: SHAP                             & $800 \pm 82$         & $3950 \pm 550$        & -      & -      \\
        \quad Variant A: LIME                             & -         & -       & -      & -      \\
        \quad Variant B: \textsc{Recursive}-$\phi$ (\textbf{ours}) & $\mathbf{40 \pm 0}$  & $\mathbf{60 \pm 0}$   & $\mathbf{276 \pm 0}$ & $\mathbf{76 \pm 0}$  \\
        \quad Variant B: \textsc{Flat}-$\phi$             & $810 \pm 165$        & $1711 \pm 653$        & $3125 \pm 1125$      & -      \\
        \quad Variant B: SHAP                             & $1080 \pm 271$       & $1620 \pm 863$        & -         & -      \\
        \quad Variant B: LIME                             & $467 \pm 267$        & -   & $367 \pm 133$        & -      \\
        \bottomrule
    \end{tabular}
    \vspace{-2pt}
\end{table}

\paragraph{Results.}

\emph{(i) Localization (Q1).} The per-step $|\phi|$ traces in \cref{fig:coinrun-phi-trace-sweep-A,fig:coinrun-phi-trace-sweep-B} (\cref{sec:extended-coinrun-lengths}) show \textsc{Recursive}-$\phi$ concentrating its mass on the realized causal steps, while \textsc{Flat}-$\phi$, SHAP, and LIME all rank non-causal steps above the causally important ones.
\emph{(ii) Attribution quality (Q2).} \textsc{Recursive}-$\phi$ recovers $100\%$ of the transfer gap on every $(\text{variant}, L)$ setting. \textsc{Flat}-$\phi$ and SHAP track each other closely on Variant~A; on Variant~B they degrade as $L$ grows, with SHAP collapsing faster. LIME never exceeds $30\%$ on any experiment setting and already times out at $L \geq 300$ on Variant~A.
\emph{(iii) Sample efficiency (Q3).} \textsc{Recursive}-$\phi$ uses $30$--$300$ rollouts with zero seed variance (pruning collapses the horizon to a sub-game small enough for exact enumeration), whereas \textsc{Flat}-$\phi$ and SHAP are variance-dominated and need $20$--$100\times$ more rollouts when they recover at all. LIME succeeds on at most $3/10$ seeds in any cell, requiring the full $5\text{k}$ cap on the one Variant~A setting where it recovers.

\subsection{Experiment 3: Sample Efficiency at Scale on MiniGrid}
\label{sec:results-longhorizon}

\begin{wrapfigure}{r}{0.20\linewidth}
    \vspace{-12pt}
    \centering
    \includegraphics[width=\linewidth]{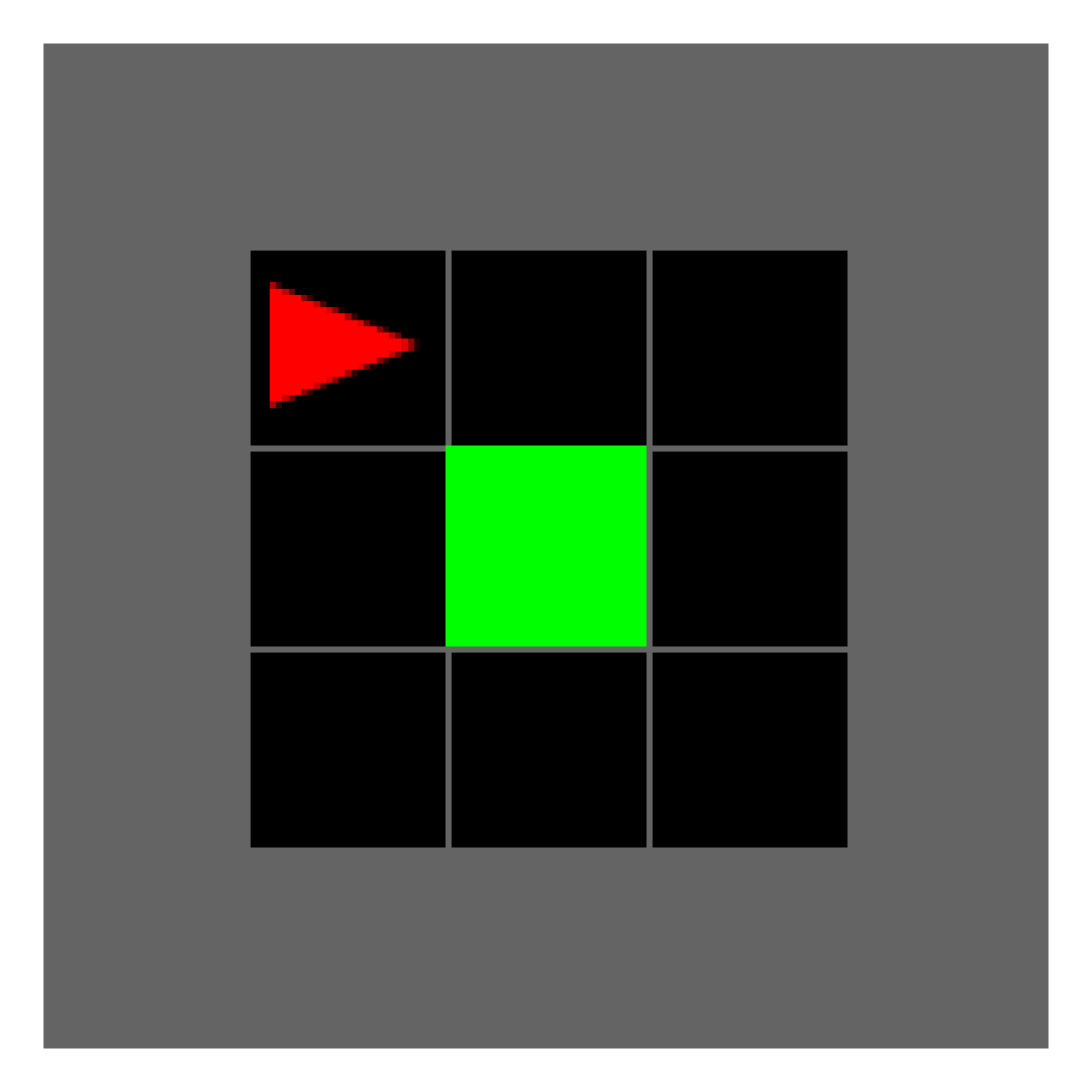}
    \caption{A custom MiniGrid world.}
    \label{fig:centergoal-env}
    \vspace{-12pt}
\end{wrapfigure}
Experiment 3 focuses on sample efficiency (\textbf{Q3}) and pushes the length ($L$) of the trajectory to be explained from thousands of steps in previous experiments to \textbf{1 million} steps. We design a custom MiniGrid \texttt{CenterGoal5x5} (\cref{fig:centergoal-env}): a $5\times5$ room walled into a walkable interior with the goal at the center. A hand-crafted policy loops the 8-cell inner ring for $n$ steps before heading to the goal; varying $n$ controls $L$. The perturbation is a \emph{recurring} reward error where a fixed penalty is subtracted at $K = 1, 2, 5$ randomly sampled time steps, and the ground-truth attribution set is exactly those steps. The deliberate simplicity (the causally important steps are known) lets us isolate \emph{how the rollout budget grows with horizon}, independent of any modelling noise.

\begin{figure}[t]
    \centering
    \begin{subfigure}[b]{0.25\linewidth}
        \centering
        \includegraphics[width=\linewidth]{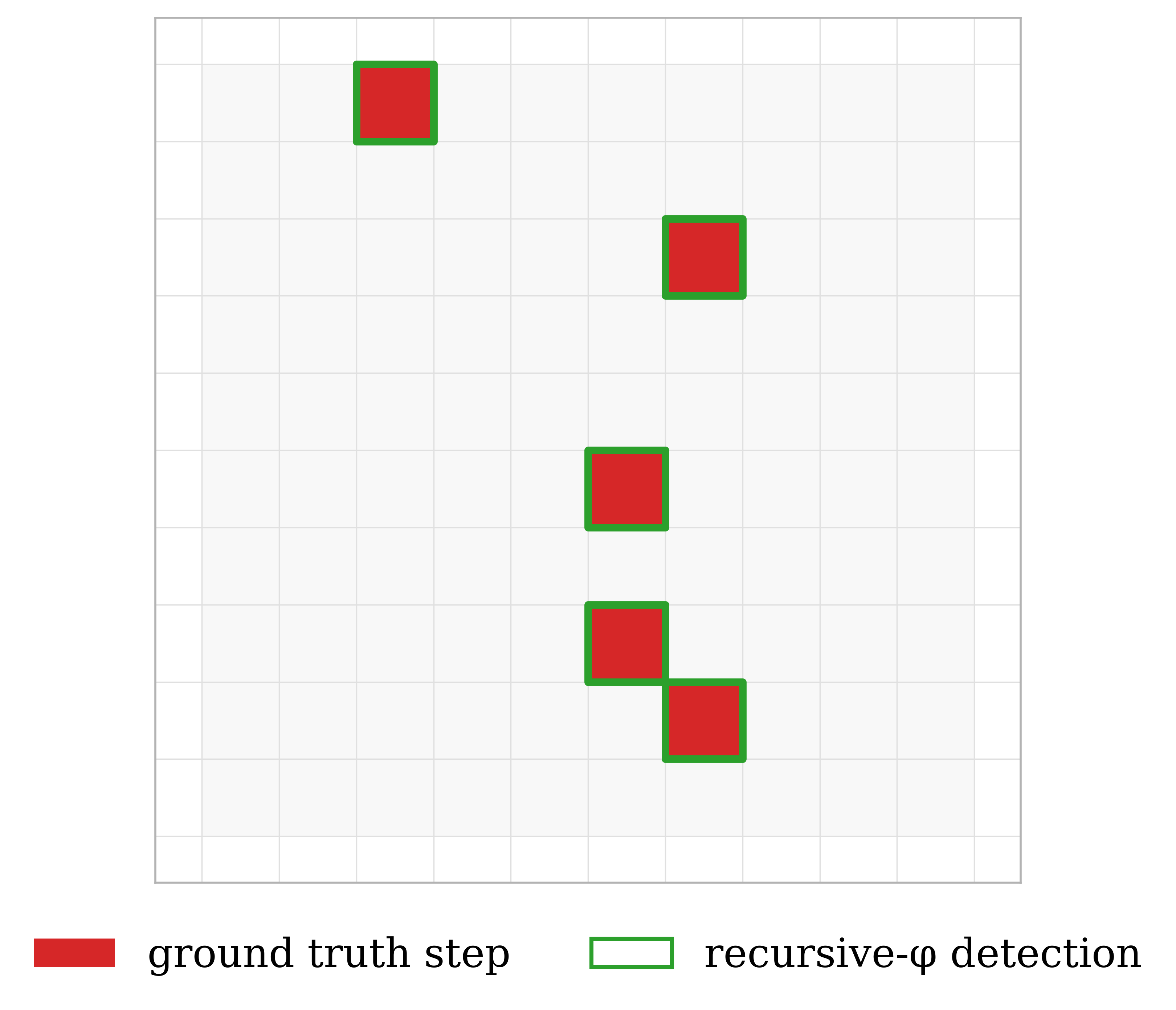}
        \caption{\xadd{Planted steps, $K = 5$.}}
        \label{fig:phi-sweep-pixelmap-100}
    \end{subfigure}
    \hfill
    \begin{subfigure}[b]{0.35\linewidth}
        \centering
        \includegraphics[width=\linewidth]{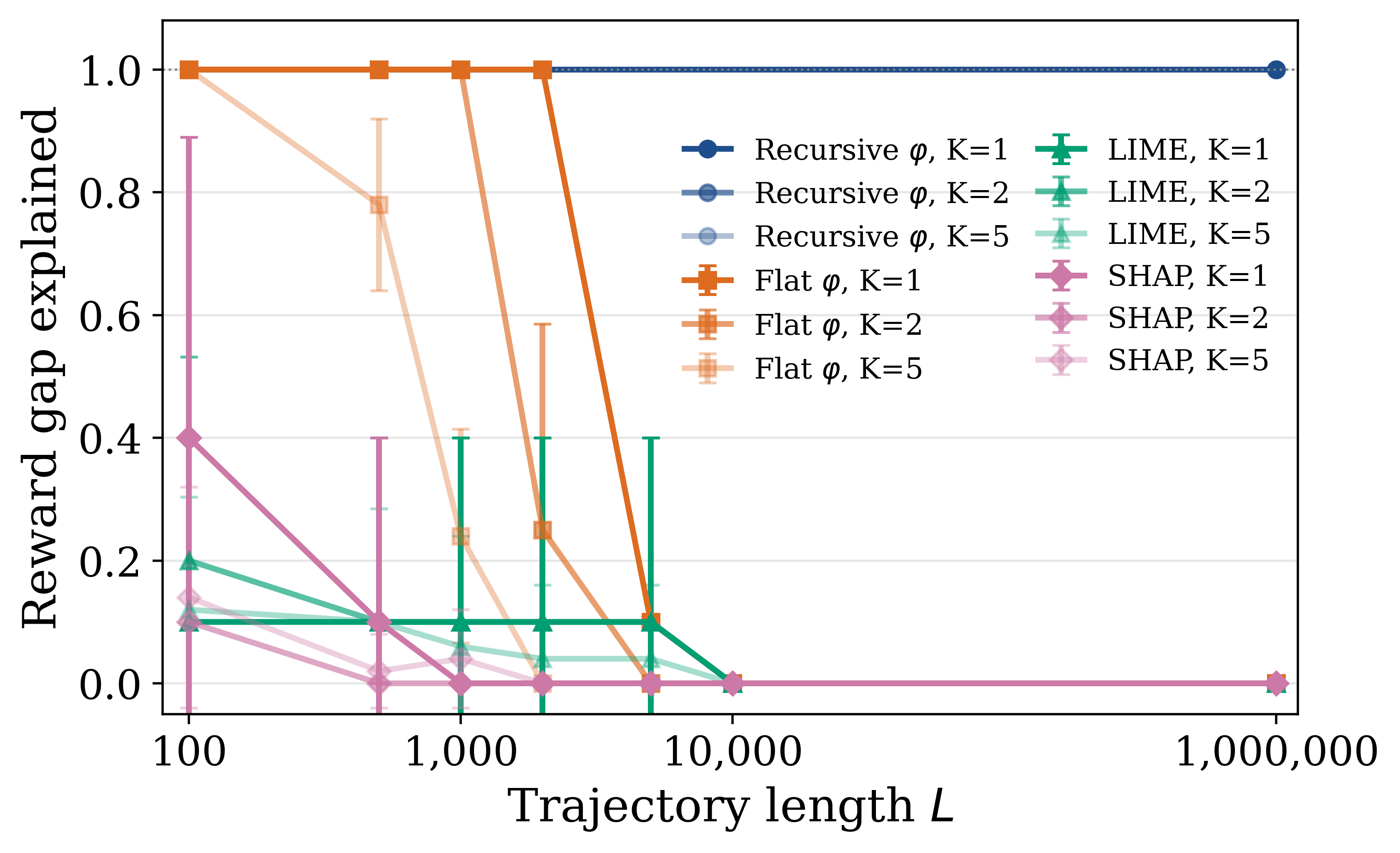}
        \caption{Transfer gap explained vs $L$.}
        \label{fig:phi-sweep-recovery}
    \end{subfigure}
    \hfill
    \begin{subfigure}[b]{0.35\linewidth}
        \centering
        \includegraphics[width=\linewidth]{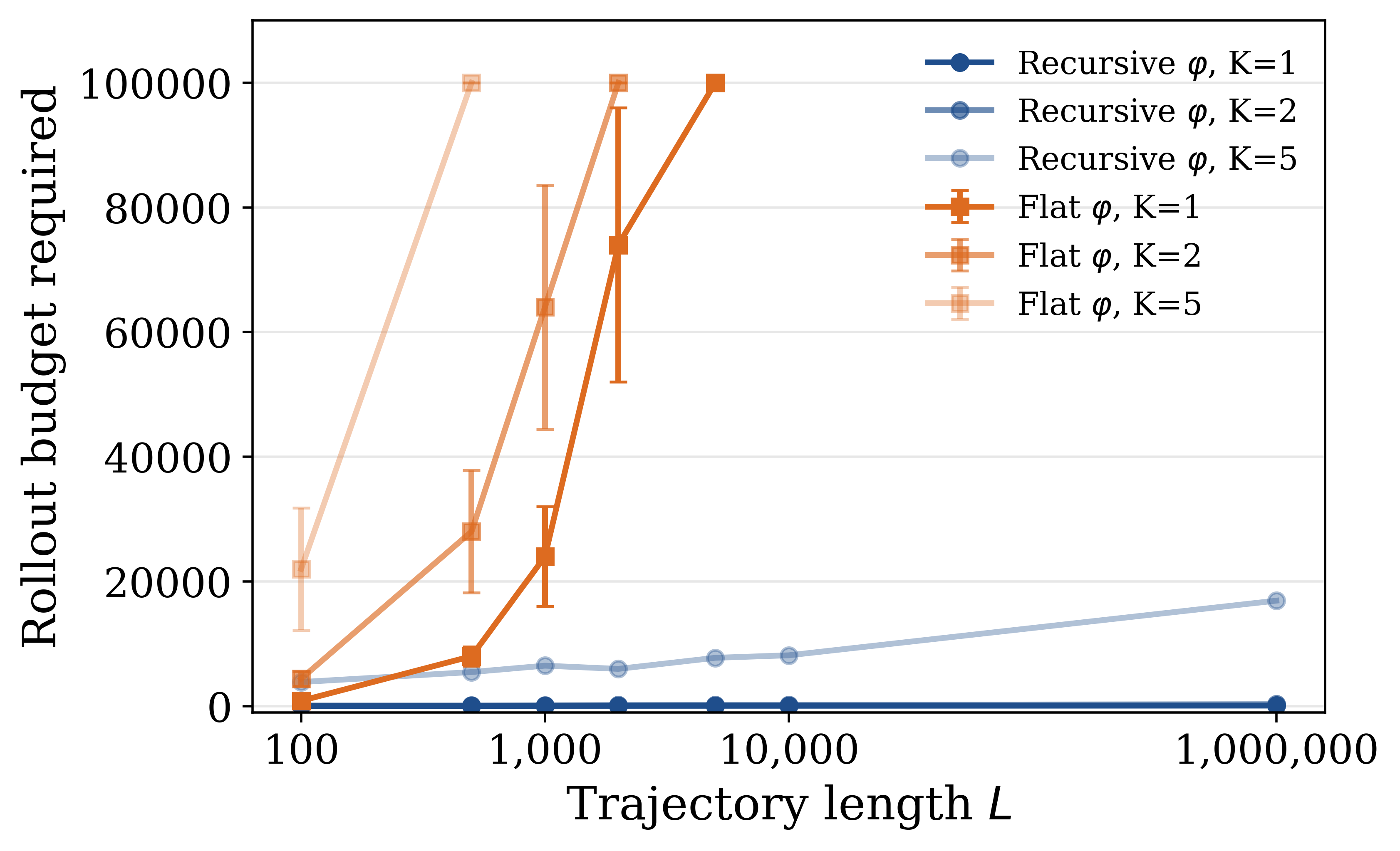}
        \caption{Rollouts required vs $L$.}
        \label{fig:phi-sweep-budget}
    \end{subfigure}
    \caption{\textbf{Scaling to $L = 1M$ on MiniGrid \texttt{CenterGoal}.} 
    }
    \label{fig:phi-sweep}
    \vspace{-15pt}
\end{figure}

\paragraph{Results (\cref{fig:phi-sweep}).}
The pixel map in \cref{fig:phi-sweep-pixelmap-100} \xadd{outlines the $K = 5$ planted steps at $L = 100$ because \textsc{Recursive}-$\phi$'s top-$K$ equals them, and for every $K \in \{1, 2, 5\}$} \textsc{Recursive}-$\phi$ places all mass exactly on the planted penalty steps with the rest at zero, confirming the accurate localization of \textsc{Recursive}-$\phi$. 
\Cref{fig:phi-sweep-recovery} shows that \textsc{Recursive}-$\phi$ recovers $100\%$ of the transfer gap at \emph{every} $L$ up to $1$M across all number of perturbations $K$, whereas \textsc{Flat}-$\phi$ matches \xadd{at every $K$} only at $L = 100$ and \xadd{at $L = 5000$ recovers on at most $1/10$ seeds} even at the largest budget we sweep. \xadd{SHAP explains at most $40\%$ of the gap (at $L = 100$, $K = 1$, where $4/10$ seeds recover) and LIME at most $20\%$, and beyond $L = 100$ neither recovers on more than $1/10$ seeds}, so we omit their budget curves from \cref{fig:phi-sweep-budget} as the means are not statistically meaningful. The decisive comparison is the rollout cost (\cref{fig:phi-sweep-budget}) against the other causal baseline, \textsc{Flat}-$\phi$, whose budget scales near exponentially with $L$ and blows past $10^{5}$ rollouts already at $L = 500$ for $K = 5$, while recursive's budget grows essentially logarithmically. This matches the $O(2^K \log_B T)$ cost of \cref{alg:recursive-shapley-main}. So a $10^{4}\times$ increase in $L$ costs less than a $5\times$ increase in rollouts budget. \textsc{Recursive}-$\phi$ therefore demonstrates the strong scaling property for trajectories up to \textbf{1M steps} that are structurally inaccessible to prior methods.

\subsection[Experiment 4: Learned World Models]{\xadd{Experiment 4: Learned World Models}}
\label{sec:results-learned-wm}

\begin{figure}[t]
    \centering
    \includegraphics[width=\linewidth]{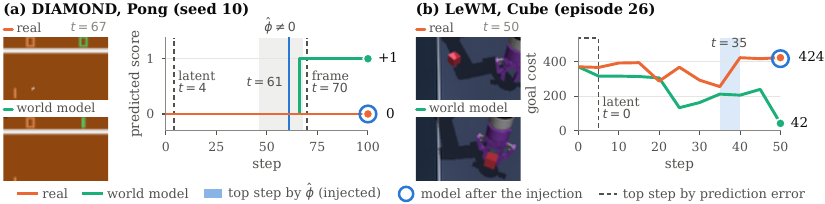}
    \caption{\xadd{\textbf{Two single-injection repairs.} (a)~DIAMOND hallucinates that the ball passes the opponent at $t = 66$ and the agent scores a point; injecting the real transition at a top-ranked step ($t = 61$) by $\hat{\phi}$ restores the real score. (b)~The cube is dropped, then LeWM hallucinates that it picks it up again; injecting step $35$, with the largest $\phi$, moves the predicted goal cost from $42$ to $418$ vs. real cost at $424$.}}
    \label{fig:learned-wm-cases}
    \vspace{-12pt}
\end{figure}

\xadd{We apply \textsc{Recursive}-$\phi$ to two pretrained neural world models, DIAMOND~\citep{Alonso_2024} (diffusion) on Pong and LeWM~\citep{maes2026leworldmodel} (JEPA) on OGBench Cube~\citep{park2025ogbench}, with the simulator as the real world. Lacking ground truth, we score an attribution by \emph{repair}: we inject real transitions at the top-ranked steps of the world-model rollout and check whether it predicts the real outcome (\cref{sec:extended-learned-wm}).
}

\begin{wraptable}{r}{0.48\linewidth}
    \vspace{-12pt}
    \centering
    \footnotesize
    \setlength{\tabcolsep}{4pt}
    \captionsetup{font=footnotesize}
    \caption{\xadd{\textbf{Injections needed for repair} ($\downarrow$; mean $\pm$ std over $14$ Pong episodes and $45$ Cube episodes).}}
    \label{tab:learned-wm-baselines}
    \vspace{-4pt}
    \xadd{\begin{tabular}{lcc}
        \toprule
        Ranking & DIAMOND (Pong) & LeWM (Cube) \\
        \midrule
        \textsc{Recursive}-$\phi$ & $\mathbf{2.7 \pm 1.7}$ & $\mathbf{5.2 \pm 2.0}$ \\
        Frame error & $6.0 \pm 5.2$ & -- \\
        Latent error & $7.3 \pm 3.5$ & $6.4 \pm 2.1$ \\
        \bottomrule
    \end{tabular}}
    \vspace{-8pt}
\end{wraptable}

\xadd{There are three major observations from these experiments:
\emph{(i) The gap is one-sided and grows aligning with our assumptions that errors don't cancel out.} 
Over $48$ Pong seeds, DIAMOND over-predicts the return by $0.6 \pm 0.9$ points after $100$ steps and $23.9 \pm 5.0$ after $2000$; on Cube, LeWM predicts success for $21$ of $23$ failed episodes. Both models err within $50$--$100$ steps.
\emph{(ii) Transfer gaps from learned world models admit sparse repairs.} All $14$ Pong rollouts with a gap at $100$ steps are repaired by $2.7 \pm 1.7$ injected transitions, ranking steps by $\hat{\phi}_t$ in the direction of the gap.
On Cube, whose players are ten $5$-step action blocks, even though LeWM is less accurate and its errors accumulate faster, ranking by the exact $|\phi_t|$, $18$ of $45$ failures are repaired by fewer than five blocks.
\emph{(iii) $\phi$-values beat naive prediction-error baselines.} On both models, \textsc{Recursive}-$\phi$ needs much fewer injections than ranking by the model's one-step frame or latent prediction error (\cref{tab:learned-wm-baselines}). \cref{fig:learned-wm-cases} shows an example where 
one single-injection proposed by $\phi$-values repairs the gap while the top step of the prediction-error baselines (dashed) only detects superficial state discrepancies and repairs neither.
}

\section{Conclusions}
\label{sec:conclusion}

We introduced a framework for world models transfer gap attribution by identifying which time steps are \emph{causally responsible} for the policy's real world performance degradation. By treating each time step's dynamics selector as a player in a game and computing counterfactual Shapley values over hybrid rollouts that selectively replace simulated/world-model predictions with turbulant real world transitions, our method produces precise per-step causal attributions that explains the return gap. The newly proposed \textsc{Recursive}-$\phi$ is shown to be more sample efficient and scales to long trajectories by exploiting the sparsity of causal errors with a coarse-to-fine recursive estimation scheme. Experiments on Atari, ProcGen and MiniGrid environments with controlled failures demonstrate that \textsc{Recursive}-$\phi$ correctly identifies the steps that are causally important.
These results corroborate with the central premise of our work: not all model errors matter equally, and causal reasoning is necessary to distinguish the critical few from the irrelevant many. Our work paves the way for more principled policy transfer from world models to real world.

\newpage
\begin{ack}
This research is supported in part by the NSF, ONR, AFOSR, DoE, Amazon, JP Morgan, The Alfred P. Sloan Foundation and MATS. We would also like to thank Jeffrey Heninger from MATS for his thoughtful review and suggestions.
\end{ack}

\bibliography{bibwithID,extra}
\bibliographystyle{abbrvnat}

\newpage
\appendix
\section*{Appendices}
\DoToC

\section{Limitations and Future Work}
\label{sec:limitations-future-work}

\subsection{Limitations.}
Our method's computational cost, though reduced by coarse-to-fine estimation, still scales with the number of causal steps $K$, which may be prohibitive for very long trajectories with dense causal structure (i.e., when the sparsity assumption is violated). Additionally, the current framework assumes sign consistency of the causal steps which may not hold if within a rollout, there are both real world transition discrepancies that help the agent and those that hurt it. But the assumption usually holds empirically because systematic world model errors usually lead to performance loss rather than gain under a well-defined task reward function and an optimized policy under the world model. From a higher conceptual level, we assume the environment and world model can be described by MDP-SCMs while in practice, confounders are pervasive and non-MDP models are prevalent in decision making tasks~\citep{crlsurvey}. For the world model transfer gap attribution under more general environment models beyond MDP-SCMs is an interesting direction for future work.
\xadd{Our main evaluation relies on controlled sim-to-sim perturbations, in which the perturbation set, and hence the number of causal steps $K$ used for scoring, is known by design; on learned world models (\cref{sec:results-learned-wm}) there is no such ground truth, and our study is applied to two learned world models, scored by repair.}

\subsection{Future directions.}
Several extensions are worth pursuing. First, \xadd{extending the study of learned world models (\cref{sec:results-learned-wm}) to longer horizons, larger budgets, and more models would test how far} the attributions remain informative when the model's state representation differs from the simulator's. Second, the ranked attribution could be used to drive \emph{active model improvement}: collecting additional training data specifically at the high-attribution steps to improve the world model where it matters most. Third, integrating the diagnostic into online model-based planning: flagging high-attribution predictions before the agent commits to a plan could yield a safety mechanism that detects when the world model's errors are likely to cause real-world failures.

\begin{figure}[t]
    \centering
    \begin{subfigure}[t]{\linewidth}
        \centering
        \includegraphics[width=\linewidth]{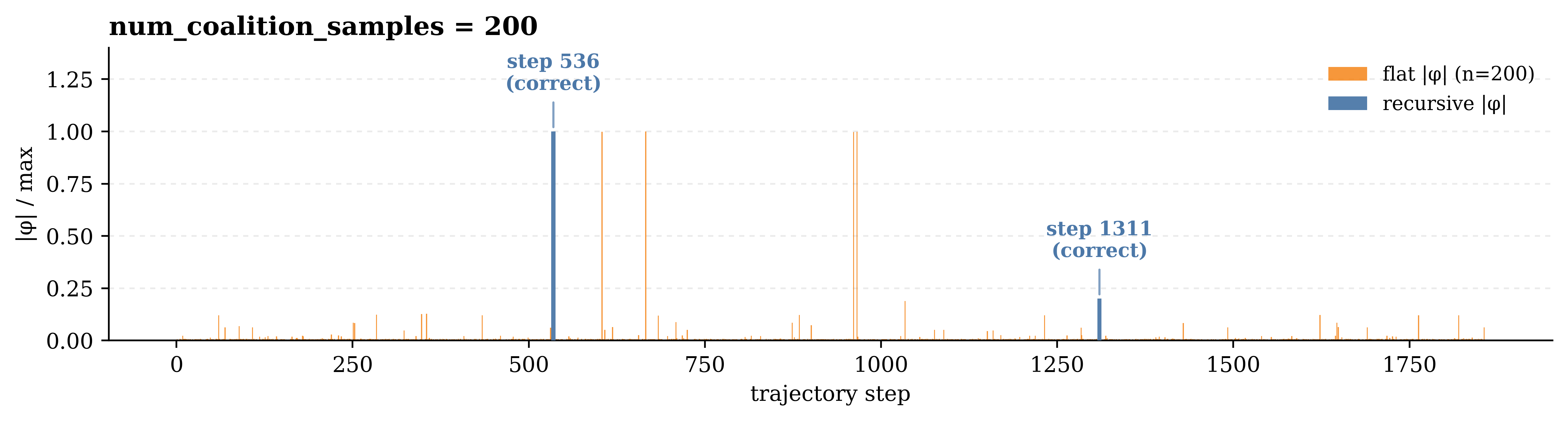}
        \caption{$M = 200$.}
        \label{fig:atari-phi-n200}
    \end{subfigure}
    \\[3pt]
    \begin{subfigure}[t]{\linewidth}
        \centering
        \includegraphics[width=\linewidth]{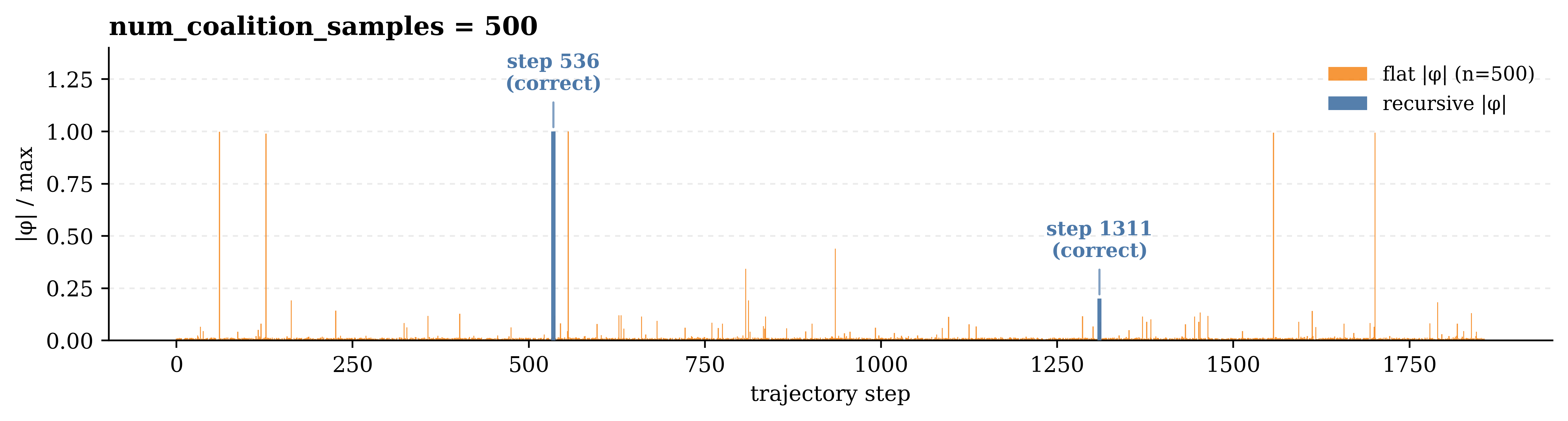}
        \caption{$M = 500$.}
        \label{fig:atari-phi-n500}
    \end{subfigure}
    \\[3pt]
    \begin{subfigure}[t]{\linewidth}
        \centering
        \includegraphics[width=\linewidth]{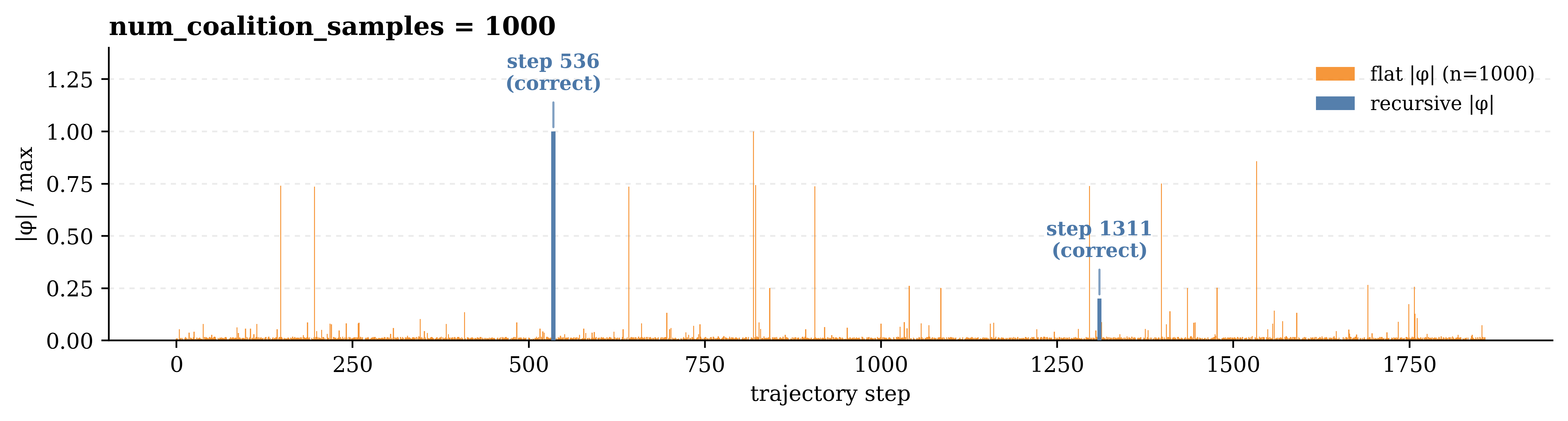}
        \caption{$M = 1000$.}
        \label{fig:atari-phi-n1000}
    \end{subfigure}
    \\[3pt]
    \begin{subfigure}[t]{\linewidth}
        \centering
        \includegraphics[width=\linewidth]{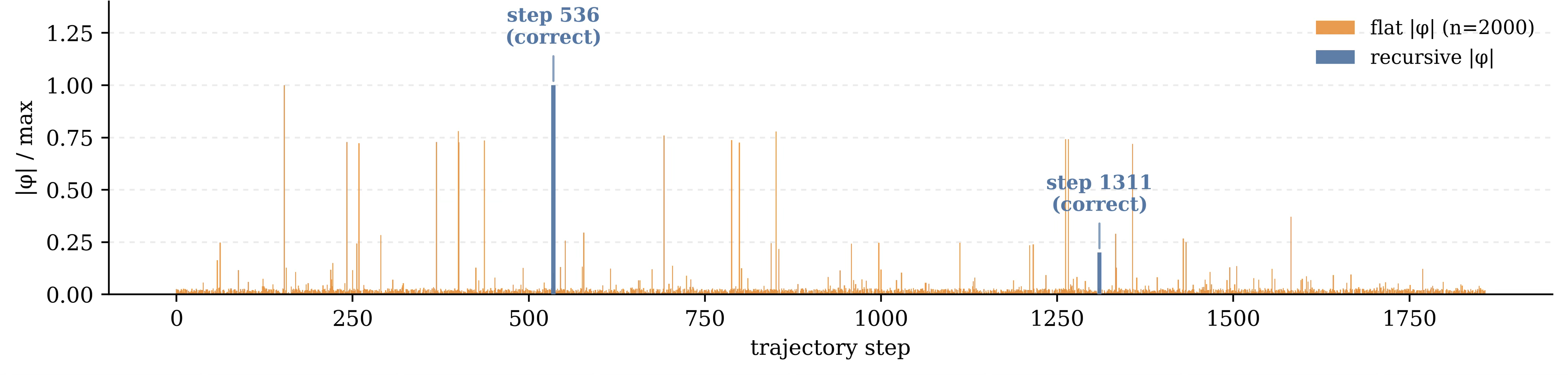}
        \caption{$M = 2000$.}
        \label{fig:atari-phi-n2000}
    \end{subfigure}
    \caption{Per-step $|\phi|$ traces (normalised by the per-method max) on the same $\sim\!1860$-step Pong trajectory across coalition budgets $M \in \{200, 500, 1000, 2000\}$. Recursive (blue) places its two largest spikes on the ground-truth causal blackouts at steps $536$ and $1311$ at every budget; flat (orange) is dominated by sampling variance and at no budget in this range does its top-$2$ recover both causal sites.}
    \label{fig:atari-phi-trace-sweep}
\end{figure}

\section{Detailed Experiment Setup}
\label{sec:exp design details}

\textbf{Common Protocol.} The major experiments share the sim-to-sim protocol: an unperturbed simulator stands in for the world model $\mathcal{M}_{\mathrm{wm}}$, a wrapped simulator with controlled error stands in for the real environment $\mathcal{M}_{\mathrm{env}}$, and the realised perturbation set is read off the wrapper as the mechanistic ground truth. Counterfactual rollouts use ALE's \texttt{clone\_state}/\texttt{restore\_state} API for Atari, ProcGen \texttt{gym3} state cloning for CoinRun, and direct state serialisation for MiniGrid; in all three cases the on-track branch of \textsc{CtfSim} (\cref{alg:ctf-sim}) reuses the observed transition exactly, while \xadd{coalition steps and off-track steps are generated by the world model, i.e., the unperturbed simulator restored to the corresponding state. The perturbed simulator $\mathcal{M}_{\mathrm{env}}$ is never reset or queried by \textsc{Recursive}-$\phi$ or \textsc{Flat}-$\phi$. The same policy $\pi$ acts in every counterfactual rollout,} so $\phi$-values measure whether each transition's deviation from the ground truth causally affected the return. Every cell is repeated over $10$ independent coalition-sampling seeds with \xadd{the observed trajectory's} environment seed and perturbation set held fixed, so per-seed variation in the flat estimators (\textsc{Flat}-$\phi$, SHAP, LIME) is averaged out rather than masked by a lucky draw. \textsc{Recursive}-$\phi$ uses split factor $B = 2$, discount $\gamma = 1$, and the optimal proposal $Q^{\star}_n$ of \citet{li2026credit} on each surviving sub-game. \xadd{Each experiment's setup below gives its maximum budget per round $M$ and its pruning threshold $\epsilon$. A round draws at most $M$ coalitions, and} once $2^{|\mathcal{A}|} \leq M$ the algorithm switches to exact enumeration on the remaining active set, eliminating sampling variance. \xadd{An enumerated round evaluates only $2^{|\mathcal{A}|}$ coalitions, so a run usually uses a small part of its maximum budget, and the rollout counts we report are the coalitions it actually evaluated.} \textsc{Flat}-$\phi$ uses the same $Q^{\star}_T$ proposal but allocates the full coalition budget over the $T$-player game in a single round; SHAP~\citep{lundberg2017unified} and LIME~\citep{ribeiro2016should} use the same $T$-player game but factual rollouts and the other thing that changes is how the value-gap signal $Y - Y^{\mathbf{z}}$ is converted into a per-step score (kernel-SHAP weights for \textsc{Flat}-$\phi$ and SHAP, sparse local linear regression for LIME).

\textbf{SHAP and LIME for Transfer Gap Attribution.} We adapt SHAP and LIME to the problem of transfer gap attribution. The influence function of the input instance is given by rolling out the policy again in the real world. Setting/Perturbing any of the steps to be 1 in the input instance means that we don't inject errors at that time step. For those steps with 0s, we randomly drawn with another seed (could be different from the seed used by the given trajectory) to decide whether an error is injected. The major difference between this factual simulation and the counterfactual simulation is that the environment randomness is not held the same as the one in the given trajectory. Thus, attribution methods based on such factual simulations evaluated influence functions \xadd{credit the errors that hurt on average over the drawn seeds rather than in the given trajectory. They find the causal steps of a specific trajectory instance only where the two coincide, e.g.\ at $L = 300$ in Experiment~1, where the causal blackout also costs points in $15$ of the $29$ other start states}. \xadd{Where the two differ, these baselines recover only by chance. At $L = 100$ the factual game ranks steps $52$ and $53$ above the causal step $48$ (factual Shapley magnitudes $0.59$ and $0.56$ against $0.38$). At $L = 500$ the causal step $140$ nearly ties with steps $54$ and $463$ (factual Shapley magnitudes $0.61$, $0.64$ and $0.56$). At $L = 1000$, where $|P| = 20$ rules out exact values, repairing the causal step $219$ first or last gains $0.46$ points on average over start states, against $0.48$ for step $934$ and $0.47$ for step $780$. On the seeds that reach the largest budget, LIME's top step is still split among these steps. The sample-efficiency metric below counts a seed as recovered at the first swept budget that restores the clean return, so LIME recovers on most seeds at $L \in \{500, 1000\}$ without its estimate settling on the causal step.}

\textbf{Sample-efficiency Metric.} For each method we sweep the coalition / sample budget over a grid over trajectory length and effective number of errors ($K$) (and environment variants for CoinRun) and record the smallest budget at which removing the method's top-$K$ steps from the injected set restores the clean return on a re-rollout. $K$ is identified by design or from \textsc{Recursive}-$\phi$ outputs and verified before using. ``-'' in \cref{tab:atari-results,tab:coinrun-results} marks cells where fewer than three of the ten coalition seeds recover within the largest budget tried, so the conditional mean is not statistically meaningful. 

\textbf{Experiment 1 Setup (\cref{sec:results-obsocclusion}).} We use \texttt{PongNoFrameskip-v4} via Gymnasium~\citep{gymnasium} with the standard Atari wrappers (\texttt{frameskip=4}, frame-stack $k = 4$, $84 \times 84$ grayscale) and the RL-Zoo3 PPO agent~\citep{rl-zoo3,schulmanProximalPolicyOptimization2017} acting deterministically (argmax over logits, so identical observations yield identical actions across coalition rollouts). We sweep trajectory length $L \in \{100, 200, 300, 500, 1000, 3000\}$ with environment seed $100$. The perturbation is an observation blackout: at a fixed set of step indices the wrapper zeroes the stacked-frame observation passed to the policy. The injected indices are drawn by shuffling $\{0, \ldots, L-1\}$ and slicing a length-dependent window of that shuffle ($|P| = 10$ for $L \in \{100, 200, 300, 500\}$ and $|P| = 20$ for $L \in \{1000, 3000\}$); fixing both shuffle seed and window keeps the perturbation set identical across coalition seeds and methods. \xadd{The full-game rollout of \cref{fig:teaser-cumreward,fig:atari-phi-trace} ($\sim\!1860$ steps) contains $15$ blackouts, two of which are causal.} Most blackouts are absorbed by the next few paddle moves, so the causal subset $K \leq |P|$ is recovered per trajectory as the smallest top-$K$ ranking under recursive $|\phi|$ whose removal from the injected set restores the clean return; the same $K$ is then handed to the flat / SHAP / LIME baselines for an apples-to-apples comparison. \xadd{\textsc{Recursive}-$\phi$ uses a maximum budget per round of $M = 100$ and $\epsilon = 10^{-10}$, and evaluates $28$ to $156$ coalitions in total (\cref{tab:atari-results}).} \textsc{Flat}-$\phi$ and SHAP are swept over coalition budgets $M \in \{100, 200, 500, 1000, 2000, 5000\}$ \xadd{(\textsc{Flat}-$\phi$ uses only $M \in \{2000, 5000\}$ for $L \geq 500$ and extends the grid to $10\,000$ at $L = 200$)}; LIME is swept over the same sample-count grid. \xadd{SHAP and LIME are run at every $L$ of \cref{tab:atari-results}. Each of their factual rollouts, in \cref{tab:atari-results} and \cref{fig:atari-phi-trace}, resets the environment with a seed drawn uniformly from $\{0, \ldots, 9999\}$, the observed seed $100$ included. In \cref{fig:atari-phi-trace} the baselines use $200$ coalitions, with coalition seed $100$ for \textsc{Flat}-$\phi$ and $101$ for SHAP and LIME. LIME ranks step $536$ among its five largest values on $9$ of the $10$ seeds of \cref{tab:atari-results} at $L = 3000$ with $200$ samples, but never ranks step $1311$ high. In the figure its value at step $536$ is $0.84$ of its maximum and sits behind the \textsc{Recursive}-$\phi$ bar. In this setup a reset seed only sets the number of no-op steps at the start of the episode, so these seeds give $30$ distinct start states, and $3\%$ of them give the start state of the observed trajectory.}

\textbf{Experiment 2 Setup (\cref{sec:results-mechanism}).} CoinRun~\citep{cobbeLeveragingProceduralGeneration2020} with a deterministic argmax cleanba-PPO policy (checkpoint from \href{https://huggingface.co/cleanrl/CoinrunHard-v0-cleanba_ppo_envpool_procgen-seed1}{clearnrl}, $15$ discrete actions, $3{\times}64{\times}64$ RGB observations) and trajectory lengths $N \in \{100, 300, 500, 1000\}$. Both variants are stochastic wrappers around the unmodified ProcGen \texttt{gym3} simulator: \emph{Variant A (poisonous coin)} draws $\mathrm{Bernoulli}(p_A)$ at every coin pickup and on success replaces the $+10$ reward with $-10$, leaving the pixel stream unchanged; \emph{Variant B (windy day)} draws $\mathrm{Bernoulli}(p_B)$ at every \texttt{UP}-containing action and on success applies a one-cell horizontal wind pushing the agent left, so the jump falls short. We use $p_A = 1.0$ (every coin flip fires) and $p_B = 0.7$, giving realised perturbation counts in the tens at $N = 1000$ on Variant B but a much smaller causal support because most short jumps are absorbed by the policy. The realised perturbation set is read off the wrapper attribute as ground truth; for the recovery metric a repair rollout replays the same wrapper with $\text{perturb\_steps} = P \setminus \text{top-}K$ and the budget is recorded as recovered when its return matches the clean return exactly. \textsc{Flat}-$\phi$, SHAP, and LIME are each swept over the budget grid $\{100, 200, 500, 1000, 2000, 5000\}$. \textsc{Recursive}-$\phi$ uses \xadd{a maximum} budget $M = 100$ per round \xadd{and $\epsilon = 10^{-10}$, and evaluates $28$ to $276$ coalitions in total (\cref{tab:coinrun-results})}.

\textbf{Experiment 3 Setup (\cref{sec:results-longhorizon}).} The custom \texttt{MiniGrid-CenterGoal5x5-v0} environment is a $5{\times}5$ room with the goal at the centre; a hand-crafted policy executes the deterministic clockwise pattern \texttt{[forward, forward, right]} for $n$ inner-ring steps, then takes the BFS shortest path to the goal. Total trajectory length $T = n + (\text{steps to goal})$, so $n \in \{100, 500, 1000, 2000, 5000\}$ for SHAP, LIME and \textsc{Flat}-$\phi$, and additionally $\{10\,000, 10^6\}$ only for \textsc{Recursive}-$\phi$ (which is cheap enough to push to $T = 1\text{M}$ and is the only curve drawn in \cref{fig:phi-sweep-budget} at the very end). The perturbation is a deterministic time penalty applied by a \texttt{TimePenaltyWrapper}: a constant $-0.1$ reward is subtracted at exactly $K \in \{1, 2, 5\}$ pre-committed randomly sampled time steps. $K$ is exactly the cardinality of the causal set and the recovery / recall-at-$K$ metric reduces to checking whether the top-$K$ of the attribution output equals the planted set. \textsc{Flat}-$\phi$, SHAP, and LIME share the budget grid $\{100, 500, 1000, 4000, 8000, 10000, 20000, 40000, 60000, 100000\}$. \textsc{Recursive}-$\phi$ uses \xadd{a maximum budget per round of $M = 10\,000$ for $n \leq 5000$ and $M = 70\,000$ for $n \in \{10\,000, 10^6\}$, with $\epsilon = 10^{-3}$ at $n = 100$, $10^{-5}$ for $500 \leq n \leq 10\,000$, and $10^{-8}$ at $n = 10^6$. At most $2K$ blocks stay active, so every round is enumerated and a run evaluates only $28$ to $16\,916$ coalitions in total (\cref{fig:phi-sweep-budget}), far below $M \cdot \lceil \log_2 T \rceil$}.

\xadd{\textbf{Experiment 4 Setup (\cref{sec:results-learned-wm}).} \emph{DIAMOND on Pong.} We use the pretrained Atari-100k Pong checkpoint released with DIAMOND~\citep{Alonso_2024}, which bundles the diffusion world model, its reward model, and an actor-critic policy trained inside the world model; actions are sampled from the policy. The real environment is \texttt{ALE/Pong-v5} with sticky actions (repeat probability $0.25$), frame skip $4$, and $64{\times}64$ RGB observations, whereas the checkpoint was trained on the deterministic game. Real and world-model rollouts start from the same four-frame context; in the world-model rollout the policy acts on the imagined frames (three denoising steps per frame) and rewards come from the reward model. The gap statistics use $48$ seeds and $2000$ steps. Attribution uses the first $100$ steps with a maximum budget per round of $M = 1500$ coalitions, $B = 2$, $\gamma = 1$, and the same sampling-noise seed in every counterfactual rollout; rounds with $2^{|\mathcal{A}|} \leq M$ are enumerated and pruned at $\epsilon = 10^{-4}$, the others are sampled and pruned at $\epsilon = 10^{-10}$. We attribute all $14$ of the $48$ seeds whose world-model rollout under this noise seed differs in return from the real one over these steps; the model over-predicts on each of them, by $1$ to $3$ points. These rollouts are separate from the $2000$-step ones of \cref{tab:pong-wm-gap}, hence the different count at $100$ steps. Attribution takes $4371$ counterfactual rollouts per trajectory on average ($884$ to $6276$). Since an imagined frame never equals the real one exactly, the on-track test of \cref{eqn:sigma-z} is replaced by its most permissive form: at a step outside the coalition, the observed action, reward, and next frame are injected and teacher-forced into the model's context. For the repair score we inject at the top-$k$ ranked steps, run the world model elsewhere, and report the smallest $k \in \{1, \ldots, 10, 12, 15, 20, 30, 50, 75, 100\}$ at which the predicted return equals the real one. We use two rankings of the attributions: by magnitude, $|\hat{\phi}_t|$, as in the controlled experiments, and in the direction of the gap, i.e., by decreasing $\hat{\phi}_t \operatorname{sign}(\Delta)$, which puts the steps whose attribution has the sign of the gap first. As baselines, we rank steps by the model's one-step prediction error along the real trajectory: given the real frames and actions as context, the model predicts the next frame, and the step is scored by the mean squared error of this frame (frame error) or by its $\ell_2$ distance to the real frame in the feature space of the reward model's encoder (latent error). Each baseline needs one model call per step.}

\xadd{\emph{LeWM on Cube.} We use the pretrained Cube checkpoint of LeWM~\citep{maes2026leworldmodel} with its planner (CEM with $300$ samples and $30$ iterations over $5$ action blocks of $5$ actions each, replanning every $25$ steps) on OGBench Cube~\citep{park2025ogbench}. Following the evaluation protocol of the model, an episode starts from an expert state, its goal is the expert state $25$ steps later, and the budget is $50$ steps; we use $30$ start/goal pairs in which the cube moves by more than $6$\,cm. The outcome is the planner's own objective, the squared distance between the final latent and the goal embedding (goal cost), and an episode counts as predicted to succeed when its imagined goal cost is below $1.2$ times the largest real goal cost among successful episodes. LeWM predicts one latent per block of five environment actions, so a block is the model's elementary transition and the players are the ten executed action blocks. At a block outside the coalition, the real latent and its successor are injected if the counterfactual latent is within $\ell_2$ distance $10$ of the real one. We use a maximum budget per round of $M = 1500$, $B = 2$, and $\epsilon = 10^{-4}$; the last of the four rounds needs $2^{10} = 1024$ coalitions, so every round is enumerated and the attributions are exact. We count a repair when the predicted goal cost is within $10\%$ of the free rollout's error from the real cost and on the same side of the success threshold. Candidates are optimistic failures, i.e., real failures that the model predicts to succeed: the $21$ among the $30$ pairs and $24$ from further planner rollouts on new start/goal pairs, $45$ in total. The collection stopped once a first pass at $M = 300$ had found ten episodes with a repair of fewer than five blocks; all $45$ candidates are attributed at $M = 1500$, and all of them are reported. The imagined goal cost is below the real one in $43$ of the $45$. Blocks are ranked by magnitude and in the direction of the gap as on Pong, and injected for $k = 1, \ldots, 10$. The baseline ranks blocks by the $\ell_2$ error of the model's one-step latent prediction from the real latents; LeWM has no decoder, so a frame error is not available.}

\section{Extended Experiment Results}
\label{sec:extended-results}

We mainly provide extra experiment results in comparing \textsc{FLAT}-$\phi$ against \textsc{Recursive}-$\phi$ because from the main experiment section, both non-causal baselines perform poorly due to their inability to pinpoint causal steps using associational influence functions.

\subsection{Atari Pong: detection traces across coalition budgets}
\label{sec:extended-atari-budgets}

\cref{fig:atari-phi-trace} in the main paper reports the per-step $|\phi|$ trace \xadd{at $L = 3000$ for a single coalition budget}. To make the comparison less anecdotal, \cref{fig:atari-phi-trace-sweep} sweeps the coalition budget $M \in \{200, 500, 1000, 2000\}$ on the same $\sim\!1860$-step Pong rollout, with the same two ground-truth causal blackouts at steps $536$ and $1311$ and the same \xadd{thirteen} non-causal blackouts. Recursive uses the budget once per refinement round (with $B = 2$, $\epsilon = 10^{-10}$); flat uses the full budget over the $T$-player game.

\textbf{Recursive is budget-stable.}
Across all four panels the recursive trace looks essentially identical: a saturated spike at step $536$ and a smaller but isolated spike at step $1311$, with every other position at nearly zero. The reason is structural rather than statistical, once the coarse-to-fine pruning collapses the inactive bulk of the horizon (after the first two or three rounds), the surviving sub-game is small enough that any of the budgets in $\{200, 500, 1000, 2000\}$ already exceeds $2^{|\mathcal{A}|}$ and the algorithm switches to exact enumeration on the remaining blocks. Increasing $M$ buys nothing once enumeration kicks in, which is why the recursive trace is visually invariant to $M$ in this regime. The two causal sites are the unambiguous top-$2$ at every budget, so top-$K$ recovery of the clean return is achieved already at $M = 200$.

\textbf{Flat improves but does not converge in this range.}
Flat's trace, in contrast, changes substantially with $M$ and never localises both causal sites cleanly. At $M = 200$ several non-causal positions ($\approx 600$, $\approx 660$, $\approx 970$) saturate at the per-method max and dwarf the causal site at step $1311$, so the flat top-$2$ misses both ground-truth blackouts. At $M = 500$ the dominant spurious spikes shift (now $\approx 60$, $\approx 125$, $\approx 1700$), step $536$ rises into the top tier, but step $1311$ remains buried under multiple non-causal peaks. At $M = 1000$ a single non-causal position near step $810$ still saturates, and although step $536$ is now reliably among the tallest spikes, step $1311$ is comparable to four or five distractors and would still be missed by a top-$2$ readout. At $M = 2000$ flat is closest to recursive, step $536$ is among the largest peaks and step $1311$ is visible, but a non-causal spike near step $150$ still saturates and the gap between causal and non-causal mass is not yet decisive. The per-method max-normalisation in each panel hides the absolute shrinkage of flat's variance with $M$, but what matters operationally, the \emph{ranking} of causal vs.\ non-causal positions, only flips reliably at budgets well above $2000$ (consistent with the $\geq 25\times$ gap reported in \cref{tab:atari-results}).

\textbf{Implication for budget selection.}
The qualitative picture across the four panels reinforces the takeaway of the main-paper sample-efficiency curve: the practical question is not ``how large does $M$ need to be for flat to converge'' but ``does the coarse-to-fine scheme reach exact enumeration on the surviving sub-game.'' Once it does, recursive's output is deterministic up to the on-track coupling and increasing $M$ has no effect; until flat does, its top-$K$ readout is governed by sampling variance and can rank a non-causal blackout above a causal one even at $M = 2000$ on a horizon of less than $2000$ steps. This is exactly the regime that motivates the recursive scheme.

\subsection{Atari Pong: detection traces across trajectory lengths}
\label{sec:extended-atari-lengths}

\begin{figure}[ht]
    \centering
    \begin{subfigure}[t]{.9\linewidth}
        \centering
        \includegraphics[width=\linewidth]{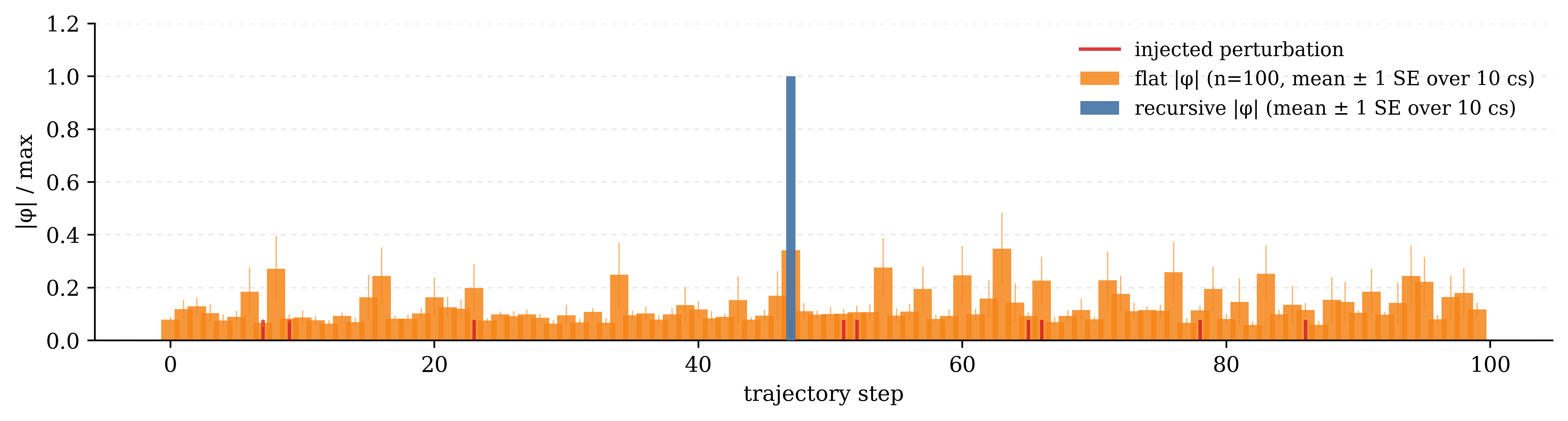}
        \caption{$L = 100$, $|P| = 10$, $K = 1$, gap $= 2$. Min flat budget $\in [500, 1000]$ over $10$ cs (10/10 recovered).}
        \label{fig:traj-sweep-L100}
    \end{subfigure}
    \\[3pt]
    \begin{subfigure}[t]{.9\linewidth}
        \centering
        \includegraphics[width=\linewidth]{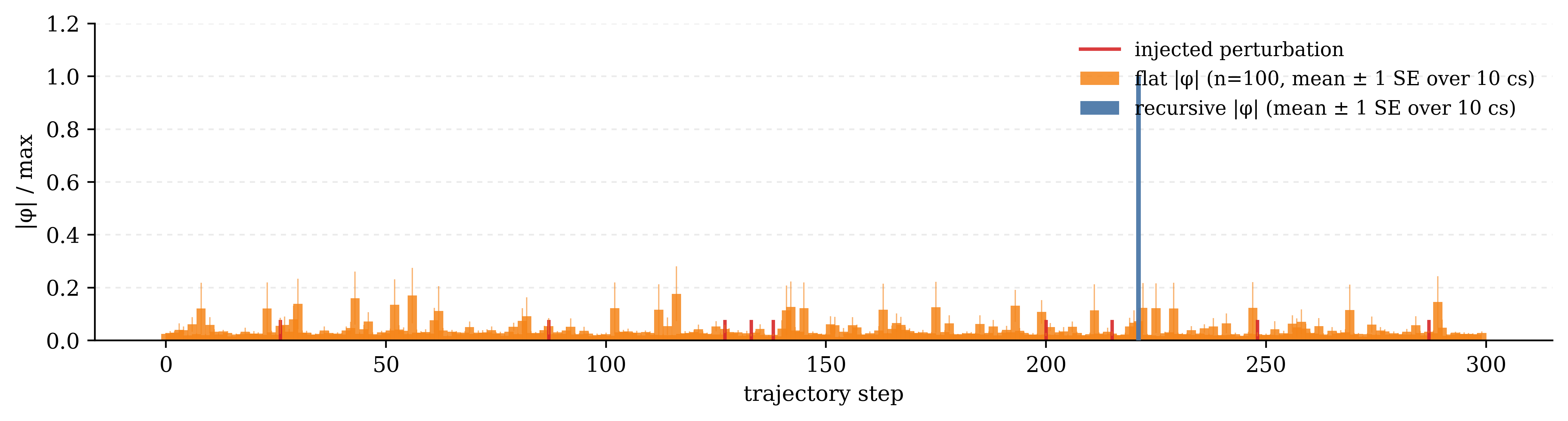}
        \caption{$L = 300$, $|P| = 10$, $K = 1$, gap $= 1$. Min flat budget $\in [1000, 5000]$ over $10$ cs (10/10 recovered).}
        \label{fig:traj-sweep-L300}
    \end{subfigure}
    \\[3pt]
    \begin{subfigure}[t]{.9\linewidth}
        \centering
        \includegraphics[width=\linewidth]{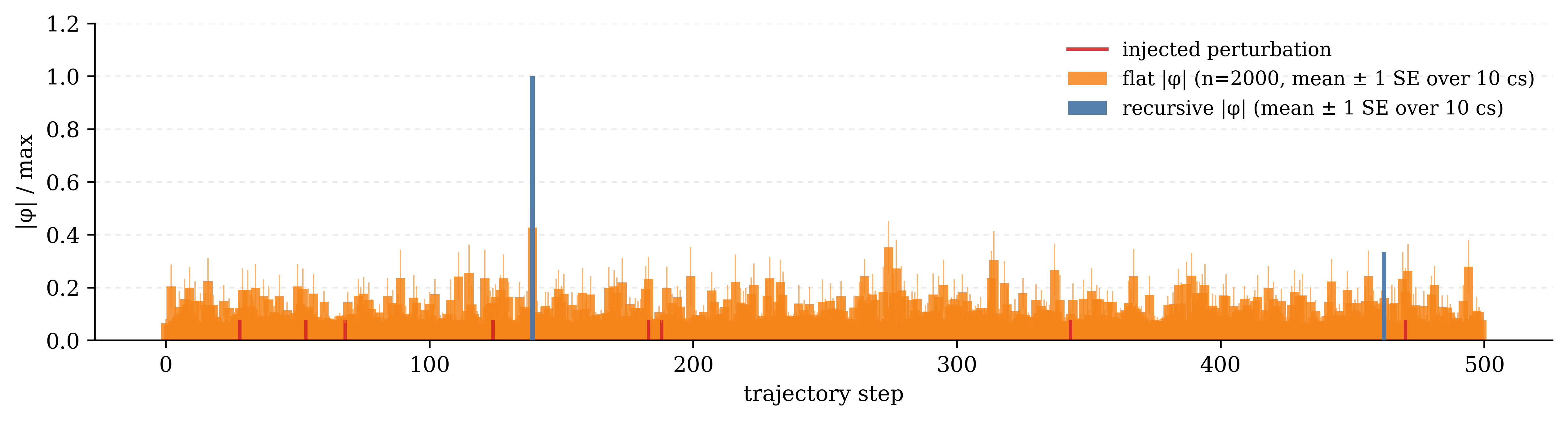}
        \caption{$L = 500$, $|P| = 10$, $K = 1$, gap $= 4$. \xadd{Flat swept at $M \in \{2000, 5000\}$: $1/10$ cs recovered (at $M = 2000$).}}
        \label{fig:traj-sweep-L500}
    \end{subfigure}
    \\[3pt]
    \begin{subfigure}[t]{.9\linewidth}
        \centering
        \includegraphics[width=\linewidth]{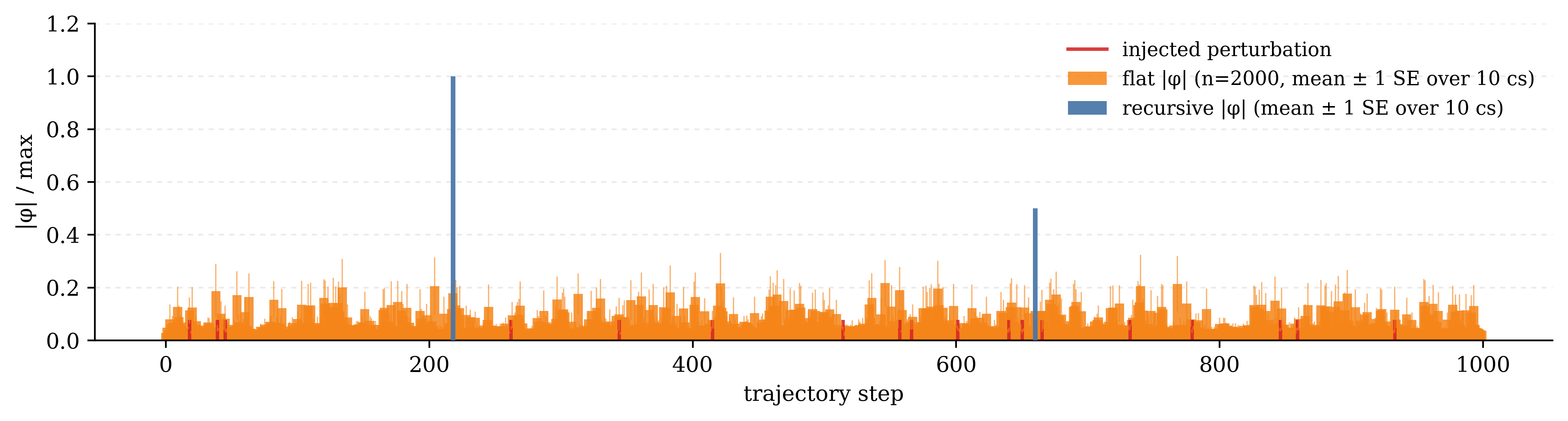}
        \caption{$L = 1000$, $|P| = 20$, $K = 1$, gap $= 3$. \xadd{Flat swept at $M \in \{2000, 5000\}$: $0/10$ cs recovered.}}
        \label{fig:traj-sweep-L1000}
    \end{subfigure}
    \\[3pt]
    \begin{subfigure}[t]{.9\linewidth}
        \centering
        \includegraphics[width=\linewidth]{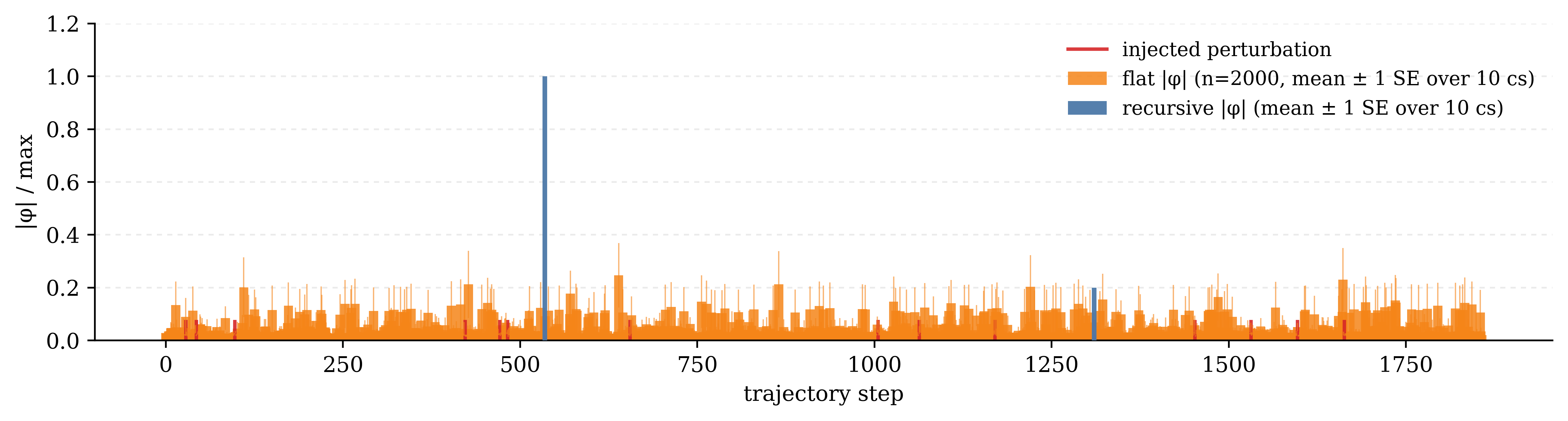}
        \caption{$L = 3000$, $|P| = 20$, $K = 2$, gap $= 3$. \xadd{Flat swept at $M \in \{2000, 5000\}$: $0/10$ cs recovered.}}
        \label{fig:traj-sweep-L3000}
    \end{subfigure}
    \caption{Per-step $|\phi|$ traces across $L$. Blue: \textsc{recursive}-$\phi$. Orange: \textsc{flat}-$\phi$ at $M = 100$ \xadd{(recursive's $M$) for $L \leq 300$ and at $M = 2000$ for $L \geq 500$}. Red ticks: injected blackouts\xadd{. Removing recursive's top-$K$ restores the clean return on $10/10$ seeds at every length.}}
    \label{fig:traj-length-sweep}
\end{figure}

\cref{tab:atari-results} reports the headline sample-efficiency numbers from the $L$-sweep as a single point per length. \cref{fig:traj-length-sweep} unpacks them into per-step $|\phi|$ traces (\xadd{$|\phi|$ divided by its per-seed maximum,} mean $\pm$ 1 SE over $10$ coalition-sampling seeds; env seed and perturbation set held fixed within each length). Red ticks mark every injected blackout\xadd{ (at $L = 3000$ the game ends at step $1858$, so panel (e) spans $1858$ steps)}, and each \xadd{subcaption} records the gap, the recovered $K$, and the \xadd{outcome of the flat budget sweep} over the $10$ seeds.

\textbf{Sparse causal support persists as $L$ grows.}
Although $|P| \in \{10, 20\}$ blackouts are injected at every length, only $K \in \{1, 1, 1, 1, 2\}$ are causal at $L \in \{100, 300, 500, 1000, 3000\}$. \textsc{Recursive}-$\phi$ recovers the full clean return by removing a single blackout for $L \leq 1000$ and two for $L = 3000$. The remaining injected sites (red ticks visible in every panel) are \xadd{not needed to restore the clean return. Most} are absorbed within a few paddle moves and \xadd{receive no credit}, even when they fall close in time to a gateway blackout (e.g.\ \xadd{steps $472$ and $483$, shortly before the gateway blackout at step $536$ at $L = 3000$). Two carry a smaller effect of their own, step $463$ at $L = 500$ and step $661$ in the cluster of five blackouts between steps $600$ and $700$ at $L = 1000$, but neither is part of the recovered top-$K$ (see below).} The transfer gap is not always equal to $K$: a single gateway blackout can perturb the agent's policy state long enough to shift the rally cadence and shave more than one point off the score within the fixed $L$-step window, which is why the gap is $2, 1, 4, 3, 3$ while $K$ stays at $1$ or $2$. This is the $K \ll |P|$ regime that the coarse-to-fine scheme is designed for.

\textbf{Recursive is causally faithful and budget-stable at every $L$.}
At every length, the blue recursive bars \xadd{peak on the gateway blackout (and on the second causal blackout, step $1311$, at $L = 3000$)} and sit at nearly zero everywhere else\xadd{, apart from one smaller spike at $L \in \{500, 1000\}$ discussed below}, with error bars too small to see on the spike heights. The reason is structural: once pruning collapses the inactive bulk of the horizon (one or two rounds suffice in this regime), the surviving sub-game has size $|\mathcal{A}|$ small enough that $2^{|\mathcal{A}|} \leq M = 100$ and the algorithm switches to exact enumeration on the remaining blocks. From that round onward the recursive readout is deterministic up to the on-track coupling, which is why the per-seed standard error on the recursive bars is zero, \xadd{and} $10/10$ coalition seeds restore the clean return at every length. A second\xadd{,} smaller spike appears at \xadd{$L \in \{500, 1000\}$ on another injected blackout (step $463$ at $L = 500$, step $661$ at $L = 1000$). Removing only that blackout recovers part of the gap ($2$ of $4$ points at $L = 500$, $2$ of $3$ at $L = 1000$), while removing the gateway blackout alone restores the clean return}, so \xadd{the second spike} is correctly ranked below the gateway step and excluded from the recovered top-$K$.

\textbf{Flat at $M = 100$ is below threshold and degrades with $L$.}
At $L = 100$ and $L = 300$, the orange flat bars at the same $M = 100$ \xadd{as recursive's maximum budget per round} are dominated by sampling variance: at $L = 100$ several non-causal positions are within a factor of two of the causal spike's height, and at $L = 300$ the causal site barely separates from the floor of distractors. The minimum recovering flat budget rises sharply with $L$: $[500, 1000]$ at $L = 100$ and $[1000, 5000]$ at $L = 300$ across the $10$ coalition seeds, both well above the $M = 100$ shown. \xadd{For $L \geq 500$ the flat sweep uses only $M \in \{2000, 5000\}$. It recovers on $1/10$ seeds at $L = 500$ (at $M = 2000$) and on none at $L \in \{1000, 3000\}$. Panels (c)--(e) show flat at $M = 2000$.} This is the regime in which the scaling gap of the two estimators becomes operationally decisive: at $L = 1000$ recursive uses $148$ rollouts to localise the single gateway blackout, while flat does not converge in $> \xadd{30}\times$ that budget.

\textbf{Coalition-seed variance.}
Recursive's terminal output is invariant to the coalition seed in this regime: every panel shows zero standard error on the spike heights because the surviving sub-game collapses to $|\mathcal{A}| \leq \log_2 M$ players within a couple of refinement rounds and the algorithm switches to exact enumeration. Flat's per-seed minimum recovering budget, by contrast, varies by up to $5\times$ at $L = 300$ ($1000$--$5000$ across the $10$ seeds), reflecting how strongly the kernel-SHAP estimator depends on which coalitions happen to be drawn. Reporting the typical-seed budget rather than the luckiest is therefore essential to a fair comparison; using a single seed for flat would have given a $5\times$ band of headline numbers at $L = 300$ alone.

\subsection{ProcGen CoinRun: detection traces across trajectory lengths}
\label{sec:extended-coinrun-lengths}

\begin{figure}[ht]
    \centering
    \begin{subfigure}[t]{\linewidth}
        \centering
        \includegraphics[width=\linewidth]{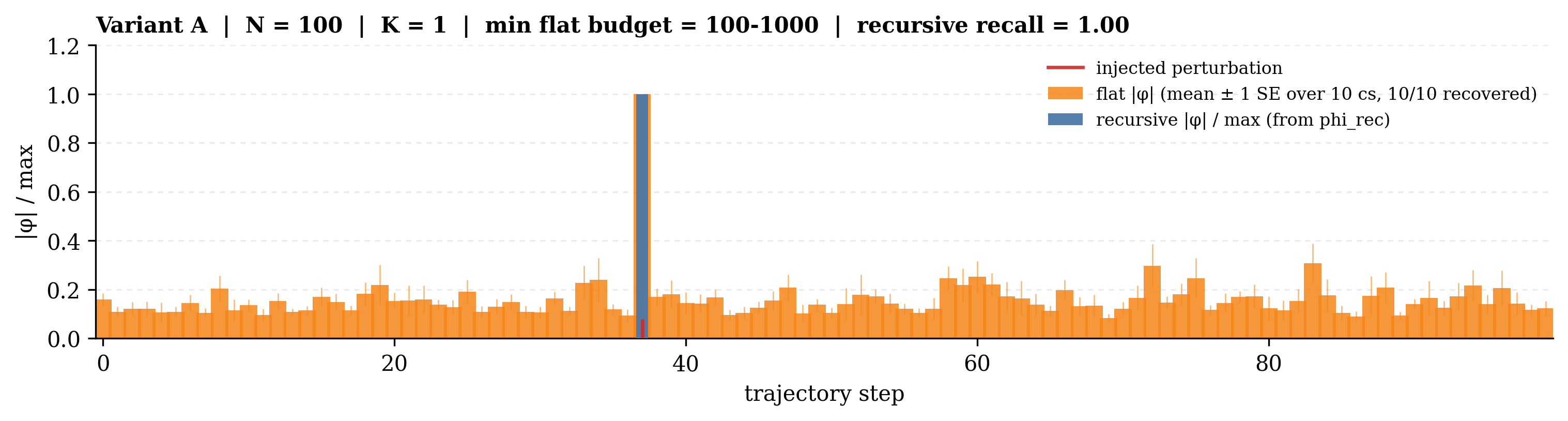}
        \caption{$N = 100$. $K = 1$, min flat budget $\in [100, 1000]$ over $10$ cs ($10/10$ recovered).}
        \label{fig:phi-coinrun-A-N100}
    \end{subfigure}
    \\[3pt]
    \begin{subfigure}[t]{\linewidth}
        \centering
        \includegraphics[width=\linewidth]{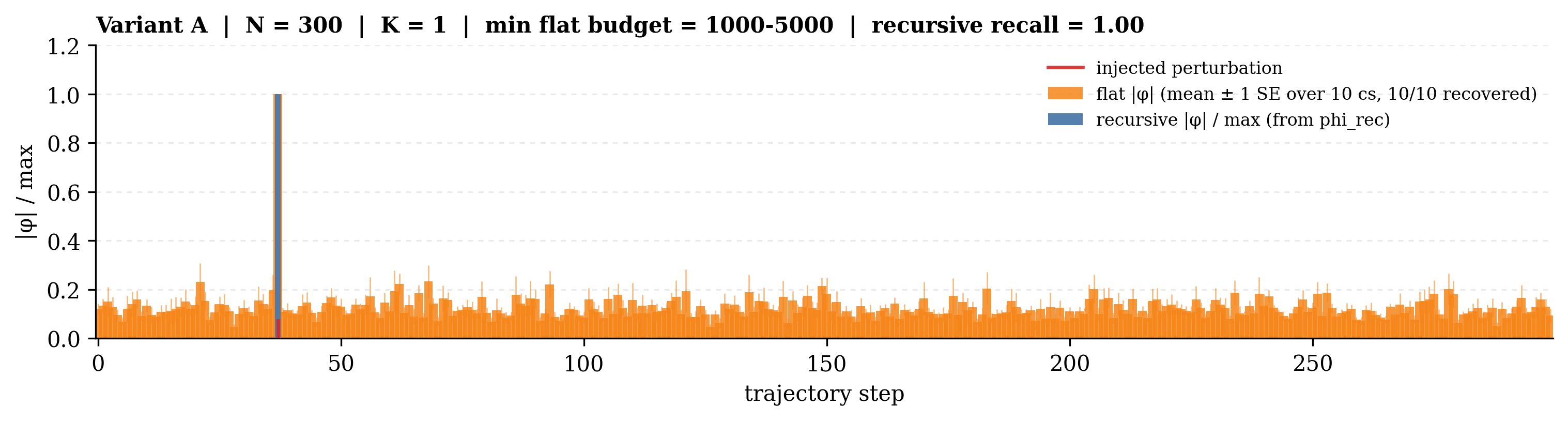}
        \caption{$N = 300$. $K = 1$, min flat budget $\in [1000, 5000]$ ($10/10$ recovered).}
        \label{fig:phi-coinrun-A-N300}
    \end{subfigure}
    \\[3pt]
    \begin{subfigure}[t]{\linewidth}
        \centering
        \includegraphics[width=\linewidth]{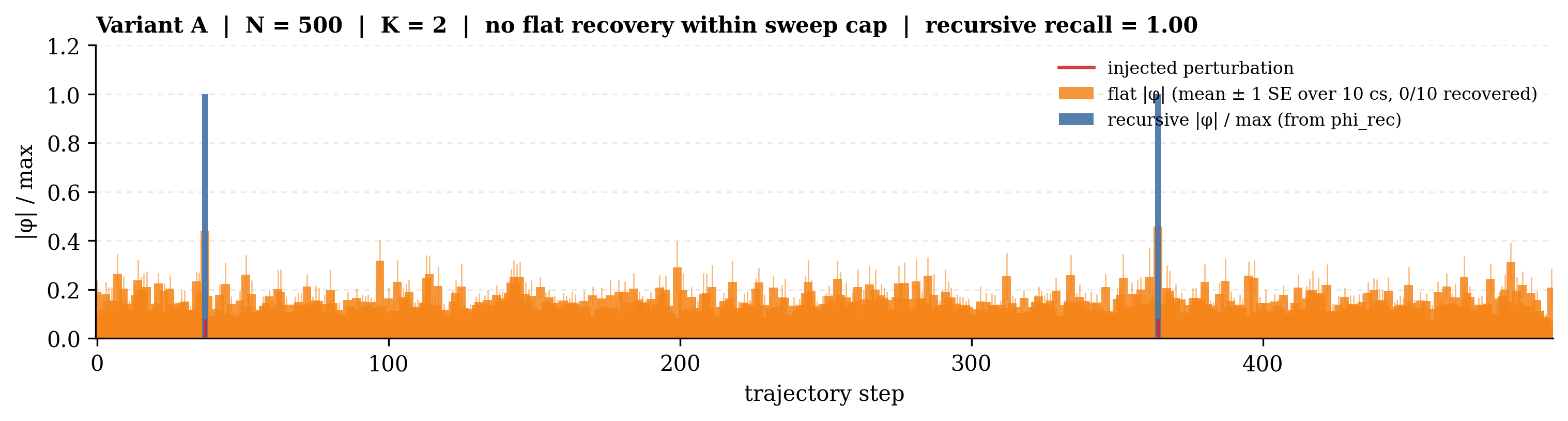}
        \caption{$N = 500$. $K = 2$. No flat recovery within $5\text{k}$ cap on any of the $10$ cs.}
        \label{fig:phi-coinrun-A-N500}
    \end{subfigure}
    \\[3pt]
    \begin{subfigure}[t]{\linewidth}
        \centering
        \includegraphics[width=\linewidth]{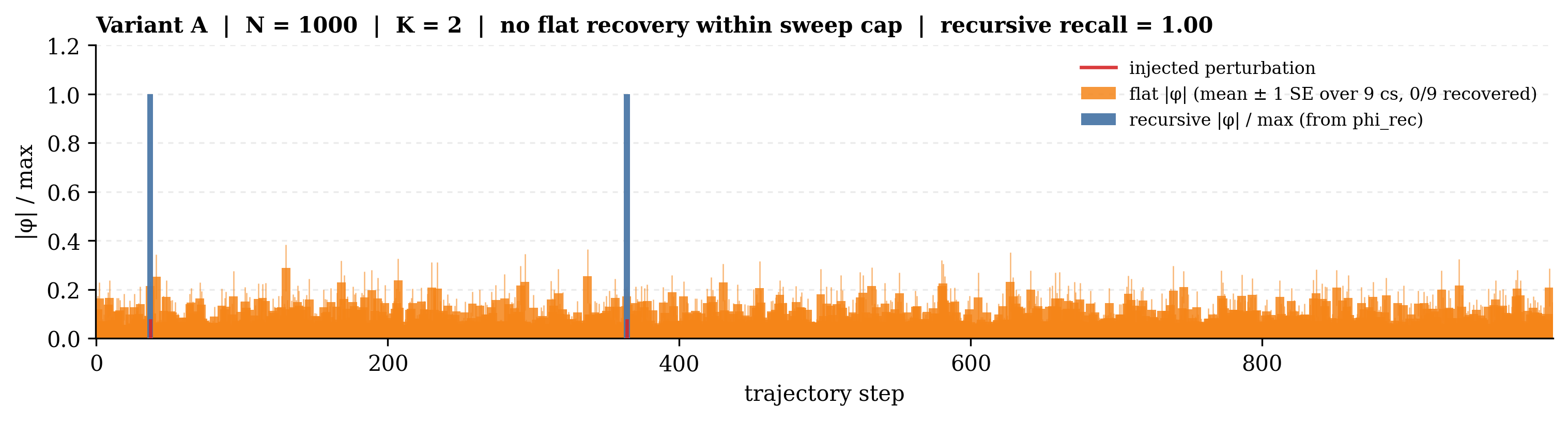}
        \caption{$N = 1000$. $K = 2$. No flat recovery within $5\text{k}$ cap on $9/9$ available cs.}
        \label{fig:phi-coinrun-A-N1000}
    \end{subfigure}
    \caption{\textbf{Variant A (poisonous reward): per-step $|\phi|$ traces across $N$.} Orange: flat $|\phi|/\max$, mean $\pm$ 1 SE across $10$ coalition-sampling seeds at each seed's recovering budget (or $5\text{k}$ if no budget recovered). Blue: recursive $|\phi|/\max$ from the deterministic enumerate-on-survive run. Red ticks at the bottom mark the injected perturbation steps. Recursive recovers $100\%$ of the transfer gap on every panel.}
    \label{fig:coinrun-phi-trace-sweep-A}
\end{figure}

\begin{figure}[ht]
    \centering
    \begin{subfigure}[t]{\linewidth}
        \centering
        \includegraphics[width=\linewidth]{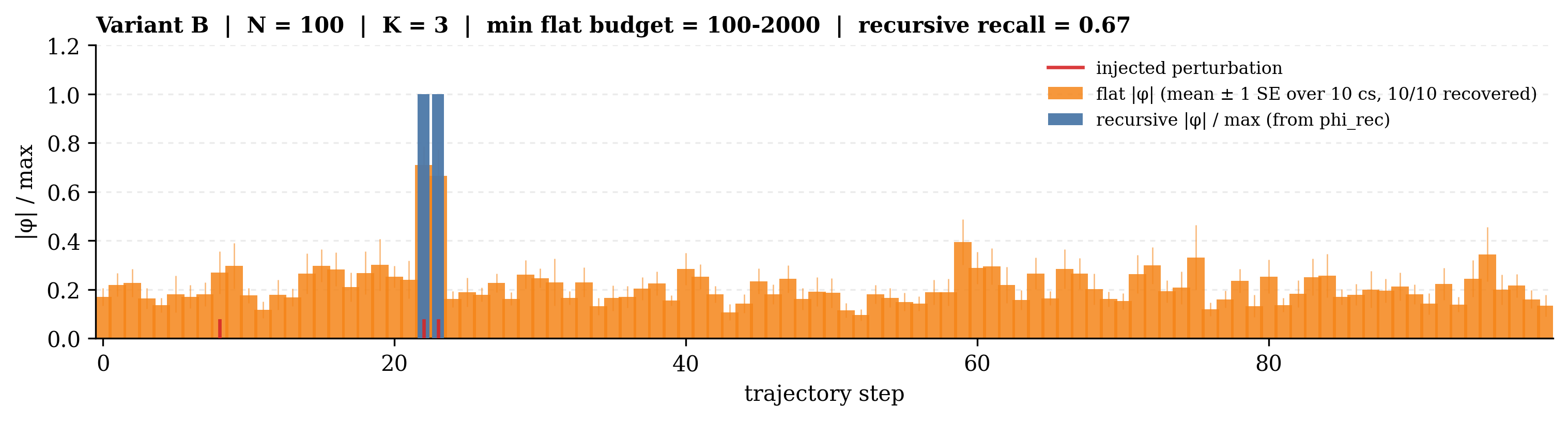}
        \caption{$N = 100$. $K = 3$, min flat budget $\in [100, 2000]$ ($10/10$ recovered).}
        \label{fig:phi-coinrun-B-N100}
    \end{subfigure}
    \\[3pt]
    \begin{subfigure}[t]{\linewidth}
        \centering
        \includegraphics[width=\linewidth]{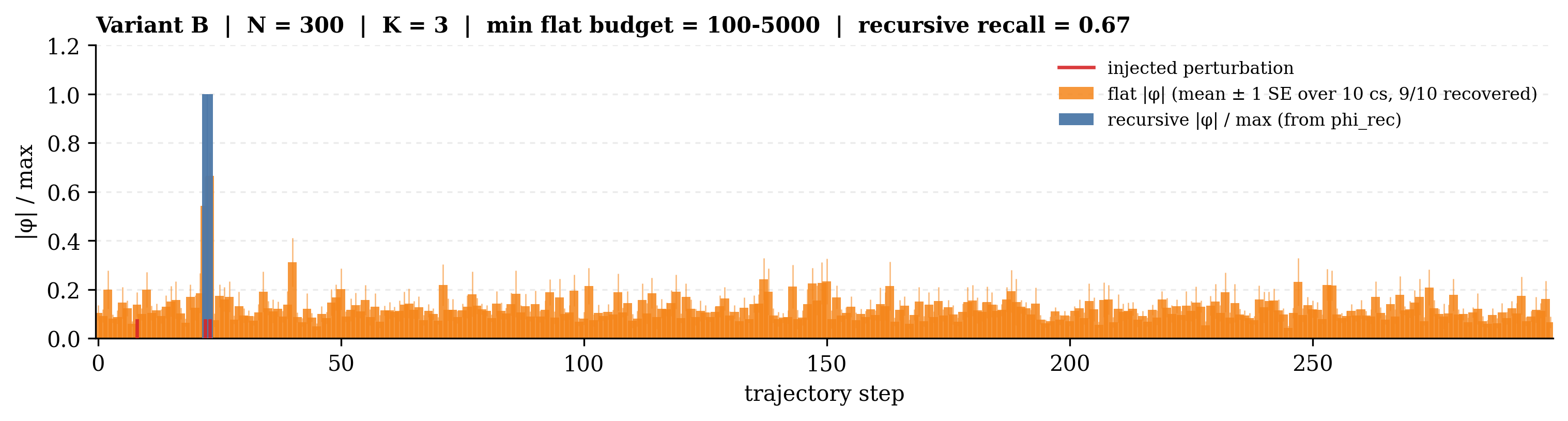}
        \caption{$N = 300$. $K = 3$, min flat budget $\in [100, 5000]$ ($9/10$ recovered).}
        \label{fig:phi-coinrun-B-N300}
    \end{subfigure}
    \\[3pt]
    \begin{subfigure}[t]{\linewidth}
        \centering
        \includegraphics[width=\linewidth]{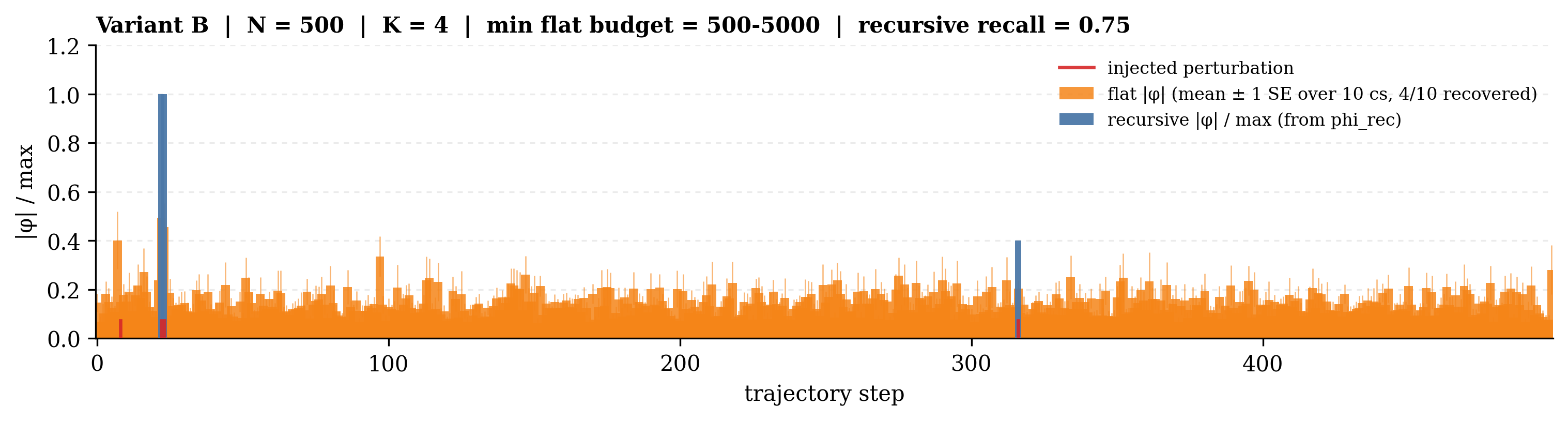}
        \caption{$N = 500$. $K = 4$, min flat budget $\in [500, 5000]$ ($4/10$ recovered).}
        \label{fig:phi-coinrun-B-N500}
    \end{subfigure}
    \\[3pt]
    \begin{subfigure}[t]{\linewidth}
        \centering
        \includegraphics[width=\linewidth]{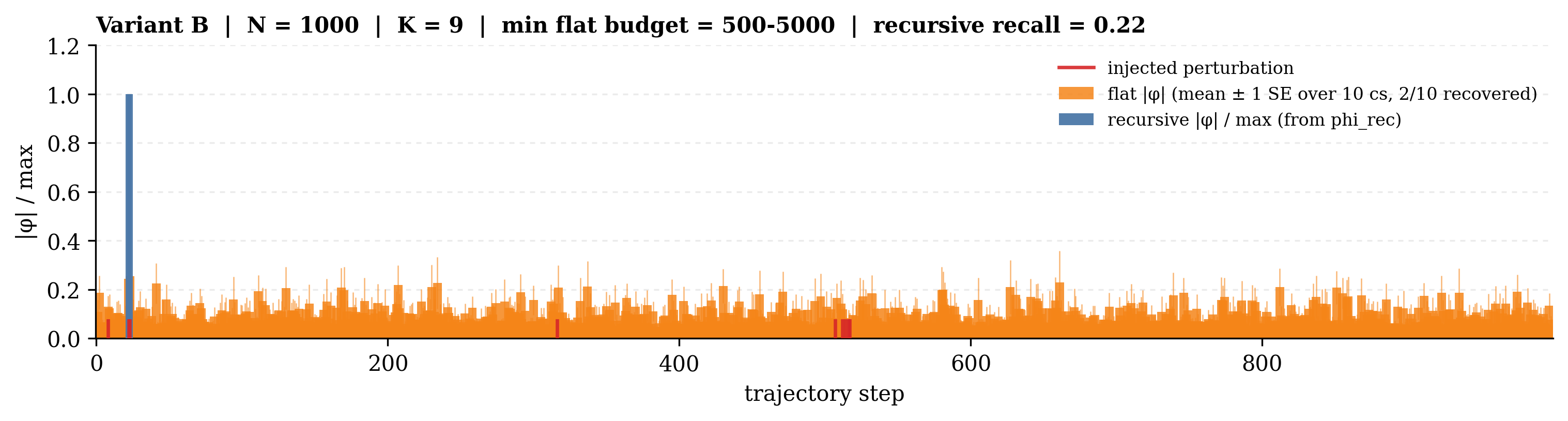}
        \caption{$N = 1000$. $K = 9$ (recursive realisation) / $10$ (flat realisation), min flat budget $\in \{500, 5000\}$ ($2/10$ recovered).}
        \label{fig:phi-coinrun-B-N1000}
    \end{subfigure}
    \caption{\textbf{Variant B (windy day): per-step $|\phi|$ traces across $N$.} Same conventions as \cref{fig:coinrun-phi-trace-sweep-A}. Recursive recovers $100\%$ of the transfer gap on every panel; the gap between recursive's top-$K$ and the (mechanistic) injected set widens with $N$: most realised short-jumps are absorbed by the policy and only the gateway events carry non-zero recursive $\phi$.}
    \label{fig:coinrun-phi-trace-sweep-B}
\end{figure}

\cref{tab:coinrun-results} reports the headline numbers for both CoinRun mechanism-error variants as a single (recovery, min-budget) pair per $(\text{variant}, N)$. \cref{fig:coinrun-phi-trace-sweep-A,fig:coinrun-phi-trace-sweep-B} unpack them into per-step $|\phi|$ traces, one figure per variant. Orange bars are flat $|\phi|/\max$ aggregated as mean $\pm$ 1 SE across the $10$ coalition-sampling seeds, with each per-seed trace taken from the smallest budget at which top-$K$ removal recovered the clean return (or, for seeds that never recovered within the $5\text{k}$ cap, from the $5\text{k}$ attempt). Blue bars are recursive $|\phi|/\max$ from the deterministic exact-enumeration run ($M = 100$ per round, $B = 2$). Red ticks mark the injected perturbation set.

\textbf{Recursive concentrates on the gateway steps at every $N$.}
For Variant A (\cref{fig:coinrun-phi-trace-sweep-A}) the perturbation set is exact ($p_A = 1.0$, so every reward-eligible terminal flips), and recursive's blue mass sits exactly on the injected steps at all four lengths: a single bar at step $37$ for $N \in \{100, 300\}$ and bars at $\{37, 364\}$ for $N \in \{500, 1000\}$. Recall is therefore $1.0$ on Variant A across the sweep and the full clean-buggy gap is closed with $\leq 2$ counterfactuals. Variant B (\cref{fig:coinrun-phi-trace-sweep-B}) is harder by design, the perturbation is stochastic ($p_B = 0.7$) and many short-jumps are absorbed by the policy. Recursive's blue mass concentrates on a strict subset of the realised perturbations: at $N = 1000$ only $2$ of the $10$ realised short-jumps carry non-zero recursive $\phi$, yet repairing those two alone restores the entire return. This is the same mechanism-vs-causation gap that the deterministic CenterGoal pixel map in \cref{fig:centergoal-pixelmap-10000} cleanly isolates: ground truth is mechanistic, recursive picks the gateway steps whose consequences cascade.

\textbf{Flat is dominated by sampling variance until $N$ exceeds the budget cap.}
At $N = 100$, flat does eventually localise the causal step on every coalition seed (the orange peak coincides with the blue bar in \cref{fig:phi-coinrun-A-N100,fig:phi-coinrun-B-N100}), but only after $620 \pm 114$ rollouts on Variant A and $810 \pm 165$ on Variant B, already $>20\times$ recursive's deterministic cost of $28$/$40$ rollouts. By $N = 300$ the per-seed minimum recovering budget for flat reaches $3800 \pm 611$ on Variant A and $1711 \pm 653$ on Variant B, the orange profile becomes visibly noisier (roughly uniform $\sim 0.15$ floor with the causal spike sitting only $\sim 5\times$ above it), and one of the ten seeds has already exhausted the $5\text{k}$ cap on Variant B. From $N = 500$ onwards the flat profile is essentially flat-noise on Variant A: max-normalised $|\phi|$ is uniform across the horizon and the seeds that did recover are dominated by lucky coalition draws rather than by a separated causal signal. Variant B at $N = 500, 1000$ shows the same pattern with $4/10$ and $2/10$ recovering, respectively, and the budget standard error is comparable to the mean, the few successful seeds are the upper tail of a heavy-tailed distribution, not a stable estimator of where the causal mass lies.

\textbf{Coalition-seed variance and the failure mode.}
Recursive's per-seed standard error is zero on these traces because the surviving sub-game collapses to $|\mathcal{A}|$ small enough that $2^{|\mathcal{A}|} \leq M = 100$ and exact enumeration kicks in within one or two rounds; from there the readout is deterministic up to the on-track coupling. Flat has no such collapse: the kernel-SHAP estimator must allocate its budget over all $T$ positions, and the variance grows with $T$. The most informative diagnostic is therefore not the average min-budget but the recovery rate, the fraction of the $10$ coalition seeds that ever recovered within the cap. That rate degrades from $10/10$ at $N = 100$ to $2/10$ at $N = 1000$ on Variant B (and to $0/10$ on Variant A, $N = 500$). The Variant B, $N = 1000$ panel is the cleanest illustration: $K = 9$--$10$ realised short-jumps are scattered across the $1000$-step trajectory, recursive concentrates blame on $2$ of them and recovers the full gap in $76$ rollouts, while flat's $5\text{k}$ cap recovers on only two seeds with a $\pm 2250$-rollout standard error on those two, a budget on the same order as recursive's, but with a $50\times$ larger price tag and a $20\%$ recovery rate.

\subsection{MiniGrid \texttt{CenterGoal}}
\label{sec:extended-centergoal-pixelmap}

\begin{figure}[ht]
    \centering
    \includegraphics[width=\linewidth]{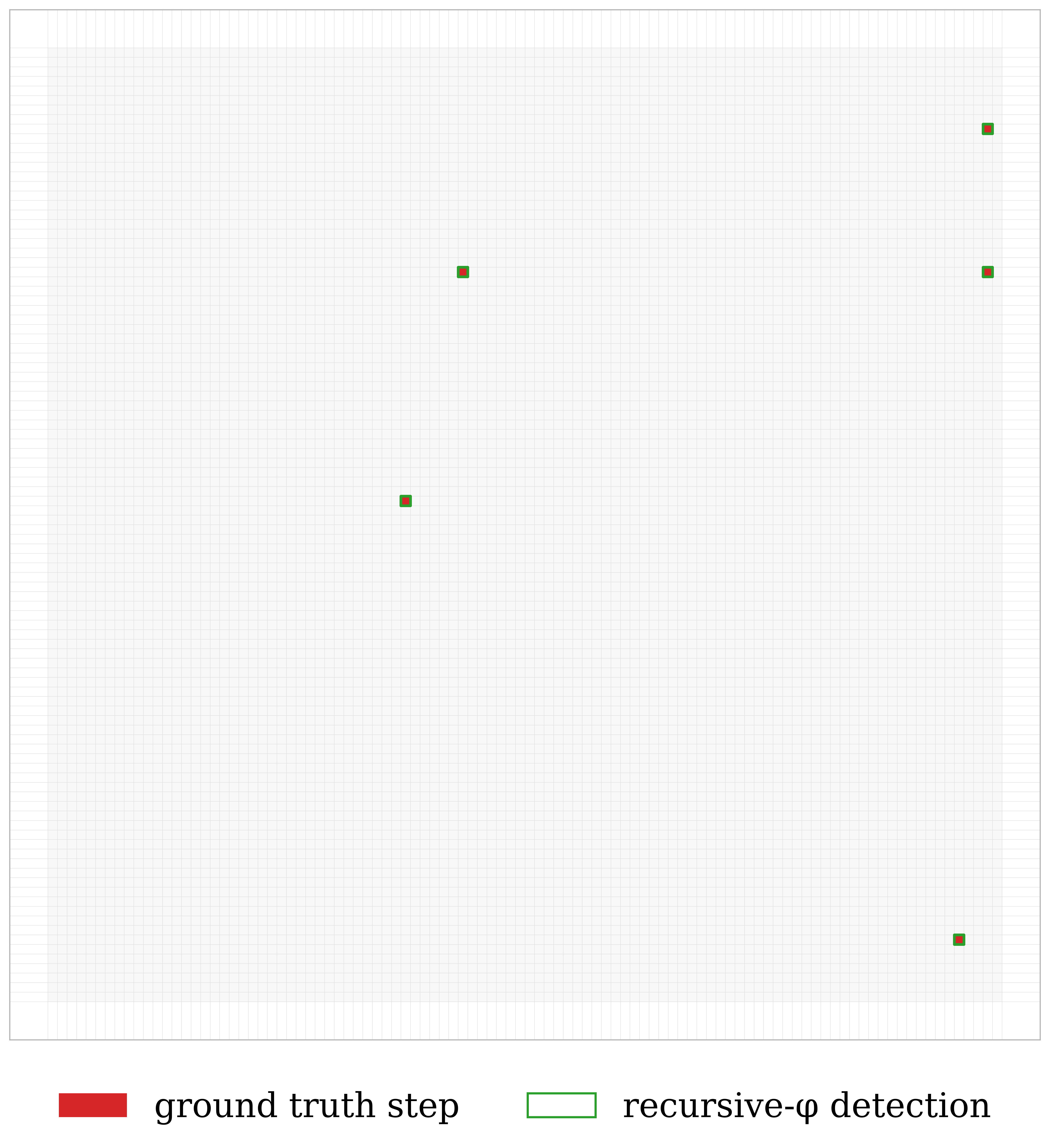}
    \caption{\textbf{\xadd{Pixel map of the planted steps} on \texttt{MiniGrid-CenterGoal5x5-v0} at $n_{\mathrm{cw}} = 10\,000$ ($K = 5$).} Step index $i$ is folded into a $100 \times 100$ grid (row $= i \mathbin{//} 100$, column $= i \bmod 100$)\xadd{. Red pixels mark the five planted penalty steps, and a green outline marks the steps of \textsc{Recursive}-$\phi$'s top-$K$}. \textsc{Recursive}-$\phi$ places non-zero mass on exactly the five planted penalty steps and zero everywhere else, so the recovered top-$K$ matches the planted set on $10/10$ coalition seeds whereas other baselines fail across all seeds at this length.}
    \label{fig:centergoal-pixelmap-10000}
\end{figure}

\cref{fig:phi-sweep-pixelmap-100} in the main paper shows the \xadd{planted steps and \textsc{Recursive}-$\phi$'s top-$K$ as a pixel map at $L = 100$ ($K = 5$)}, where every planted penalty step is visible at single-pixel resolution. \cref{fig:centergoal-pixelmap-10000} extends the same diagnostic to $n_{\mathrm{cw}} = 10\,000$ (so $T \approx 10^4$ once the BFS-to-goal tail is appended), the longest length at which a single-pixel-per-step layout still fits on a page. The step index $i$ is folded into a $100 \times 100$ grid (row $= i \mathbin{//} 100$, column $= i \bmod 100$) so contiguous stretches of the rollout appear as horizontal bands; the planted penalty steps are at the $K = 5$ positions sampled randomly.

The penalty mechanism in this experiment is fully deterministic: \texttt{TimePenaltyWrapper} subtracts a fixed $-0.1$ at exactly the planted set, the policy is a hand-crafted clockwise-then-BFS-to-goal walk, and the environment is reset under a fixed seed. Mechanistic ground truth therefore equals causal ground truth here: every planted step is causally responsible for the corresponding $-0.1$ in the return, and recursive's top-$K$ should coincide exactly with the planted set, not merely overlap with it. \cref{fig:centergoal-pixelmap-10000} confirms this: \xadd{\textsc{Recursive}-$\phi$'s top-$5$ in the $10\,000$-step map are exactly the five planted positions, and its $\phi$ is} nearly zero \xadd{on the rest of the grid}. This is the cleanest demonstration of the localisation property at $L \approx 10^4$\xadd{:} the \xadd{coalitions} \textsc{Recursive}-$\phi$ \xadd{actually uses} to recover the entire causal set \xadd{only about double from $L = 100$ ($3860$ to $8148$ at $K = 5$) for a $100\times$ longer trajectory} (\cref{fig:phi-sweep-budget}), because pruning collapses the \xadd{$9\,997$} non-causal steps within the first few rounds and the surviving sub-game is small enough for exact enumeration. None of the flat baselines (\textsc{Flat}-$\phi$, SHAP, LIME) reaches this regime within the largest budget we sweep ($M = 10^5$)\xadd{. By $L = 5000$ their gap-explained curves in \cref{fig:phi-sweep-recovery} fall to at most $10\%$: \textsc{Flat}-$\phi$ recovers on $1/10$ seeds at $K = 1$ and on none at $K \in \{2, 5\}$, SHAP on none, and LIME only at $K \in \{1, 2\}$ on the one seed whose rollout seeds include that of the observed trajectory}.

\subsection[Learned world models]{\xadd{Learned world models}}
\label{sec:extended-learned-wm}

\xadd{This section reports the results on learned world models that are not in \cref{sec:results-learned-wm}. All of them come from the runs described in \cref{sec:exp design details}.}

\begin{table}[ht]
    \xaddcolor
    \centering
    \small
    \caption{\xadd{\textbf{World-model optimism on Pong.} Predicted minus real return of DIAMOND's policy over $48$ seeds (mean $\pm$ std), and the number of seeds on which the model over-predicts; it under-predicts on none.}}
    \label{tab:pong-wm-gap}
    \vspace{2pt}
    \begin{tabular}{lcccccc}
        \toprule
        Steps & $100$ & $200$ & $500$ & $1000$ & $1500$ & $2000$ \\
        \midrule
        Gap (points) & $0.6 \pm 0.9$ & $1.6 \pm 1.4$ & $5.8 \pm 2.4$ & $11.9 \pm 3.5$ & $18.0 \pm 4.3$ & $23.9 \pm 5.0$ \\
        Seeds over-predicted & $16$ & $32$ & $47$ & $47$ & $48$ & $48$ \\
        \bottomrule
    \end{tabular}
\end{table}

\begin{table}[ht]
    \xaddcolor
    \centering
    \footnotesize
    \setlength{\tabcolsep}{3.5pt}
    \caption{\xadd{\textbf{Repair on learned world models under each ranking.} \emph{Injections}: smallest number of top-ranked steps (Pong) or blocks (Cube) whose injection reproduces the real outcome. \emph{Sparse}: rollouts repaired by fewer than $10$ steps or $5$ blocks. \emph{Top-1}: rollouts repaired by the top-ranked injection alone. \emph{Gap closed}: $100\,(1 - |\hat{Y}_k - Y| / |\hat{Y}_0 - Y|)$, where $Y$ is the real outcome and $\hat{Y}_k$ the one predicted after the top-$k$ injections, so an overshoot counts against. Entries are mean $\pm$ std over rollouts.}}
    \label{tab:learned-wm-topk}
    \vspace{2pt}
    \begin{adjustbox}{max width=\linewidth}
    \begin{tabular}{lccccccc}
        \toprule
        & & & & \multicolumn{4}{c}{Gap closed by the top-$k$ injections (\%)} \\
        \cmidrule(lr){5-8}
        Ranking & Injections $\downarrow$ & Sparse $\uparrow$ & Top-1 $\uparrow$ & $k = 1$ & $k = 2$ & $k = 5$ & $k = 10$ \\
        \midrule
        \multicolumn{8}{l}{\emph{DIAMOND on Pong, $14$ rollouts of $100$ steps}} \\
        \quad $\hat{\phi}$, direction of the gap ($M = 1500$) & $2.7 \pm 1.7$ & $14$ & $3$ & $42 \pm 39$ & $74 \pm 39$ & $93 \pm 18$ & $100 \pm 0$ \\
        \quad $\hat{\phi}$, magnitude ($M = 1500$) & $5.4 \pm 4.7$ & $12$ & $2$ & $32 \pm 37$ & $43 \pm 62$ & $63 \pm 46$ & $88 \pm 31$ \\
        \quad Frame prediction error & $6.0 \pm 5.2$ & $12$ & $3$ & $29 \pm 43$ & $40 \pm 45$ & $65 \pm 41$ & $82 \pm 54$ \\
        \quad Latent prediction error & $7.3 \pm 3.5$ & $11$ & $0$ & $4 \pm 13$ & $1 \pm 43$ & $38 \pm 59$ & $93 \pm 18$ \\
        \midrule
        \multicolumn{8}{l}{\emph{LeWM on Cube, $45$ episodes of $10$ blocks}} \\
        \quad $\phi$, magnitude (exact) & $5.2 \pm 2.0$ & $18$ & $2$ & $-1 \pm 132$ & $-3 \pm 140$ & $50 \pm 75$ & $100$ \\
        \quad $\phi$, direction of the gap (exact) & $7.0 \pm 2.8$ & $10$ & $2$ & $20 \pm 33$ & $17 \pm 65$ & $28 \pm 94$ & $100$ \\
        \quad Latent prediction error & $6.4 \pm 2.1$ & $9$ & $0$ & $-62 \pm 199$ & $-63 \pm 200$ & $20 \pm 109$ & $100$ \\
        \bottomrule
    \end{tabular}
    \end{adjustbox}
\end{table}

\xadd{\textbf{Optimism.} \cref{tab:pong-wm-gap} extends the Pong gap statistics: DIAMOND's over-prediction grows roughly linearly with the horizon, about $12$ points per $1000$ steps, and it never under-predicts. On Cube, the planner fails on $23$ of the $30$ start/goal pairs, and LeWM's imagined outcome is a success for $21$ of them. For the six most optimistic episodes, the imagined goal cost is below the real one on average at every block boundary. The difference is largest at the end of each $25$-step plan ($-264 \pm 67$ at step $25$ and $-295 \pm 63$ at step $50$).}

\xadd{\textbf{Repair.} \cref{tab:learned-wm-topk} reports the repair experiment under both rankings of the attributions. On Pong, ranking in the direction of the gap repairs the $14$ rollouts with $1, 1, 1, 2, 2, 2, 2, 2, 2, 3, 4, 4, 5, 7$ injections, and ranking by magnitude with $1, 1, 2, 2, 2, 2, 4, 4, 5, 6, 7, 9, 15, 15$. On Cube, the magnitude ranking repairs $2$ episodes with one block, $1$ with two, $4$ with three, and $11$ with four, i.e., $18$ of $45$ with fewer than five ($3.3 \pm 1.0$ blocks on these); the other $27$ need five to nine.}

\xadd{\textbf{The two rankings.} Following which ranking to repair is better differs between the models: the direction of the gap needs fewer injections than the magnitude on $8$ Pong rollouts and more on $1$, but more on $23$ Cube episodes and fewer on $2$. We think this is due to the interaction effects of the errors. Once one step is fixed, the causal contribution of other errors detected in the original trajectory also changes. Thus, we recommend to try to fix the top detected steps by both ranks.}

\xadd{\textbf{Baselines.} On Pong, the direction-of-gap ranking needs fewer injections than the frame-error ranking on $9$ rollouts, more on $3$, and as many on $2$ (two-sided sign test, $p = 0.15$), and fewer than the latent-error ranking on $12$ and more on $1$ ($p = 0.003$); for the magnitude ranking these counts are $8$, $4$, $2$ and $11$, $2$, $1$. On Cube, the magnitude ranking needs fewer blocks than the latent-error ranking on $24$ episodes, more on $12$, and as many on $9$ ($p = 0.065$); the direction-of-gap ranking does not improve on the latent-error ranking (fewer on $16$, more on $25$). The latent-error ranking puts the first block on top in $22$ of the $45$ episodes: the first prediction is made from a single context frame and has the largest one-step error on average.}

\xadd{\textbf{The two cases of \cref{fig:learned-wm-cases}.} In the Pong case, $\hat{\phi}$ is non-zero on steps $46$--$67$, and the six top-ranked steps in the direction of the gap, which tie, are $61$--$66$, the steps leading to the imagined point. Each of them restores the real score when injected alone, as do $12$ of the $22$ flagged steps, against $1$ of the other $78$. The frame-error ranking puts step $70$ on top, four steps after the imagined point, where the model draws the agent's own paddle a few pixels off; the latent-error ranking puts step $4$ on top, the start of the rally, at which the ball and the opponent's paddle appear. Both are genuine one-step mispredictions, but injecting either leaves the predicted score at $+1$, and the two rankings need $9$ and $10$ injections. Along the real trajectory, the one-step frame error at steps $61$--$66$ is $2$ to $5$ times its median, against $14$ times at step $70$: the false point arises in the free rollout, where the imagined game has drifted from the real one, and a one-step error computed from the real context does not see it. In the Cube case, the model transition at step $35$ carries $-265$ of a total attribution of $-382$. The latent-error ranking puts the first block on top; injecting it moves the predicted goal cost from $42$ to $293$, and five more blocks are needed for a repair. The attribution ranks this block third. LeWM has no decoder, so the world-model frame in \cref{fig:learned-wm-cases}(b) is the real frame whose latent is nearest to the imagined one.}

\xadd{\textbf{Budget and error bound.} On Cube at $M = 1500$ the four rounds use $4$, $16$, $256$, and $1024$ coalitions, i.e., all of them are enumerated: the attributions sum to the gap on all $45$ episodes, and they equal the flat Shapley values on the ten episodes of the first pass, for which we computed those. At $M = 300$ the last round is sampled; the largest deviation from the exact values ranges from $19$ to $259$ goal-cost units across these ten episodes, and $10$ instead of $18$ of the $45$ episodes are repaired by fewer than five blocks. With $M = 30$ coalitions per round, the mean absolute error is $20.0 \pm 12.3$ against a mean $|\phi_t|$ of $20.2$. Taking, per episode, the largest per-round block estimation error as $\alpha$ and the largest per-round interaction as $\eta$ gives $\alpha = 24.8$ and $\eta = 40.6$ on average, so the bound in part (c) of \Cref{thm:recursive-correctness} holds on all ten episodes, but it is loose: interactions between blocks are of the same order as the attributions themselves, unlike in the controlled settings, where blocks without a causal step are exact null players. A sparse repair on a learned world model therefore calls for a per-round budget large enough to enumerate the surviving blocks.}

\xadd{\textbf{Stopping rule.} The stopping rule of \Cref{sec:faq}, which takes top-ranked steps until their signed sum reaches $90\%$ of $\sum_t \hat{\phi}_t$, works well on Cube but on Pong due to zeroing out some $\phi$-values, it does not apply as the sampled estimates do not sum to the gap.}

\clearpage
\subsection[Case studies]{\xadd{Case studies}}
\label{sec:extended-case-studies}

\begin{figure}[H]
    \centering
    \includegraphics[width=\linewidth]{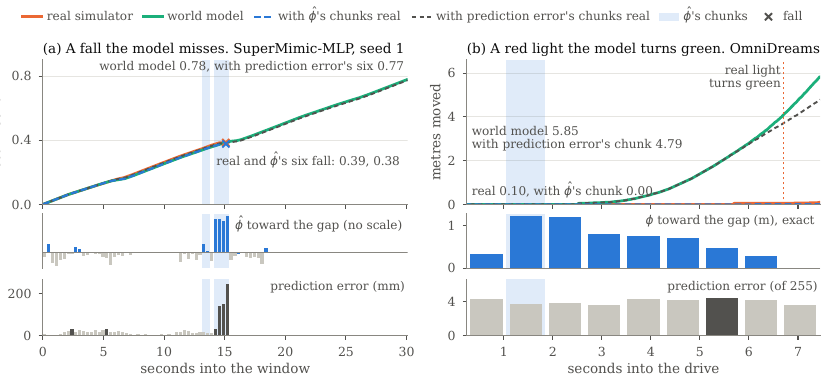}
    \caption{\xadd{\textbf{The humanoid and driving case studies.} Top: the return so far over the $30$\,s window in (a), where a line stops at a fall and the scoring stops, and the metres moved along the logged path from $1$\,s in (b). Below: $\phi$ per chunk in the direction of the gap (gray against it), sampled in (a) and exact in (b), and the world model's prediction error from the real state. Shaded chunks are the ones injected for the $\phi$ rollout, and dark bars the ones injected for the prediction-error rollout: six each in (a), one each in (b). Prediction errors after the real fall in (a) are excluded.}}
    \label{fig:humanoid-driving-cases}
\end{figure}

\xadd{Additionally, we studied debugging four rollouts from learned world models with our algorithm one by one: the two of \cref{fig:learned-wm-cases}, one on a simulated humanoid, and one in autonomous driving (\cref{tab:case-studies,fig:humanoid-driving-cases}). Videos of all four are on the project page, \href{https://mingxuan.me/debugwm.html}{\texttt{mingxuan.me/debugwm.html}}. We chose each rollout by hand, by the criteria given below, so they show how a world model can fail and how the attributions respond.}

\begin{table}[ht]
    \xaddcolor
    \centering
    \footnotesize
    \setlength{\tabcolsep}{3.5pt}
    \caption{\xadd{\textbf{Four case studies.} \emph{Rollouts}: counterfactual rollouts used by \textsc{Recursive}-$\phi$, or all $2^9$ coalitions for the exact $\phi$ in driving. \emph{Real} and \emph{Imagined}: the outcome, which is the return over its maximum on the humanoid and the points on Pong, and the metres driven at a red light in driving and the goal cost on Cube, where lower is better. The last three columns give the smallest number of top-ranked players whose injection repairs the rollout ($\downarrow$), with the number of attributions needed to fix ranked in the direction of the gap or by magnitude.}}
    \label{tab:case-studies}
    \vspace{2pt}
    \begin{adjustbox}{max width=\linewidth}
    \begin{tabular}{lccccccc}
        \toprule
        Case & Steps/Chunks & Rollouts lengths & Real & Imagined & Direction & Magnitude & Prediction error \\
        \midrule
        SuperMimic-MLP, missed fall & $94$ & $16\,108$ & $0.39$, falls & $0.78$ & $\mathbf{10}$ & $11$ & $21$ \\
        OmniDreams, red light & $9$ & $512$ & $0.10$ & $5.85$ & $\mathbf{1}$ & $\mathbf{1}$ & $6$ (frame) \\
        DIAMOND, Pong & $100$ & $3356$ & $0$ & $+1$ & $\mathbf{1}$ & $5$ & $9$ (frame), $10$ (latent) \\
        LeWM, Cube & $10$ & $1300$ & $424$ & $42$ & $\mathbf{1}$ & $\mathbf{1}$ & $6$ (latent) \\
        \bottomrule
    \end{tabular}
    \end{adjustbox}
\end{table}

\xadd{\textbf{Humanoid setup.} NDM Gym~\citep{ndmgym2026} trains tracking policies for the Unitree G1 on LAFAN1 motions~\citep{harvey2020robust} by backpropagating through a learned world model, following SuperMimic~\citep{liao2026supermimic}, and evaluates them in Isaac Lab with MuJoCo-Warp. SuperMimic has no official code release, so the results below concern NDM Gym's implementation and its published checkpoints. Each checkpoint holds a policy and its world model, and we use those with an MLP world model (SuperMimic-MLP). A window is $1504$ control steps ($30$\,s at $50$\,Hz) from a state the simulator reached, divided into $94$ players of $16$ steps ($0.32$\,s). Each player is played by the simulator or by the world model, and the policy acts in closed loop on whichever state results. The outcome is the undiscounted tracking reward over its maximum. Both worlds use NDM Gym's analytic reward and termination, so no outcome model is learned. An episode ends when the pelvis strays $0.25$\,m from the reference height or tilts too far from the reference, and scoring stops there. We call this a fall.}

\xadd{\textbf{Attribution and repair on the humanoid.} We use a maximum budget per round of $M = 5000$ coalitions, $B = 2$, and $\gamma = 1$, which gives seven rounds in total. The last rounds are sampled. We score a repair for every $k = 1, \ldots, 94$ given the $\phi$-value results. The baseline ranks chunks by the world model's error over one chunk: from the real state, the model rolls out open-loop through the executed actions for $16$ steps, and the chunk is scored by the mean position error of the tracked bodies. Chunks after a real fall are excluded, since the recording is chaotic by then. A rollout near a fall also depends on numerical rounding, which changes with the batch size, so we replayed the sets named below in $128$ or $256$ replays.}

\xadd{\textbf{SuperMimic-MLP: a fall the model misses.} Over $1691$ rollouts clips we tested, the median absolute transfer gap is $0.004$, yet imagination survives $23$ of the $28$ clips that contain a real fall. These clips cover five falls. We attributed all five with the fall at mid-clip: the direction of the gap needs $5$ to $14$ chunks (median $11$), the magnitude $10$ to $49$ (median $17$), and prediction error $20$ to $59$ (median $22$).} 

\xadd{We take a fall clip that happens constantly through multiple replays for analysis. The robot sits on the floor and starts to rise at about $14.6$\,s. In the simulator it tips over backwards at about $15.1$\,s, with a return of $0.39$. In imagination it stands up and keeps tracking, with a return of $0.78$. The ten top-ranked chunks by $\phi$ close the gap cleanly. For naive prediction error baselines, chunks $44$--$47$ have the four largest prediction errors, up to $250$\,mm, and over the last three the real and predicted motions look different. Yet injecting only these four real chunks closes merely $3\%$ of the gap, and the world model still hallucinates that the robot stands up in all $256$ replays. The real chunks to blame as detected by $\phi$, $41$ and $42$, have prediction errors of $16$ and $14$\,mm, which only rank $25$th and $28$th of $94$ under the prediction-error ranking.}

\xadd{\textbf{Driving setup.} Alpamayo~1.5~\citep{nvidia2025alpamayo} drives in closed loop in AlpaSim~\citep{alpasim2025} from one front camera. The real world is NuRec~\citep{nurec2025}, a neural reconstruction of a logged drive that renders the camera from wherever the car goes. The world model is OmniDreams~\citep{basant2026omnidreams}, which generates each frame from the HD map, the car's path, the other road users, and the frames before it. Only the frames differ between the two worlds. A window is nine chunks of $0.8$\,s, from $0.25$ to $7.45$\,s, and we compute $\phi$ exactly over all $2^9 = 512$ coalitions. The baseline ranks chunks by the mean pixel difference between NuRec's frames and OmniDreams' predictions of them, made $0.27$\,s at a time from the real state. A drive counts as repaired once it moves no further than the real one.}

\xadd{\textbf{OmniDreams: a red light turned green.} At night the car waits at a red light, which NuRec renders red until $5.5$\,s and green by $6.7$\,s. AlpaSim's score has no term for red lights, so the outcome is the distance driven along the logged path from $1$\,s to $7.45$\,s. On NuRec's frames Alpamayo waits, moving $0.09$\,m on average over $10$ seeds. On OmniDreams' frames the red washes out within a second and is green by $2$\,s, and Alpamayo drives off, moving $3.28$\,m on average. We study the seed with the largest gap, $0.10$\,m against $5.85$\,m. The top chunk by $\phi$, $1.05$--$1.85$\,s, repairs the drive alone: with NuRec's frames there, OmniDreams ends on amber instead of green and the car does not move. By prediction error this chunk is the third least wrong of nine ($3.7$ against $3.6$ to $4.4$ out of $255$), since predicted from the real state the light stays red. It turns green only once OmniDreams draws from its own frames. Prediction error's top chunk, $5.05$--$5.85$\,s, comes three seconds after the imagined light turned green and leaves the car moving $4.79$\,m, and it takes six chunks to repair the drive. NuRec is itself a reconstruction, so the gap is between two models of the same drive. The project page shows the frames and Alpamayo's reasoning at each decision.}

\xadd{\textbf{DIAMOND on Pong: an imagined point.} We use a rollout that a single top-ranked step closes the transfer gap. In this clip, nobody scores within $100$ steps, and the opponent returns the ball at $t = 64$. But in imagination the ball passes the opponent at $t = 66$ and the agent is awarded a point. \cref{sec:extended-learned-wm} gives the attributions and baselines of this rollout. We also note that the repair restores the score but not every frame: after $t = 64$ the repaired rollout loses track of the ball instead of drawing the ball bouncing back while the score is correct.}

\xadd{\textbf{LeWM on Cube: an imagined grasp.} The episode we pick starts with the cube in the gripper. In the simulator the gripper opens within the first few steps, the cube land on the table, and it is never picked up again, so the final goal cost is $-424$. LeWM hallucinates that it picks the cube up again: its predicted final cost is $-42$, below the success threshold of $-331$, and its final latent is nearest to a frame with the cube in the gripper. The exact attributions are largest at step $35$ ($-265$), step $25$ ($-89$), and step $0$ ($-58$), fixing those close the transfer gap. The latent-error ranking puts step $0$ first, after which the goal cost is increased to -292 closer to real, but then it picks step $181$ and $13$ which contribute nothing to closing the gap.}

\xadd{\textbf{Across the cases.} In all four, the top picks of prediction error are genuine mispredictions that do not close the transfer gap. We argue that this is due to the error compounding effect of learned world models. The real cause of the final failure usually starts with a small misprediction which is nearly negligible under the prediction error metric or even manual checks. And those big prediction error steps are usually ones too close to the failure or even after the failure. In such cases, only \textsc{Recursive}-$\phi$ finds the easily overlooked root causes.}

\section{Extended Related Work}
\label{sec:extended-related-work}

Existing approaches to diagnosing world-model failures are largely indirect.
Aggregate metrics measure reconstruction loss or reward prediction error across trajectories \citep{schrittwieserMasteringAtariGo2020,dreamer}; short imagined rollouts from real data quantify how model errors compound over multi-step rollouts \citep{jannerWhenTrustYour2019a}.
Saliency maps, structural causal models, and Shapley decompositions over state features explain the agent's policy at a given state but do not attribute blame to the learned dynamics over a rollout (see \emph{Explaining RL agents} below).
None of these lines answers the targeted question: at which time steps did the world model's prediction errors cause the policy to underperform?
Most directly, \citet{li2026credit} introduce counterfactual Shapley values for trajectories with a flat coalition estimator; we extend their construction with a recursive coarse-to-fine algorithm that scales from $L \le 300$ to $L \approx 10^6$.
Given a fixed policy and a single observed failure trajectory, our method localizes which transitions in the world model are causally responsible for the return gap; we are unaware of prior work that addresses this question for a learned world model under a fixed policy.

\textbf{Feature attribution.}
Shapley values \citep{shapley:book1952} provide a principled way to distribute credit among players in a cooperative game.
In machine learning, SHAP \citep{lundberg2017unified} instantiates this idea for individual predictions, treating input features as players and a trained model as the characteristic value function.
LIME \citep{ribeiro2016should} fits a sparse local linear surrogate around each prediction rather than invoking the cooperative-game framework.
Neither method was designed to attribute credit in multi-step rollouts, where an error at step $t$ propagates through subsequent transitions.
Our method treats trajectory time steps as players and uses counterfactual simulation as the characteristic value function, so attributions reflect causal responsibility for the return gap rather than correlational feature importance.

\textbf{Explaining RL agents.}
A growing body of work seeks to explain the behavior of trained RL agents.
Perturbation-based saliency maps highlight which observation regions drive an agent's action choice \citep{DBLP:conf/icml/GreydanusKDF18visatari}.
Approaches that learn a causal graph during RL use it to generate counterfactual explanations of agent actions \citep{DBLP:conf/aaai/Madumal0SV20}.
Shapley-based methods apply the cooperative-game framework to attribute an agent's action choices, expected return, or predicted return to state features \citep{explaining_rl_with_shapley_values}.
All three lines target the policy's action selection at a state and surface either correlations or local causal links at a given state, rather than attributing blame to the learned dynamics over a rollout.
We instead attribute return loss to specific transitions of the world model, asking which transitions, not which features, caused the failure.

\textbf{Model-based RL and world models.}
Model-based RL agents learn a world model to reduce sample complexity and enable planning \citep{suttonDynaIntegratedArchitecture1991,hafner2018learning,schrittwieserMasteringAtariGo2020,dreamer}.
World models have achieved strong results across domains: \citet{Alonso_2024} demonstrate that a diffusion-based world model achieves state-of-the-art Atari 100k performance among agents trained entirely within a world model; for a broad survey of world models in robot learning, see \citet{hou2026world}.
A well-known failure mode is \emph{model exploitation}: errors in the learned dynamics compound over multi-step rollouts and steer the policy toward spurious high-reward regions \citep{jannerWhenTrustYour2019a}.
Recent work trains more robust world models by targeting data collection at environments where world model error is highest, guided by reward-free curricula \citep{rigter2023rewardfree}.
\citet{gzel2025imagined} train agents entirely inside a learned world model with an adaptive autocurriculum over imagined initial states, generalizing to novel settings without real-environment interaction.
These lines all aim to build world models that work better on average; none localizes which transitions in a specific failure episode are causally responsible for the return gap.
Most closely, \citet{Buesing2018Wouldacoulda} counterfactually simulates alternative actions under shared random seeds to score policies inside a learned model.
Prior work either mitigates model error uniformly, by limiting rollout length or reweighting model-based and model-free updates, or counterfactually scores policies rather than transitions; neither line localizes which specific transitions caused the underperformance.
Holding the policy fixed, we identify which specific transitions in the world model's rollout are causally responsible for the return gap, providing targeted feedback for model improvement.

\textbf{Sim-to-real transfer.}
Sim-to-real transfer methods train policies in simulation and deploy them in the real world, relying on domain randomization over visual or dynamic parameters \citep{tobin2017domain,peng2017simtoreal,akkaya2019solvingcube} to bridge the simulator-real gap.
Teacher--student pipelines extend domain randomization by distilling privileged-state policies into visual ones under aggressive rendering randomization, enabling zero-shot transfer for humanoid loco-manipulation \citep{xue2025opening}.
A related line uses structured benchmarks with explicit causal factors to study transfer under controlled interventions \citep{ahmedCausalWorldRoboticManipulation2021}.
A separate line learns the simulator entirely: \citet{ha2018world} train policies inside a learned world model, \citet{wu2022daydreamer} extend this to physical robots, and \citet{bruce2024genie,yang2023learning} scale learned interactive simulators to internet-scale video data.
Domain randomization and learned simulators both target aggregate transfer: robustness over a distribution of environments, not diagnosis of any specific failure episode.
Our method is complementary: given a specific failure episode, it pinpoints which simulator transitions caused the return gap, supporting targeted simulator improvement rather than aggregate robustness.

\textbf{Safe and robust RL.}
Safe RL incorporates constraints alongside reward maximization \citep{garcia2015comprehensive}; \citet{huang2023safedreamer} integrate Lagrangian-based safety constraints into a world model, either via per-step online planning (OSRP) or background policy optimization over imagined rollouts (BSRP).
Robust RL trains policies that are tolerant to a family of model perturbations on transitions \citep{wiesemann2013robust,panaganti2022robust} or on observations \citep{zhang2020robustdrl}.
Both safe and robust RL are prescriptive and online: they aim to prevent failures during deployment by optimizing over constraints or perturbation families, whereas our method is a post-hoc diagnostic applied to a fixed policy after an observed failure.
SafeDreamer specifically is distinguished from our work by its goal: it adapts the policy at deployment time to satisfy constraints, while we hold the policy fixed and attribute the return gap to specific simulator transitions.
Our diagnostic is complementary: by identifying exactly which transitions drove a failure, it can directly inform what perturbation families robust RL should model, or which transition types require additional safety data.

\section{Theory Details}
\label{sec:theory-details}
In this section, we provide detailed discussion on assumptions, definitions and proofs.

\subsection{World Model Transfer Gap and Causal Attribution}
\label{sec:phi-values-causal}
Below is the formal version of the properties that a causal attribution for the world model transfer gap should satisfy (\Cref{def:causal world-model attribution}).

\begin{definition}[Attribution for World Model Transfer Gap, Formal]
    \label{def:causal world-model attribution formal}
    A world-model transfer gap attribution function $\phi$ maps a trajectory, a sequence of dynamics selectors, a world model, and a policy to a real vector denoting each step's attribution to the transfer gap,
    \begin{align}
        \phi: \tau \times \{{\mathrm{wm}}, {\mathrm{env}}\}^T \times \mathcal{M}_{\mathrm{wm}} \times \Pi \times [T] \mapsto \mathbb{R}^T
    \end{align}
    where $\{{\mathrm{wm}}, {\mathrm{env}}\}^T$ is the sequence of dynamics selectors determining whether at time step $t$, $\{f_S, r\}$ is from $\mathcal{M}_{\mathrm{wm}}$ or $\mathcal{M}_{\mathrm{env}}$.
    Each desideratum below is a mapping $D_i: \Omega \times \mathbb{C} \times \Phi \to \{0,1\}$, where $\Omega$ is the space of SCMs (each $\mathcal{M} \in \Omega$ packaging a world-model/environment pair $(\mathcal{M}_{\mathrm{wm}}, \mathcal{M}_{\mathrm{env}})$ with a trajectory $\tau$ and a policy $\pi$), $\mathbb{C}$ is the space of causal measures, and $\Phi$ is the space of attribution functions. The attribution function $\phi$ is causal if, for every step $t \in [T]$, $D_i(\mathcal{M}, c, \phi) = 1$ holds for $i = 1, \dots, 4$:
    \begin{enumerate}[label={$\mathbf{D}_\arabic*$}, leftmargin=20pt, topsep=0pt, parsep=0pt, itemsep=4pt]
        \item \textbf{Causal Admissibility}: Steps that do not cause the performance gap have zero credit:
        \begin{align}
            c = 0 \;\implies\; \phi = \mathbf{0}.
        \end{align}
        \item \textbf{Causal Power}: Steps that cause the performance gap are assigned non-zero credit:
        \begin{align}
            \exists\, \mathbf{u}, z:\; c \neq 0 \;\implies\; \phi \neq \mathbf{0}.
        \end{align}
        \item \textbf{Causal Normality}: Steps that are more likely to affect the outcome have more credit:
        \begin{align}
            c_\mathcal{M} = c_{\mathcal{M}'} \;\land\; P^\mathcal{M}_{\bar{c}}(Z) \not\equiv P^{\mathcal{M}'}_{\bar{c}}(Z) \;\land\; P^\mathcal{M}_{+}(z) \geq P^{\mathcal{M}'}_{+}(z) \;\implies\; \phi_\mathcal{M} > \phi_{\mathcal{M}'}.
        \end{align}
        \item \textbf{Causal Effect Scaling}: Steps affecting the outcome more are assigned more credit:
        \begin{align}
            c_\mathcal{M} \geq c_{\mathcal{M}'} \;\land\; P^\mathcal{M}(Z) \equiv P^{\mathcal{M}'}(Z) \;\land\; P^\mathcal{M}(c_Z) \not\equiv P^{\mathcal{M}'}(c_Z) \;\implies\; \phi_\mathcal{M} > \phi_{\mathcal{M}'}.
        \end{align}
    \end{enumerate}
    Quantification: all premises universally quantified unless marked $\exists$.
    Notation: $c$, $c_\mathcal{M}$ abbreviate $c(\mathcal{M}, \mathbf{u}, X, Y, z)$ --- the local discrepancy between $\mathcal{M}_{\mathrm{wm}}$ and $\mathcal{M}_{\mathrm{env}}$ at step $t$ as seen through its downstream effect on the return $Y$;
    $\phi$, $\phi_\mathcal{M}$ abbreviate $\phi_t(\mathcal{M})$, the per-step attribution at $t$;
    $P \equiv P'$ denotes distributional equality;
    $P^\mathcal{M}_{\bar{c}}(Z) \coloneqq P^\mathcal{M}(Z \mid c \neq 0)$;
    $P^\mathcal{M}_{+}(z) \coloneqq \mathrm{sign}(c_\mathcal{M}) \cdot P^\mathcal{M}(z)$.
\end{definition}

\begin{proof}[Proof for \Cref{thm:phi is causal}]
    We show that the $\phi$-values under our transfer gap attribution setting with NTE $f(\mathbf{z}) = Y^\tau_{\mathcal{M}_{\mathrm{env}}} - Y_{\sigma_{\mathbf{z}}}$ from \cref{eq:ctf nte wm,eqn:sigma-z} satisfies each of $\mathbf{D_1}$--$\mathbf{D_4}$ in \Cref{def:causal world-model attribution formal}.

    \emph{Setting.} The ``players'' are the per-step dynamics selectors: a coalition $\mathbf{z} \in \{0,1\}^T$ specifies, for each step $t$, whether the transition $(x_t, r_t, s_{t+1})$ in the counterfactual rollout is drawn from the world model $\mathcal{M}_{\mathrm{wm}}$ (when $z_t = 1$) or copied from the observed real-environment trajectory (when $z_t = 0$). The ``causal measure'' $c \coloneqq c(\mathcal{M}, \mathbf{u}, X_t, Y, z)$ is the functional dependence of the return $Y$ on this selector at step $t$ --- equivalently, the local discrepancy between $\mathcal{M}_{\mathrm{wm}}$ and $\mathcal{M}_{\mathrm{env}}$ at $t$ as seen through its downstream impact on the return. The ``SCM'' $\mathcal{M}$ here packages the pair $(\mathcal{M}_{\mathrm{wm}}, \mathcal{M}_{\mathrm{env}})$ together with the trajectory $\tau$, the policy $\pi$, and the exogenous draws $(\mathbf{u}, \mathbf{u}')$ in \cref{eq:ctf nte wm}; the ``baseline distribution'' $P^\mathcal{M}(Z)$ is the corresponding distribution over counterfactual contexts (i.e., the off-track rollouts induced by $\mathbf{u}'$ and $\pi$ under $\mathcal{M}_{\mathrm{wm}}$).

    \textbf{D1 (Causal Admissibility).}
    Suppose $c = 0$: at step $t$, the world model is functionally indistinguishable from the real environment for $Y$ across every context $(\mathbf{u}, z)$. Then flipping $z_t$ between $0$ and $1$ leaves the counterfactual return unchanged, $Y_{\sigma_{\mathbf{z} \cup \{t\}}} = Y_{\sigma_{\mathbf{z}}}$ for every coalition $\mathbf{z}$, so every marginal contribution vanishes:
    $f(\mathbf{z} \cup \{t\}) - f(\mathbf{z}) = Y_{\sigma_{\mathbf{z}}} - Y_{\sigma_{\mathbf{z} \cup \{t\}}} = 0.$
    By the Shapley null-player axiom, $\phi_t = 0$, so a step at which the world model agrees with the real environment receives no credit for the transfer gap.

    \textbf{D2 (Causal Power).}
    Suppose there exist exogenous draws $\mathbf{u}$ and a context $z$ such that $c(\mathcal{M}, \mathbf{u}, X_t, Y, z) \neq 0$, i.e., for some configuration of the other steps, swapping $\mathcal{M}_{\mathrm{env}}$ for $\mathcal{M}_{\mathrm{wm}}$ at step $t$ shifts the return. Then there is a coalition $\mathbf{z}^*$ (matching that context) with nonzero marginal contribution $f(\mathbf{z}^* \cup \{t\}) - f(\mathbf{z}^*) \neq 0$. The Shapley weights $w(\mathbf{z}) = \frac{|\mathbf{z}|!(T - |\mathbf{z}| - 1)!}{T!}$ are strictly positive for every $\mathbf{z}$, so
    \[
        \phi_t = \sum_{\mathbf{z}: z_t = 0} w(\mathbf{z})\bigl[f(\mathbf{z} \cup \{t\}) - f(\mathbf{z})\bigr]
    \]
    contains a strictly weighted nonzero term. Under \Cref{assum:sign-consistency} (causal steps share a sign, so marginal contributions cannot systematically cancel across coalitions), this entails $\phi_t \neq 0$: any step at which the world model can move the return relative to the real world is given nonzero credit.

    \textbf{D3 (Causal Normality).}
    Consider two world-model/environment pairs $\mathcal{M}, \mathcal{M}'$ with identical local discrepancies at step $t$ ($c_\mathcal{M} = c_{\mathcal{M}'}$) but different baseline distributions over counterfactual contexts, with $P^\mathcal{M}_{+}(z) \geq P^{\mathcal{M}'}_{+}(z)$ (the baseline under $\mathcal{M}$ assigns weakly higher probability to contexts in which the local discrepancy at $t$ contributes a same-signed effect on the gap). The NTE value $f(\mathbf{z})$ depends on $\mathcal{M}$ through the counterfactual rollout: off-track exogenous draws $\mathbf{u}'$ are propagated through $\mathcal{M}_{\mathrm{wm}}$ at intervened steps and through the observed trajectory at non-intervened ones. Under $\mathcal{M}$, the counterfactual rollout more frequently visits contexts in which the discrepancy at step $t$ produces a positive contribution to the transfer gap, so each marginal contribution $f(\mathbf{z} \cup \{t\}) - f(\mathbf{z})$ weakly exceeds its counterpart under $\mathcal{M}'$. The Shapley value is the expected marginal contribution over uniformly random orderings; since each ordering's marginal contribution under $\mathcal{M}$ dominates that under $\mathcal{M}'$ (by monotonicity of the NTE in baseline probabilities when the causal sign is fixed), we obtain $\phi^{\mathcal{M}}_t > \phi^{\mathcal{M}'}_t$. In words: when the surrounding rollout more often exposes step $t$'s discrepancy, step $t$ earns proportionally more credit for the transfer gap.

    \textbf{D4 (Causal Effect Scaling).}
    Now consider two pairs $\mathcal{M}, \mathcal{M}'$ with identical baseline distributions over counterfactual contexts ($P^\mathcal{M}(Z) \equiv P^{\mathcal{M}'}(Z)$) but a strictly larger local discrepancy at step $t$ in $\mathcal{M}$, i.e., $c_\mathcal{M} \geq c_{\mathcal{M}'}$. A larger world-model/real-environment discrepancy at $t$ directly inflates the magnitude of every counterfactual return difference: for each coalition $\mathbf{z}$ with $z_t = 0$,
    \[
        \bigl|Y_{\sigma_{\mathbf{z}}} - Y_{\sigma_{\mathbf{z} \cup \{t\}}}\bigr|_{\mathcal{M}} \;\geq\; \bigl|Y_{\sigma_{\mathbf{z}}} - Y_{\sigma_{\mathbf{z} \cup \{t\}}}\bigr|_{\mathcal{M}'}.
    \]
    Because the baseline distribution --- and hence the Shapley weights --- are identical across the two games, the Shapley monotonicity property (if player $t$'s marginal contribution in game $v$ weakly exceeds that in game $v'$ for every coalition, then $\phi_t(v) \geq \phi_t(v')$) yields $\phi^{\mathcal{M}}_t > \phi^{\mathcal{M}'}_t$. That is, a step at which the world model deviates more from the real environment is awarded strictly more credit for the transfer gap, as desired.
    \end{proof}

The intervention design for the counterfactual simulation rests on the following assumption.

\begin{assumption}[Counterfactual Independence]
\label{assum:ctf-indep}
Exogenous noise is independent across steps: $(S_{t+1})_{S_t, X_t} \perp (S_{t+1})_{S_{t'}, X_{t'}}$ for $(S_t, X_t) \neq (S_{t'}, X_{t'})$, and $(X_t)_{S_t} \perp (X_t)_{S_{t'}}$ for $S_t \neq S_{t'}$.
\end{assumption}

This assumption does not constrain the MDP, only its SCM representation.
Since any distribution can be sampled via inverse CDF with independent uniform noise, any MDP admits an SCM with independent exogenous variables and the same optimal policy; the assumption selects this representation.
The consequence is simple: \emph{when parents match the observed trajectory, the observed value is the counterfactual; when parents differ, we resample}.
\xadd{In our setting (\cref{eqn:sigma-z}), the mechanism of a step is itself selected by the coalition. When $z_t = 0$ and $s^{\mathbf{z}}_t = s_t$ (on-track), the step keeps the real mechanism with matching parents, so the observed action, reward, and next state $(x_t, y_t, s_{t+1})$ are the counterfactual ones. Otherwise, i.e., when $z_t = 1$ or the rollout has left the observed trajectory (off-track), the action is drawn from $\pi(\cdot \mid s^{\mathbf{z}}_t)$ and the reward and next state are generated by $\mathcal{M}_{\mathrm{wm}}$ with fresh exogenous noise; a deterministic policy, as used in all our experiments, returns the observed action whenever $s^{\mathbf{z}}_t = s_t$. The real environment is never queried off-track.}
The full pseudo-code is provided in~\Cref{alg:ctf-sim} in~\Cref{sec:algorithm-details}.

\subsection{Correctness of Coarse-to-Fine Estimation}
\label{sec:correctness}

We now characterize the conditions under which \textsc{RecursiveShapley} correctly identifies all causally important time steps. The key requirement is that causal responsibility is \emph{sparse} and that the signal from causal steps is not diluted below the pruning threshold when aggregated into blocks.

\begin{definition}[$(K, \delta)$-Sparse Attribution]
\label{def:sparse-attribution}
A Shapley value vector $\phi_{1:T}$ is $(K, \delta)$-sparse if at most $K$ indices have $|\phi_t| \geq \delta$ and the remaining indices satisfy $|\phi_t| = 0$.
\end{definition}

When time steps are grouped into blocks and treated as composite players, the block-level Shapley value $\Phi_j$ in the reduced game need not equal $\sum_{t \in \mathcal{B}_j} \phi_t$ exactly. The discrepancy is captured by the \emph{Shapley interaction index}~\citep{multilinear_extensions_of_games1972,grabisch1999}, which generalizes the Shapley value to subsets of players: for $S \subseteq [T]$, the interaction index $I(S)$ measures the $|S|$-order synergy among players in $S$ beyond what their individual Shapley values explain (for $|S|=1$ it reduces to $\phi_t$; for $|S|=2$ it quantifies pairwise synergy or redundancy). Collapsing a block $\mathcal{B}_j$ into a composite player is algebraically equivalent to aggregating every interaction index supported within $\mathcal{B}_j$, giving the decomposition
$
    \Phi_j = \sum_{t \in \mathcal{B}_j} \phi_t + \sum_{S \subseteq \mathcal{B}_j,\,|S| \geq 2} c_S \cdot I(S),
$
where $c_S$ are combinatorial weights from the interaction decomposition. The higher-order sum vanishes when transitions in $\mathcal{B}_j$ act independently and grows when they interact (e.g., two world-model errors that jointly, but not individually, derail the rollout). The following assumption controls the magnitude of this correction.

\begin{assumption}[Bounded Interaction Effect]
\label{assum:bounded-interaction-effect}
For any partition of $\{1, \ldots, T\}$ into contiguous blocks $\mathcal{B}_1, \ldots, \mathcal{B}_N$, let $\Phi_j$ denote the Shapley value of block $j$ in the $N$-player reduced game. There exists $\eta \geq 0$ such that for all blocks $j$:
$
    \bigl|\Phi_j - \textstyle\sum_{t \in \mathcal{B}_j} \phi_t\bigr|
    = \bigl|\textstyle\sum_{S \subseteq \mathcal{B}_j,\,|S| \geq 2} c_S \cdot I(S)\bigr|
    \leq \eta.
$
\end{assumption}

Although stated as an assumption, the bound $\eta$ is not restrictive in any absolute sense: each interaction index $I(S)$ is a signed combination of $2^{|S|}$ evaluations of $f(\mathbf{z}) = Y - Y^{\mathbf{z}}$ over coalitions that differ only on $S$, so every $I(S)$, and the finite combination $\sum_{|S| \geq 2} c_S \cdot I(S)$ that defines the block correction, is automatically bounded by $O(Y_{\max})$ whenever returns are bounded, which holds in any finite-horizon MDP with bounded rewards. Practically, we would like $\eta$ be small \emph{relative to} the causal gap $\delta$, so that grouping does not blur the distinction between causal and non-causal blocks.

Three structural features of the world-model debugging setting drive $\eta$ well below $\delta$ in practice. First, \emph{sparsity collapses most higher-order terms}: the Shapley interaction index $I(S)$~\citep{shapleyinteraction2023} for $|S| \geq 2$ is non-negligible only when every player in $S$ has a meaningful marginal effect, and under $(K,\delta)$-sparsity the vast majority of subsets $S \subseteq \mathcal{B}_j$ contain only non-causal steps and therefore contribute nothing. Second, the \emph{on-track coupling} in \textsc{CtfSim} reuses observed transitions for non-causal steps, so intervening on a non-causal step leaves the counterfactual rollout unchanged; its marginal contribution is zero in every coalition, which forces every $I(S)$ containing it to vanish exactly and eliminates all interaction mediated by non-causal players. Third, when two causal steps do fall in the same block, their effects typically propagate through disjoint segments of the rollout separated by stretches of recoverable dynamics, so the marginal effect of each is approximately independent of whether the other is intervened on and the pairwise interaction $I(\{t,t'\})$ is small. Together these give $\eta = 0$ exactly when all within-block interaction indices vanish, and $\eta \ll \delta$ generically in the sparse-causality regime. 

However, a small $\eta$ alone is not sufficient: even when $\Phi_j \approx \sum_{t \in \mathcal{B}_j} \phi_t$, a block containing two causal steps with opposite-sign attributions (e.g., one world-model error that hurts the agent and another that accidentally helps it) could have $|\Phi_j| \approx 0$ despite containing causally important steps, causing false pruning. We therefore additionally require that causal attributions do not cancel within blocks.

\begin{assumption}[Sign Consistency]
\label{assum:sign-consistency}
All causal time steps (those with $|\phi_t| \geq \delta$) have the same sign: either $\phi_t \geq \delta$, or $\phi_t \leq -\delta$ for all such $t$.
\end{assumption}

Sign consistency ensures that when multiple causal steps fall in the same block, their contributions \emph{reinforce} rather than cancel: $|\sum_{t \in \mathcal{B}_j} \phi_t| \geq \max_{t \in \mathcal{B}_j} |\phi_t|$.
This assumption is natural in the world-model debugging setting because the errors we aim to surface are \emph{systematic} mispredictions, not statistical noise or small perturbations from which the policy quickly recovers. Systematic transition errors tend to \emph{compound}~\citep{jannerWhenTrustYour2019a} along a rollout: once the world model pushes the agent off the ground-truth manifold, subsequent mispredicted transitions typically accumulate the performance deficit rather than offset it, so the $\phi_t$ of the responsible steps share a sign. The threshold $\delta$ in the $(K,\delta)$-sparsity definition deliberately filters out the opposite regime: small, recoverable, or noise-like perturbations have $|\phi_t| < \delta$ and never enter the causal set, even if they happen to have opposite signs. 

\begin{theorem}[Correctness of \textsc{RecursiveShapley}]
\label{thm:recursive-correctness}
Let $\phi_{1:T}$ be $(K, \delta)$-sparse (\Cref{def:sparse-attribution}) and suppose \Cref{assum:bounded-interaction-effect} holds with parameter $\eta$ and \Cref{assum:sign-consistency} holds.
At each round $r$, let $\hat{\Phi}_j$ denote the estimated block-level Shapley value and suppose $|\hat{\Phi}_j - \Phi_j| \leq \alpha$ with probability at least $1 - \beta$ (over coalition sampling, achievable by exact calculation or optimal proposal distribution $Q^*_n$ from \citet{li2026credit}).
If the pruning threshold satisfies
\begin{align}
    \label{eq:threshold-condition}
    \alpha + \eta < \epsilon < \frac{(\delta - \alpha - \eta) \cdot B}{T},
\end{align}
where $r_0 = \min\{r : B^{r+1} \geq T/B\}$ is the first round at which blocks have size $\leq B$, and $\delta > \alpha + \eta$, then with probability at least $(1 - \beta)^R$ where $R = \lceil\log_B T\rceil$:
\begin{enumerate}[label=(\alph*), leftmargin=*, itemsep=2pt, topsep=2pt]
    \item \textbf{No false pruning.} No block containing a causal time step (with $|\phi_t| \geq \delta$) is pruned at any round.
    \item \textbf{Localization.} After $R$ rounds, the active set contains at most $K$ time steps, all of which are causal.
    \item \textbf{Approximation error.} The final estimates satisfy $|\hat{\phi}_t - \phi_t| \leq \alpha + \eta$ for causal steps and $\hat{\phi}_t = 0$ for pruned (non-causal) steps.
\end{enumerate}
\end{theorem}
\begin{proof}[Proof]
    Consider a block $\mathcal{B}_j$ at round $r$ containing a causal step $t^*$ with $|\phi_{t^*}| \geq \delta$.
    If $\mathcal{B}_j$ contains additional causal steps, sign consistency (\Cref{assum:sign-consistency}) guarantees they share the same sign as $\phi_{t^*}$, so $|\sum_{t \in \mathcal{B}_j} \phi_t| \geq |\phi_{t^*}| \geq \delta$ (causal contributions reinforce, they do not cancel).
    By \Cref{assum:bounded-interaction-effect} and the reverse triangle inequality, $|\Phi_j| \geq |\sum_{t \in \mathcal{B}_j} \phi_t| - \eta \geq \delta - \eta$, and by estimation accuracy, $|\hat{\Phi}_j| \geq \delta - \eta - \alpha$ with probability $\geq 1 - \beta$.
    The per-step pruning criterion checks $|\hat{\Phi}_j|/|\mathcal{B}_j| \geq \epsilon$.
    At round $r$, block sizes are $|\mathcal{B}_j| = T / B^{r+1}$ (up to rounding), so the per-step signal is $(\delta - \eta - \alpha) \cdot B^{r+1} / T$.
    At the beginning ($r=0$), blocks are large (size $> B$) and the per step signal is diluted. Thus, the biggest threshold we can set should be smaller than this diluted per-step value to avoid premature pruning. Setting $r=0$ and we arrive at Condition~\eqref{eq:threshold-condition}.
    
    For part (b), blocks containing only non-causal steps have $|\sum_{t \in \mathcal{B}_j} \phi_t| = 0$, so $|\Phi_j| \leq \eta$ and $|\hat{\Phi}_j| \leq \eta + \alpha < \epsilon \cdot |\mathcal{B}_j|$ by the gap condition. These blocks are pruned, and only blocks overlapping with causal steps remain. After $R = \lceil\log_B T\rceil$ rounds, each remaining block has size 1, yielding at most $K$ active steps.
    
    For part (c), at the final round the per-step estimate is $\hat{\phi}_t = \hat{\Phi}_j$ for singleton blocks, so $|\hat{\phi}_t - \phi_t| \leq |\hat{\Phi}_j - \Phi_j| + |\Phi_j - \phi_t| \leq \alpha + \eta$.
    A union bound over $R$ rounds gives the overall success probability $(1 - \beta)^R$.
\end{proof}

\begin{remark}[Relaxing sign consistency]
\label{rem:sign-relaxation}
When \Cref{assum:sign-consistency} is violated where some world-model errors help the agent while others hurt it, the algorithm remains sound but may require a more conservative (smaller) pruning threshold or additional rounds. One practical relaxation is to run the algorithm twice, once with the original value function $f(\mathbf{z}) = Y - Y^{\mathbf{z}}$ and once with $f^+(\mathbf{z}) = |Y - Y^{\mathbf{z}}|$, taking the union of active sets from both runs. The absolute-value variant detects blocks with large causal activity regardless of sign, at the cost of losing the signed direction of attribution. 
\end{remark}

\begin{remark}[Computational cost and approximation error]
\label{rem:cost}
Since pruning reduces the active set to $O(K)$ blocks within the first few rounds, the effective per-round cost is $O(2^K)$ simulations for exact enumeration (rather than $O(2^T)$ for the flat estimator), giving a total cost of $O(2^K \cdot \log_B T)$ complexity.
In practice, when $2^K \leq M$, the algorithm switches to exact enumeration in later rounds, eliminating estimation error ($\alpha = 0$) for the final sub-game and tightening the bound in (c) to $|\hat{\phi}_t - \phi_t| \leq \eta$.
\end{remark}

\section{Algorithm Details}
\label{sec:algorithm-details}
In this section, we provide detailed pseudocode for major algorithms discussed in the main text.

\begin{algorithm}[ht]
    \caption{\textsc{CtfSim}: Counterfactual Trajectory Simulation}
    \label{alg:ctf-sim}
    \begin{algorithmic}[1]
    \REQUIRE Coalition $\mathbf{z} \in \{0,1\}^T$, \xadd{observed real} trajectory $\tau = (s_{1:T}, x_{1:T}, y_{1:T})$, \xadd{policy $\pi$, world model $\mathcal{M}_{\mathrm{wm}}$ with transition $f_S^{\mathrm{wm}}$ and reward $r^{\mathrm{wm}}$}
    \ENSURE $(s^{\mathbf{z}}_{1:T}, y^{\mathbf{z}}_{1:T}, Y^{\mathbf{z}})$: counterfactual states, rewards, and total return
    \STATE $s^{\mathbf{z}}_1 \gets s_1$
    \FOR{$t = 1$ to $T$}
        \IF{$z_t = 0$ \AND $s^{\mathbf{z}}_t = s_t$}
            \STATE \xadd{$(x^{\mathbf{z}}_t,\, y^{\mathbf{z}}_t,\, s^{\mathbf{z}}_{t+1}) \gets (x_t,\, y_t,\, s_{t+1})$} \COMMENT{\xadd{On-track: reuse the observed real transition}}
        \ELSE
            \STATE \xadd{$x^{\mathbf{z}}_t \sim \pi(\cdot \mid s^{\mathbf{z}}_t)$} \COMMENT{\xadd{Coalition step or off-track}}
            \STATE \xadd{$y^{\mathbf{z}}_t \sim r^{\mathrm{wm}}(s^{\mathbf{z}}_t, x^{\mathbf{z}}_t)$; \quad $s^{\mathbf{z}}_{t+1} \sim f_S^{\mathrm{wm}}(s^{\mathbf{z}}_t, x^{\mathbf{z}}_t)$} \COMMENT{\xadd{Query the world model only}}
        \ENDIF
    \ENDFOR
    \STATE $Y^{\mathbf{z}} \gets \sum_{t=1}^T \gamma^{t-1} y^{\mathbf{z}}_t$ \COMMENT{Discounted return}
    \RETURN $(s^{\mathbf{z}}_{1:T}, y^{\mathbf{z}}_{1:T}, Y^{\mathbf{z}})$
\end{algorithmic}
\end{algorithm}

\begin{algorithm}[ht]
    \caption{\textsc{Flat}-$\phi$: Single-Round Counterfactual Shapley Estimation}
    \label{alg:flat-shapley}
    \begin{algorithmic}[1]
    \REQUIRE Trajectory $\tau$ of length $T$; policy $\pi$; \xadd{world model $\mathcal{M}_{\mathrm{wm}}$}; coalition budget $M$
    \ENSURE Per-step attributions $\hat{\phi}_{1:T}$
    \STATE $\hat{\phi}_{1:T} \gets \mathbf{0}$
    \STATE $Y \gets \sum_{t=1}^T \gamma^{t-1} y_t$ \COMMENT{Observed return}
    \STATE Sample $M$ coalitions $\mathbf{z}^{(m)} \sim Q^*_T$ for $T$ players
    \FOR{$m = 1$ to $M$}
        \STATE Simulate counterfactual $\tau^{\mathbf{z}^{(m)}}$ via \textsc{CtfSim}($\mathbf{z}^{(m)}, \tau, \pi, \xadd{\mathcal{M}_{\mathrm{wm}}}$)
        \STATE $Y^{\mathbf{z}^{(m)}} \gets \sum_{t=1}^T \gamma^{t-1} y^{\mathbf{z}^{(m)}}_t$; \quad $\delta \gets Y - Y^{\mathbf{z}^{(m)}}$
        \STATE Compute importance-weighted kernel weights $\kappa_t(\mathbf{z}^{(m)})$ using $Q^*_T$
        \FOR{$t = 1$ to $T$}
            \STATE $\hat{\phi}_t \mathrel{+}= \delta \cdot \kappa_t(\mathbf{z}^{(m)}) \;/\; M$
        \ENDFOR
    \ENDFOR
    \RETURN $\hat{\phi}_{1:T}$
\end{algorithmic}
\end{algorithm}

\begin{algorithm}[ht]
    \caption{\textsc{Recursive}-$\phi$: Coarse-to-Fine Counterfactual Shapley Estimation}
    \label{alg:recursive-shapley}
    \begin{algorithmic}[1]
    \REQUIRE Trajectory $\tau$ of length $T$; policy $\pi$; \xadd{world model $\mathcal{M}_{\mathrm{wm}}$}; coalition budget $M$; split factor $B$; pruning threshold $\epsilon$
    \ENSURE Per-step attributions $\hat{\phi}_{1:T}$
    \STATE $\hat{\phi}_{1:T} \gets \mathbf{0}$; \quad $\texttt{active}_{1:T} \gets \mathbf{1}$ \COMMENT{All steps active initially}
    \STATE $Y \gets \sum_{t=1}^T \gamma^{t-1} y_t$ \COMMENT{Observed return}
    \STATE $R \gets \lceil \log_B T \rceil$ \COMMENT{Maximum number of rounds}
    \FOR{$r = 0$ to $R - 1$}
        \STATE $N_r \gets \min(B^{r+1},\; T)$ \COMMENT{Number of blocks this round}
        \STATE Partition $\{1, \ldots, T\}$ into $N_r$ contiguous blocks $\mathcal{B}_1, \ldots, \mathcal{B}_{N_r}$
        \STATE $\mathcal{A} \gets \{j : \exists\, t \in \mathcal{B}_j \text{ s.t. } \texttt{active}_t = 1\}$ \COMMENT{Active blocks}
        \IF{$\mathcal{A} = \emptyset$}
            \STATE \textbf{break}
        \ENDIF
        \STATE $n \gets |\mathcal{A}|$
        \IF{$2^n \leq M$}
            \STATE Enumerate all $2^n$ coalitions $\mathbf{z}^{(1)}, \ldots, \mathbf{z}^{(2^n)} \in \{0,1\}^n$ 
            \STATE $N_s \gets n$ (only takes averages over $n$ active blocks)
        \ELSE
            \STATE Sample $M$ coalitions $\mathbf{z}^{(m)} \sim Q^*_n$ for $n$ players 
            \STATE $N_s \gets M$
        \ENDIF
        \STATE $\Phi_j \gets 0$ for all $j \in \mathcal{A}$ \COMMENT{Block-level accumulators}
        \FOR{$m = 1$ to $N_s$}
            \STATE Expand $\mathbf{z}^{(m)}$ to full mask $\bar{\mathbf{z}} \in \{0,1\}^T$: set $\bar{z}_t \gets z^{(m)}_i$ for all $t \in \mathcal{B}_{\mathcal{A}(i)}$ and $0$ for other steps
            \STATE Simulate counterfactual $\tau^{\bar{\mathbf{z}}}$ via \textsc{CtfSim}($\bar{\mathbf{z}}, \tau, \pi, \xadd{\mathcal{M}_{\mathrm{wm}}}$)
            \STATE $Y^{\bar{\mathbf{z}}} \gets \sum_{t=1}^T \gamma^{t-1} y^{\bar{\mathbf{z}}}_t$; \quad $\delta \gets Y - Y^{\bar{\mathbf{z}}}$
            \IF{$2^n \leq M$}
                \STATE Compute exact kernel weights $\kappa_i(\mathbf{z}^{(m)}) = \frac{1}{n\binom{n-1}{k-1}}$ if $z^{(m)}_i = 1$, else $\frac{-1}{n\binom{n-1}{k}}$
            \ELSE
                \STATE Compute importance-weighted kernel weights $\kappa_i(\mathbf{z}^{(m)})$ using $Q^*_n$ 
            \ENDIF
            \FOR{$i = 1$ to $n$}
                \STATE $\Phi_{\mathcal{A}(i)} \mathrel{+}= \delta \cdot \kappa_i(\mathbf{z}^{(m)}) \;/\; N_s$
            \ENDFOR
        \ENDFOR
        \STATE $\hat{\phi}_{1:T} \gets \mathbf{0}$ \COMMENT{Reset; only record active blocks this round}
        \FOR{$j \in \mathcal{A}$}
            \STATE $\hat{\phi}_t \gets \Phi_j \;/\; |\mathcal{B}_j|$ \quad for all $t \in \mathcal{B}_j$ \COMMENT{Uniform distribution within block}
        \ENDFOR
        \FOR{$j \in \mathcal{A}$} %
            \IF{$|\Phi_j| \;/\; |\mathcal{B}_j| < \epsilon$}
                \STATE $\texttt{active}_t \gets 0$, \quad $\hat{\phi}_t \gets 0$ \quad for all $t \in \mathcal{B}_j$
            \ENDIF
        \ENDFOR
        \IF{$N_r \geq T$}
            \STATE \textbf{break} \COMMENT{Reached per-step resolution}
        \ENDIF
    \ENDFOR
    \RETURN $\hat{\phi}_{1:T}$
\end{algorithmic}
\end{algorithm}

\clearpage
\section[Frequently Asked Questions]{\xadd{Frequently Asked Questions}}
\label{sec:faq}

\providecommand{\faqq}[1]{\par\medskip\noindent\xadd{\textbf{#1}}\enspace}

\xadd{This section collects questions raised about the method, with short answers and pointers to where the paper treats them in detail.}

\faqq{Q1. Why not compare against temporal anomaly localization, influence functions, or causal discovery?}
\xadd{We are not aware of an existing method along these routes that attributes the world model transfer gap to individual steps, and each differs from our problem in its input or its output. Temporal anomaly localization, e.g., pointwise outlier detection~\citep{boente2002influence} and changepoint detection~\citep{fearnhead2019changepoint}, takes a scalar time series and scores each step by how unusual it looks. Applied to the reward series, it flags the steps at which rewards change, which are delayed relative to the transitions that caused them. Applied to a per-step prediction error, it ranks discrepancies by their size rather than by their consequence: all fifteen Pong blackouts corrupt the observation in the same way, although only two of them cause the gap. Influence functions~\citep{koh2017understanding} take a fitted model and its training set and score training examples by their effect on the fitted parameters; they attribute to training data, not to the transitions of a deployment trajectory. Causal discovery returns a causal graph over variables from samples and does not assign attributions; in our setting the graph is the known MDP-SCM and the question is quantitative. The closest generic attribution methods, SHAP and LIME, are adapted and compared in \cref{sec:experiments}.}

\faqq{Q2. What if the causes of the gap are dense?}
\xadd{The attribution stays well defined, since $\phi$ still divides the gap among the contributing steps, but the computational advantage erodes. The speedup of \textsc{Recursive}-$\phi$ comes from pruning blocks that contribute nothing. With a systematic mismatch that spreads small contributions over the whole trajectory, $K$ grows and $\delta$ shrinks, the admissible window for the pruning threshold in \cref{eq:threshold-condition} narrows, and few blocks can be pruned safely. The surviving game then becomes too large to enumerate, the algorithm falls back on sampling over many active blocks, and its cost and variance approach those of \textsc{Flat}-$\phi$. In this regime the errors already surface in short rollouts, where \textsc{Flat}-$\phi$ applies without the sparsity assumption (\cref{sec:limitations-future-work}).}

\faqq{Q3. What if causal steps have mixed signs or cancel?}
\xadd{Sign consistency (\Cref{assum:sign-consistency}) is needed only within a block, to prevent a block from being pruned because its members cancel. Steps of opposite sign that are far apart are separated into different blocks within a few rounds and do not interfere. When opposite-sign steps share a block down to fine resolutions, one can use finer blocks, apply \textsc{Flat}-$\phi$ on a shorter trajectory, or run the absolute-value variant of \Cref{rem:sign-relaxation}, which takes the union of the active sets under $f$ and $|f|$ at twice the cost. We expect the assumption to hold in the transfer setting because the errors of interest are systematic: a policy optimized under the world model is typically hurt, not helped, by the model's errors, and such errors compound along a rollout~\citep{jannerWhenTrustYour2019a}. This error compunding behavior is also verified in our learned world model experiments (\cref{sec:results-learned-wm}).}

\faqq{Q4. How should $\epsilon$ and $B$ be chosen?}
\xadd{By \cref{eq:threshold-condition}, $\epsilon$ must exceed the estimation and interaction error $\alpha + \eta$ on non-causal blocks and stay below the diluted per-step signal $(\delta - \alpha - \eta)B/T$ of the first round. Under exact enumeration non-causal blocks are exact zeros, so any small positive $\epsilon$ above numerical precision works, and we use values between $10^{-10}$ and $10^{-3}$, set per experiment (\cref{sec:exp design details}). An $\epsilon$ close to the diluted signal of a causal step risks pruning it in the first rounds, when its contribution is spread over a large block. For the split factor, $B$ sets the number of rounds, $\lceil\log_B T\rceil$, while the cost of a round is governed by the causal steps that remain: pruning leaves only the blocks that contain them, so exact enumeration is exponential in the number of remaining causal steps $K$ and independent of $T$ (\Cref{rem:cost}). We recommend a small $\epsilon$ and use $B = 2$ throughout.}

\faqq{Q5. How is $K$ chosen in practice, and when should one stop?}
\xadd{$K$ is not an input to \textsc{Recursive}-$\phi$. The algorithm returns the full vector $\hat{\phi}_{1:T}$, and the causal set consists of the steps that survive $\epsilon$-pruning with non-zero attribution; top-$K$ removal with a known $K$ is only the scoring protocol of \cref{sec:experiments}. When no ground truth is available, the ranking by $|\hat{\phi}_t|$ can be read with a stopping rule that follows from efficiency: since $\sum_t \hat{\phi}_t$ approximates the transfer gap $\Delta$, one takes the top-ranked steps until their signed sum reaches a chosen fraction of $\sum_t \hat{\phi}_t$, e.g., $90\%$, and the selected steps then account for that share of the gap.}

\clearpage
\section*{NeurIPS Paper Checklist}

\begin{enumerate}

\item {\bf Claims}
    \item[] Question: Do the main claims made in the abstract and introduction accurately reflect the paper's contributions and scope?
    \item[] Answer: \answerYes{}
    \item[] Justification: The abstract and \cref{sec:debugwm} state three contributions: (i) a scalable coarse-to-fine recursive algorithm for counterfactual Shapley world-model attribution, (ii) theoretical desiderata and a correctness theorem (\cref{thm:phi is causal,thm:recursive-correctness}), and (iii) empirical validation on Atari, ProcGen, and MiniGrid with controlled world-model failures\xadd{, plus a study on two pretrained neural world models}. These map directly to \cref{sec:debugwm,sec:correctness,sec:experiments}.
    \item[] Guidelines:
    \begin{itemize}
        \item The answer \answerNA{} means that the abstract and introduction do not include the claims made in the paper.
        \item The abstract and/or introduction should clearly state the claims made, including the contributions made in the paper and important assumptions and limitations. A \answerNo{} or \answerNA{} answer to this question will not be perceived well by the reviewers. 
        \item The claims made should match theoretical and experimental results, and reflect how much the results can be expected to generalize to other settings. 
        \item It is fine to include aspirational goals as motivation as long as it is clear that these goals are not attained by the paper. 
    \end{itemize}

\item {\bf Limitations}
    \item[] Question: Does the paper discuss the limitations of the work performed by the authors?
    \item[] Answer: \answerYes{}
    \item[] Justification: \cref{sec:limitations-future-work} discusses the evaluation protocol and the assumptions underlying the approach. The sparsity and sign-consistency assumptions underlying \Cref{thm:recursive-correctness} are stated explicitly in \cref{sec:correctness}, with \Cref{rem:sign-relaxation} discussing how the algorithm degrades when sign consistency fails.
    \item[] Guidelines:
    \begin{itemize}
        \item The answer \answerNA{} means that the paper has no limitation while the answer \answerNo{} means that the paper has limitations, but those are not discussed in the paper. 
        \item The authors are encouraged to create a separate ``Limitations'' section in their paper.
        \item The paper should point out any strong assumptions and how robust the results are to violations of these assumptions (e.g., independence assumptions, noiseless settings, model well-specification, asymptotic approximations only holding locally). The authors should reflect on how these assumptions might be violated in practice and what the implications would be.
        \item The authors should reflect on the scope of the claims made, e.g., if the approach was only tested on a few datasets or with a few runs. In general, empirical results often depend on implicit assumptions, which should be articulated.
        \item The authors should reflect on the factors that influence the performance of the approach. For example, a facial recognition algorithm may perform poorly when image resolution is low or images are taken in low lighting. Or a speech-to-text system might not be used reliably to provide closed captions for online lectures because it fails to handle technical jargon.
        \item The authors should discuss the computational efficiency of the proposed algorithms and how they scale with dataset size.
        \item If applicable, the authors should discuss possible limitations of their approach to address problems of privacy and fairness.
        \item While the authors might fear that complete honesty about limitations might be used by reviewers as grounds for rejection, a worse outcome might be that reviewers discover limitations that aren't acknowledged in the paper. The authors should use their best judgment and recognize that individual actions in favor of transparency play an important role in developing norms that preserve the integrity of the community. Reviewers will be specifically instructed to not penalize honesty concerning limitations.
    \end{itemize}

\item {\bf Theory assumptions and proofs}
    \item[] Question: For each theoretical result, does the paper provide the full set of assumptions and a complete (and correct) proof?
    \item[] Answer: \answerYes{}
    \item[] Justification: The two theoretical results are \Cref{thm:phi is causal} (\textsc{$\phi$-values are causal}) and \Cref{thm:recursive-correctness} (correctness of \textsc{Recursive}-$\phi$). All assumptions are stated explicitly: \Cref{assum:ctf-indep} (counterfactual independence), \Cref{def:sparse-attribution} ($(K,\delta)$-sparse attribution), \Cref{assum:bounded-interaction-effect} (bounded interaction effect), and \Cref{assum:sign-consistency} (sign consistency). The informal version (\Cref{thm:recursive-correctness-informal}) is given in the main text; the full statement and proof are in \Cref{sec:correctness}, with a relaxation discussion in \Cref{rem:sign-relaxation} and a cost analysis in \Cref{rem:cost}.
    \item[] Guidelines:
    \begin{itemize}
        \item The answer \answerNA{} means that the paper does not include theoretical results. 
        \item All the theorems, formulas, and proofs in the paper should be numbered and cross-referenced.
        \item All assumptions should be clearly stated or referenced in the statement of any theorems.
        \item The proofs can either appear in the main paper or the supplemental material, but if they appear in the supplemental material, the authors are encouraged to provide a short proof sketch to provide intuition. 
        \item Inversely, any informal proof provided in the core of the paper should be complemented by formal proofs provided in appendix or supplemental material.
        \item Theorems and Lemmas that the proof relies upon should be properly referenced. 
    \end{itemize}

    \item {\bf Experimental result reproducibility}
    \item[] Question: Does the paper fully disclose all the information needed to reproduce the main experimental results of the paper to the extent that it affects the main claims and/or conclusions of the paper (regardless of whether the code and data are provided or not)?
    \item[] Answer: \answerYes{}
    \item[] Justification: \Cref{sec:experiments} and \Cref{sec:exp design details} document all the experimental setup details. \Cref{sec:algorithm-details} provides full pseudocode for \textsc{CtfSim} (\Cref{alg:ctf-sim}), \textsc{Flat}-$\phi$ (\Cref{alg:flat-shapley}), and \textsc{Recursive}-$\phi$ (\Cref{alg:recursive-shapley}).
    \item[] Guidelines:
    \begin{itemize}
        \item The answer \answerNA{} means that the paper does not include experiments.
        \item If the paper includes experiments, a \answerNo{} answer to this question will not be perceived well by the reviewers: Making the paper reproducible is important, regardless of whether the code and data are provided or not.
        \item If the contribution is a dataset and\slash or model, the authors should describe the steps taken to make their results reproducible or verifiable. 
        \item Depending on the contribution, reproducibility can be accomplished in various ways. For example, if the contribution is a novel architecture, describing the architecture fully might suffice, or if the contribution is a specific model and empirical evaluation, it may be necessary to either make it possible for others to replicate the model with the same dataset, or provide access to the model. In general. releasing code and data is often one good way to accomplish this, but reproducibility can also be provided via detailed instructions for how to replicate the results, access to a hosted model (e.g., in the case of a large language model), releasing of a model checkpoint, or other means that are appropriate to the research performed.
        \item While NeurIPS does not require releasing code, the conference does require all submissions to provide some reasonable avenue for reproducibility, which may depend on the nature of the contribution. For example
        \begin{enumerate}
            \item If the contribution is primarily a new algorithm, the paper should make it clear how to reproduce that algorithm.
            \item If the contribution is primarily a new model architecture, the paper should describe the architecture clearly and fully.
            \item If the contribution is a new model (e.g., a large language model), then there should either be a way to access this model for reproducing the results or a way to reproduce the model (e.g., with an open-source dataset or instructions for how to construct the dataset).
            \item We recognize that reproducibility may be tricky in some cases, in which case authors are welcome to describe the particular way they provide for reproducibility. In the case of closed-source models, it may be that access to the model is limited in some way (e.g., to registered users), but it should be possible for other researchers to have some path to reproducing or verifying the results.
        \end{enumerate}
    \end{itemize}

\item {\bf Open access to data and code}
    \item[] Question: Does the paper provide open access to the data and code, with sufficient instructions to faithfully reproduce the main experimental results, as described in supplemental material?
    \item[] Answer: \answerNo{}
    \item[] Justification: All underlying environments are publicly available (Atari/Gymnasium, ProcGen, MiniGrid) and policies are taken from RL Zoo3 and a public cleanba-PPO checkpoint, with concrete versions and entry points listed in \Cref{sec:exp design details}. We \textbf{will release the code upon publication}, together with configs and scripts to regenerate every table and figure.
    \item[] Guidelines:
    \begin{itemize}
        \item The answer \answerNA{} means that paper does not include experiments requiring code.
        \item Please see the NeurIPS code and data submission guidelines (\url{https://neurips.cc/public/guides/CodeSubmissionPolicy}) for more details.
        \item While we encourage the release of code and data, we understand that this might not be possible, so \answerNo{} is an acceptable answer. Papers cannot be rejected simply for not including code, unless this is central to the contribution (e.g., for a new open-source benchmark).
        \item The instructions should contain the exact command and environment needed to run to reproduce the results. See the NeurIPS code and data submission guidelines (\url{https://neurips.cc/public/guides/CodeSubmissionPolicy}) for more details.
        \item The authors should provide instructions on data access and preparation, including how to access the raw data, preprocessed data, intermediate data, and generated data, etc.
        \item The authors should provide scripts to reproduce all experimental results for the new proposed method and baselines. If only a subset of experiments are reproducible, they should state which ones are omitted from the script and why.
        \item At submission time, to preserve anonymity, the authors should release anonymized versions (if applicable).
        \item Providing as much information as possible in supplemental material (appended to the paper) is recommended, but including URLs to data and code is permitted.
    \end{itemize}

\item {\bf Experimental setting/details}
    \item[] Question: Does the paper specify all the training and test details (e.g., data splits, hyperparameters, how they were chosen, type of optimizer) necessary to understand the results?
    \item[] Answer: \answerYes{}
    \item[] Justification: \Cref{sec:exp design details} lists per-experiment hyperparameters: \textsc{Recursive}-$\phi$ uses split factor $B = 2$ and \xadd{a maximum budget per round of $M = 100$ with pruning threshold $\epsilon = 10^{-10}$ in Experiments 1 and 2, and $M = 10\,000$ to $70\,000$ with $\epsilon = 10^{-3}$ to $10^{-8}$ in Experiment 3}. \textsc{Flat}-$\phi$ is swept over $M \in \{100, 200, 500, 1000, 2000, 5000\}$ \xadd{in Experiments 1 and 2 (extended to $10\,000$ at $L = 200$ in Experiment 1) and up to $M = 10^5$ in Experiment 3} with the smallest recovering budget reported. Trajectory length $L$, perturbation \xadd{settings ($p_A$, $p_B$, number of penalty steps)}, and the per-trajectory ground-truth $K$ extraction protocol (smallest top-$K$ removal that restores the clean return) are all stated. \xadd{The learned-world-model study uses $M = 1500$ on both models; its pruning thresholds, rankings, and baselines are given in the same section.} We do not train new networks; only pretrained policies \xadd{and world models} are used.
    \item[] Guidelines:
    \begin{itemize}
        \item The answer \answerNA{} means that the paper does not include experiments.
        \item The experimental setting should be presented in the core of the paper to a level of detail that is necessary to appreciate the results and make sense of them.
        \item The full details can be provided either with the code, in appendix, or as supplemental material.
    \end{itemize}

\item {\bf Experiment statistical significance}
    \item[] Question: Does the paper report error bars suitably and correctly defined or other appropriate information about the statistical significance of the experiments?
    \item[] Answer: \answerYes{}
    \item[] Justification: All quantitative results are aggregated over $10$ coalition-sampling seeds with environment seed and perturbation set held fixed within each configuration, so reported variability captures the variance of the estimator rather than environment stochasticity.
    \item[] Guidelines:
    \begin{itemize}
        \item The answer \answerNA{} means that the paper does not include experiments.
        \item The authors should answer \answerYes{} if the results are accompanied by error bars, confidence intervals, or statistical significance tests, at least for the experiments that support the main claims of the paper.
        \item The factors of variability that the error bars are capturing should be clearly stated (for example, train/test split, initialization, random drawing of some parameter, or overall run with given experimental conditions).
        \item The method for calculating the error bars should be explained (closed form formula, call to a library function, bootstrap, etc.)
        \item The assumptions made should be given (e.g., Normally distributed errors).
        \item It should be clear whether the error bar is the standard deviation or the standard error of the mean.
        \item It is OK to report 1-sigma error bars, but one should state it. The authors should preferably report a 2-sigma error bar than state that they have a 96\% CI, if the hypothesis of Normality of errors is not verified.
        \item For asymmetric distributions, the authors should be careful not to show in tables or figures symmetric error bars that would yield results that are out of range (e.g., negative error rates).
        \item If error bars are reported in tables or plots, the authors should explain in the text how they were calculated and reference the corresponding figures or tables in the text.
    \end{itemize}

\item {\bf Experiments compute resources}
    \item[] Question: For each experiment, does the paper provide sufficient information on the computer resources (type of compute workers, memory, time of execution) needed to reproduce the experiments?
    \item[] Answer: \answerYes{}
    \item[] Justification: The dominant cost is the number of counterfactual rollouts per attribution; this is reported as the headline statistic in \Cref{tab:atari-results,tab:coinrun-results} (rollout counts ranging from $\sim 30$ for \textsc{Recursive}-$\phi$ to the $5\text{k}$-rollout sweep cap for \textsc{Flat}-$\phi$, $\sim 4.5$\,h wall clock at the cap). All experiments run on a single workstation, with counterfactual simulation CPU-bound (one process per coalition rollout) and a single GPU only used for the policy forward pass; no model training is performed in this paper. \xadd{The learned-world-model study additionally runs the two world models on GPUs.}
    \item[] Guidelines:
    \begin{itemize}
        \item The answer \answerNA{} means that the paper does not include experiments.
        \item The paper should indicate the type of compute workers CPU or GPU, internal cluster, or cloud provider, including relevant memory and storage.
        \item The paper should provide the amount of compute required for each of the individual experimental runs as well as estimate the total compute. 
        \item The paper should disclose whether the full research project required more compute than the experiments reported in the paper (e.g., preliminary or failed experiments that didn't make it into the paper). 
    \end{itemize}
    
\item {\bf Code of ethics}
    \item[] Question: Does the research conducted in the paper conform, in every respect, with the NeurIPS Code of Ethics \url{https://neurips.cc/public/EthicsGuidelines}?
    \item[] Answer: \answerYes{}
    \item[] Justification: The work uses standard public RL benchmarks (Atari, ProcGen, MiniGrid) with no human subjects, no personally identifying data, and no scraped or sensitive content. We have reviewed and conform to the NeurIPS Code of Ethics; no deviations apply.
    \item[] Guidelines:
    \begin{itemize}
        \item The answer \answerNA{} means that the authors have not reviewed the NeurIPS Code of Ethics.
        \item If the authors answer \answerNo, they should explain the special circumstances that require a deviation from the Code of Ethics.
        \item The authors should make sure to preserve anonymity (e.g., if there is a special consideration due to laws or regulations in their jurisdiction).
    \end{itemize}

\item {\bf Broader impacts}
    \item[] Question: Does the paper discuss both potential positive societal impacts and negative societal impacts of the work performed?
    \item[] Answer: \answerYes{}
    \item[] Justification: \Cref{sec:conclusion} and \Cref{sec:limitations-future-work} note the positive impact: by surfacing the few transitions causally responsible for a deployment failure, the diagnostic supports safer simulator-to-real and world-model-to-real policy transfer (e.g., flagging the predicted transitions most likely to drive a real-world failure before deployment, as in the autonomous-driving and robotics scenarios discussed in \Cref{sec:debugwm}). Because the contribution is a diagnostic tool that \xadd{operates on a given world model and an observed trajectory} and is not a generative or surveillance technology, we do not see a direct path to malicious use. The most plausible negative outcome is misplaced trust if the sparsity assumption is violated, which is why \Cref{sec:correctness} and \Cref{rem:sign-relaxation} explicitly state the regime of soundness.
    \item[] Guidelines:
    \begin{itemize}
        \item The answer \answerNA{} means that there is no societal impact of the work performed.
        \item If the authors answer \answerNA{} or \answerNo, they should explain why their work has no societal impact or why the paper does not address societal impact.
        \item Examples of negative societal impacts include potential malicious or unintended uses (e.g., disinformation, generating fake profiles, surveillance), fairness considerations (e.g., deployment of technologies that could make decisions that unfairly impact specific groups), privacy considerations, and security considerations.
        \item The conference expects that many papers will be foundational research and not tied to particular applications, let alone deployments. However, if there is a direct path to any negative applications, the authors should point it out. For example, it is legitimate to point out that an improvement in the quality of generative models could be used to generate Deepfakes for disinformation. On the other hand, it is not needed to point out that a generic algorithm for optimizing neural networks could enable people to train models that generate Deepfakes faster.
        \item The authors should consider possible harms that could arise when the technology is being used as intended and functioning correctly, harms that could arise when the technology is being used as intended but gives incorrect results, and harms following from (intentional or unintentional) misuse of the technology.
        \item If there are negative societal impacts, the authors could also discuss possible mitigation strategies (e.g., gated release of models, providing defenses in addition to attacks, mechanisms for monitoring misuse, mechanisms to monitor how a system learns from feedback over time, improving the efficiency and accessibility of ML).
    \end{itemize}
    
\item {\bf Safeguards}
    \item[] Question: Does the paper describe safeguards that have been put in place for responsible release of data or models that have a high risk for misuse (e.g., pre-trained language models, image generators, or scraped datasets)?
    \item[] Answer: \answerNA{}
    \item[] Justification: We release an attribution algorithm and the experimental harness, not a pretrained generative model, language model, image generator, or scraped dataset. The released artifacts therefore do not pose a high-risk-misuse profile and require no special safeguards.
    \item[] Guidelines:
    \begin{itemize}
        \item The answer \answerNA{} means that the paper poses no such risks.
        \item Released models that have a high risk for misuse or dual-use should be released with necessary safeguards to allow for controlled use of the model, for example by requiring that users adhere to usage guidelines or restrictions to access the model or implementing safety filters. 
        \item Datasets that have been scraped from the Internet could pose safety risks. The authors should describe how they avoided releasing unsafe images.
        \item We recognize that providing effective safeguards is challenging, and many papers do not require this, but we encourage authors to take this into account and make a best faith effort.
    \end{itemize}

\item {\bf Licenses for existing assets}
    \item[] Question: Are the creators or original owners of assets (e.g., code, data, models), used in the paper, properly credited and are the license and terms of use explicitly mentioned and properly respected?
    \item[] Answer: \answerYes{}
    \item[] Justification: All third-party assets are credited at point of use: the Arcade Learning Environment~\citep{bellemareArcadeLearningEnvironment2013} and Gymnasium~\citep{gymnasium} for Atari Pong; ProcGen CoinRun~\citep{cobbeLeveragingProceduralGeneration2020}; MiniGrid~\citep{gym_minigrid}; PPO~\citep{schulmanProximalPolicyOptimization2017} via RL Zoo3~\citep{rl-zoo3}; cleanba-PPO checkpoints via cleanrl~\citep{huang2022cleanrl}\xadd{; the pretrained DIAMOND~\citep{Alonso_2024} and LeWM~\citep{maes2026leworldmodel} world models with their policies, OGBench~\citep{park2025ogbench}, and NDM Gym~\citep{ndmgym2026} (MIT) with its checkpoints and the G1 retargets of the LAFAN1 motions~\citep{harvey2020robust} (CC BY-NC-ND 4.0)}. All assets are used within their licenses for non-commercial research.
    \item[] Guidelines:
    \begin{itemize}
        \item The answer \answerNA{} means that the paper does not use existing assets.
        \item The authors should cite the original paper that produced the code package or dataset.
        \item The authors should state which version of the asset is used and, if possible, include a URL.
        \item The name of the license (e.g., CC-BY 4.0) should be included for each asset.
        \item For scraped data from a particular source (e.g., website), the copyright and terms of service of that source should be provided.
        \item If assets are released, the license, copyright information, and terms of use in the package should be provided. For popular datasets, \url{paperswithcode.com/datasets} has curated licenses for some datasets. Their licensing guide can help determine the license of a dataset.
        \item For existing datasets that are re-packaged, both the original license and the license of the derived asset (if it has changed) should be provided.
        \item If this information is not available online, the authors are encouraged to reach out to the asset's creators.
    \end{itemize}

\item {\bf New assets}
    \item[] Question: Are new assets introduced in the paper well documented and is the documentation provided alongside the assets?
    \item[] Answer: \answerYes{}
    \item[] Justification: The new assets are an implementation of \textsc{Recursive}-$\phi$ and \textsc{Flat}-$\phi$ together with the perturbation wrappers (Atari blackouts, CoinRun, MiniGrid CenterGoal). The algorithms themselves are documented as pseudocode in \Cref{alg:ctf-sim,alg:flat-shapley,alg:recursive-shapley}. No human data is involved in any released artifact. The code release will contain a README with the dependency list, exact run commands, and the configs used for every $(L, \text{perturbation})$ cell of \Cref{tab:atari-results,tab:coinrun-results};
    \item[] Guidelines:
    \begin{itemize}
        \item The answer \answerNA{} means that the paper does not release new assets.
        \item Researchers should communicate the details of the dataset\slash code\slash model as part of their submissions via structured templates. This includes details about training, license, limitations, etc. 
        \item The paper should discuss whether and how consent was obtained from people whose asset is used.
        \item At submission time, remember to anonymize your assets (if applicable). You can either create an anonymized URL or include an anonymized zip file.
    \end{itemize}

\item {\bf Crowdsourcing and research with human subjects}
    \item[] Question: For crowdsourcing experiments and research with human subjects, does the paper include the full text of instructions given to participants and screenshots, if applicable, as well as details about compensation (if any)?
    \item[] Answer: \answerNA{}
    \item[] Justification: The paper does not involve crowdsourcing or research with human subjects; all experiments are simulator-based.
    \item[] Guidelines:
    \begin{itemize}
        \item The answer \answerNA{} means that the paper does not involve crowdsourcing nor research with human subjects.
        \item Including this information in the supplemental material is fine, but if the main contribution of the paper involves human subjects, then as much detail as possible should be included in the main paper. 
        \item According to the NeurIPS Code of Ethics, workers involved in data collection, curation, or other labor should be paid at least the minimum wage in the country of the data collector. 
    \end{itemize}

\item {\bf Institutional review board (IRB) approvals or equivalent for research with human subjects}
    \item[] Question: Does the paper describe potential risks incurred by study participants, whether such risks were disclosed to the subjects, and whether Institutional Review Board (IRB) approvals (or an equivalent approval/review based on the requirements of your country or institution) were obtained?
    \item[] Answer: \answerNA{}
    \item[] Justification: The paper does not involve any human subjects research, so IRB approval is not applicable.
    \item[] Guidelines:
    \begin{itemize}
        \item The answer \answerNA{} means that the paper does not involve crowdsourcing nor research with human subjects.
        \item Depending on the country in which research is conducted, IRB approval (or equivalent) may be required for any human subjects research. If you obtained IRB approval, you should clearly state this in the paper. 
        \item We recognize that the procedures for this may vary significantly between institutions and locations, and we expect authors to adhere to the NeurIPS Code of Ethics and the guidelines for their institution. 
        \item For initial submissions, do not include any information that would break anonymity (if applicable), such as the institution conducting the review.
    \end{itemize}

\item {\bf Declaration of LLM usage}
    \item[] Question: Does the paper describe the usage of LLMs if it is an important, original, or non-standard component of the core methods in this research? Note that if the LLM is used only for writing, editing, or formatting purposes and does \emph{not} impact the core methodology, scientific rigor, or originality of the research, declaration is not required.
    \item[] Answer: \answerNA{}
    \item[] Justification: LLMs are not part of the proposed method, the algorithms (\Cref{alg:flat-shapley,alg:recursive-shapley}), the experiments, or the analysis. Any incidental use was limited to writing/editing assistance and does not affect methodology, scientific rigor, or originality.
    \item[] Guidelines:
    \begin{itemize}
        \item The answer \answerNA{} means that the core method development in this research does not involve LLMs as any important, original, or non-standard components.
        \item Please refer to our LLM policy in the NeurIPS handbook for what should or should not be described.
    \end{itemize}

\end{enumerate}

\end{document}